%% file: iclr2027_conference.tex
\documentclass{article} % For LaTeX2e
\usepackage{iclr2027_conference,times}

\input{math_commands.tex}

\usepackage{hyperref}
\usepackage{url}

\usepackage{booktabs}
\usepackage{siunitx}
\usepackage[table]{xcolor}
\usepackage{graphicx}
\usepackage{subcaption}
\usepackage{booktabs} % for professional tables
\usepackage{hyperref}

\usepackage{amsmath}
\usepackage{amssymb}
\usepackage{mathtools}
\usepackage{amsthm}
\usepackage[capitalize,noabbrev]{cleveref}

\theoremstyle{plain}

\theoremstyle{definition}

\theoremstyle{remark}

\usepackage{amsmath,amssymb}
\usepackage{algorithm}
\usepackage{algpseudocode} 
\usepackage{multirow}

\usepackage{colortbl}

\usepackage{multirow}
\usepackage[table]{xcolor}
\definecolor{advgray}{gray}{0.9}

\definecolor{cleanrow}{RGB}{245,250,255}
\definecolor{advrow}{RGB}{255,245,245}
\definecolor{headrow}{RGB}{235,235,235}

\definecolor{cWhisperM}{RGB}{235,235,235}
\definecolor{cWav2vec}{RGB}{252,252,252}
\definecolor{cMMS}{RGB}{235,235,235}
\definecolor{cWhisperL}{RGB}{252,252,252}

\definecolor{CG}{HTML}{555555}  % CleanGray

\usepackage{array}
\newcolumntype{B}{>{\bfseries}l}
\newcolumntype{C}{>{\bfseries}c}

\usepackage[textsize=tiny]{todonotes}

\usepackage[utf8]{inputenc} % allow utf-8 input
\usepackage[T1]{fontenc}    % use 8-bit T1 fonts
\usepackage{url}            % simple URL typesetting
\usepackage{amsfonts}       % blackboard math symbols
\usepackage{nicefrac}       % compact symbols for 1/2, etc.
\usepackage{microtype}      % microtypography
\usepackage{xcolor}         % colors

\usepackage{wrapfig}

\title{Know When to Hold 'em: Correct-Token Retention in Uniform-State Diffusion Language Models}
\author{Mojtaba Nafez \\
EPFL \& Idiap \\
Lausanne, Switzerland \\
\texttt{mojtaba.nafez@epfl.ch} \\
\And
James Henderson \\
Idiap Research Institute \\
Martigny, Switzerland \\
\texttt{james.henderson@idiap.ch} \\
}

\iclrfinalcopy % Uncomment for camera-ready version, but NOT for submission.
\begin{document}

\maketitle

\input{sections/0_abstract}
\input{sections/1_introduction}
\input{sections/3_preliminary}

\input{sections/4_methods}
\input{sections/5_experiments}
\input{sections/6_ablation_study}
\input{sections/7_conclusion}

\clearpage
\input{sections/7_5_ICLR-Policy}

\bibliography{iclr2027_conference}
\bibliographystyle{iclr2027_conference}

\clearpage

\appendix
\tableofcontents
\section{Appendix}
\input{sections/8_appendix}

\end{document}

%% file: math_commands.tex
\usepackage{amsmath,amsfonts,bm}

\def\eqref#1{equation~\ref{#1}}
\def\1{\bm{1}}

\DeclareMathAlphabet{\mathsfit}{\encodingdefault}{\sfdefault}{m}{sl}
\SetMathAlphabet{\mathsfit}{bold}{\encodingdefault}{\sfdefault}{bx}{n}

%% file: sections/0_abstract.tex
\begin{abstract}
Uniform-state diffusion models (USDMs) can revise any token at any denoising step, which lets them correct their own mistakes, a key advantage over masked diffusion. Self-correction, however, requires both revising incorrect tokens and retaining correct ones, and we show that current USDMs lack the latter.
Even under greedy-tail decoding, 
state-of-the-art USDMs (DUO, UDLM, and uniform-noise SEDD) keep revising 173--270 of 512 positions at every step, and these large, uncoordinated edits collapse sample diversity. A random-token corruption experiment traces this deficit to the models themselves: they reconstruct clean and corrupted tokens with nearly identical accuracy, even though clean tokens are easier targets. A decomposition of the validation NELBO shows that training barely rewards retention: incorrect predictions are heavily penalized at corrupted positions but almost free at clean ones. We propose Correct-Token Retention Regularization (CTR-Reg), a simple but effective auxiliary loss that trains the model to retain tokens left unperturbed by the forward process and requires no change to the sampler. CTR-Reg improves clean-token accuracy by 26.5 percentage points on average across six benchmarks, while leaving corrupted-token accuracy virtually unchanged, and its per-step revisions converge to only 3--11 positions. With just five greedy-tail steps, generative perplexity more than halves under CTR-Reg for all three models 
while diversity is preserved, and these gains hold across sampling budgets.
Our results identify correct-token retention as a key missing ingredient for self-correcting diffusion language models, and demonstrate an effective fix.
\end{abstract}

%% file: sections/1_introduction.tex
\section{Introduction}
Diffusion models have achieved remarkable success in image, video, and audio generation \citep{pmlr-v37-sohl-dickstein15,ho2020denoising,polyak2025moviegencastmedia,huang2023noise2musictextconditionedmusicgeneration,rombach2021highresolution,RetrivalAugment_Diff_model}, driven largely by iterative refinement: progressively correcting errors over many denoising steps yields realistic and diverse samples \citep{zhao2026informed,dhariwal2021diffusion,song2023consistency,ma2025inferencetimescalingdiffusionmodels}. Recent work has extended diffusion to discrete data, with promising results in language modeling \citep{sahoo2025the,shi2024simplified,lou2024discrete,austin2021structured}: unlike left-to-right autoregressive models, diffusion language models decode many tokens in parallel using bidirectional context, enabling substantially faster generation while remaining competitive on downstream tasks \citep{labs2025mercuryultrafastlanguagemodels,song2025seeddiffusionlargescalediffusion}.

Discrete diffusion models interpolate between clean data $x$ and a fixed prior $\pi$ via the forward process $q_t(z_t \mid x; \alpha_t) = \mathrm{Cat}(z_t; \alpha_t x + (1-\alpha_t)\pi)$, with a monotonically decreasing noise schedule $\alpha_t \in [0,1]$ \citep{austin2021structured,sahoo2024simple}. The choice of $\pi$ yields two families. Masked Diffusion Models (MDMs), or absorbing-state diffusion \citep{austin2021structured,sahoo2024simple}, set $\pi$ to a \texttt{[MASK]} token: a masked position is revealed once during denoising and never revisited, permanently locking in any early mistake. Uniform-state Diffusion Models (USDMs) \citep{schiff2024discreteguidance,sahoo2025the} instead set $\pi$ to the uniform distribution over the vocabulary; since a corrupted token is then indistinguishable from a clean one, the model must reconsider every position at every step, enabling self-correction \citep{deschenaux2026the}.

Self-correction, however, requires two distinct capabilities: revising incorrect tokens and retaining correct ones (\emph{correct-token retention}); we find that current USDMs lack the latter. The deficit surfaces under the \emph{Greedy-Tail Sampler} of DUO \citep{sahoo2025the}, which appends one per-position argmax step after ancestral sampling; we also consider $k$ such steps, each a single forward pass followed by a per-position argmax, with no stochastic sampling. Under this sampler, three state-of-the-art USDMs, DUO, UDLM \citep{schiff2024discreteguidance}, and SEDD with uniform noise \citep{lou2024discrete} (hereafter SEDD), exhibit a substantial collapse in sample diversity, measured by entropy (Figure~\ref{fig:greedy-tail-lowstep}).

To identify its source, Figure~\ref{fig:token-change-vs-greedy-steps} tracks how many positions change prediction at each greedy-tail step: 173--270 of the 512 positions are revised at every step, with no sign of convergence. Such churn indicates a lack of correct-token retention: because positions are predicted independently rather than jointly, these large-scale simultaneous edits drive the collapse, since even locally plausible edits, applied by the hundreds at once, can compound into globally inconsistent sequences \citep{gu2018nonautoregressive}. Samplers such as ancestral sampling mitigate this by limiting the average amount of revision per step, but do not determine \emph{which} positions are revised; absent a reliable model preference for correct tokens, this selection is governed by sampling stochasticity.

To disentangle revision from retention, we design a random-token corruption experiment (Figure~\ref{fig:clean-corrupt-bars}): we split a corpus into 512-token chunks, replace a random $p\%$ of tokens in each with tokens drawn uniformly from the vocabulary, reconstruct each sequence with a single forward pass, and report accuracy separately on the corrupted $p\%$ and the clean $(100-p)\%$ of tokens, averaged over $p \in \{10,20,30,40\}$. For current state-of-the-art USDMs the two accuracies are nearly identical, even though clean tokens are the easier target at these corruption rates: the models modify tokens indiscriminately, failing to recognize and preserve already-correct ones.

\input{figures/motivation-figure}

We trace this deficit to the training objective: standard NELBO training provides little incentive for retention. Decomposing the validation NELBO of current USDMs by position type (\Cref{fig:nelbo_four_cases}), we find that at low noise levels an incorrect prediction costs 13--100 nats more than a correct one at a corrupted position, but only 0.09--1.32 nats more at a clean one. Overall, mispredicted clean positions make up 23--31\% of all positions yet only 3.5--5.2\% of the NELBO, versus 90--92\% for mispredicted corrupted ones. We therefore propose \textbf{C}orrect-\textbf{T}oken \textbf{R}etention \textbf{Reg}ularization (\textbf{CTR-Reg}), a simple auxiliary loss term that supplies this missing signal by explicitly training the model to retain tokens left unperturbed by the noising process, and hence already correct.

% Motivated by this analysis, we propose \textbf{C}orrect-\textbf{T}oken \textbf{R}etention \textbf{Reg}ularization (\textbf{CTR-Reg}), a simple auxiliary loss term that explicitly trains the model to retain tokens left unperturbed by the noising process, which are therefore already correct.

Our contributions are:
(i) we identify a \emph{correct-token retention} deficit in USDMs: they revise correct and incorrect tokens indiscriminately, reconstructing both with nearly identical accuracy, and thus collapse in sample diversity under greedy-tail sampling;
(ii) we trace this deficit to the NELBO, which barely penalizes incorrect predictions at clean positions, and propose CTR-Reg, which supplies this missing signal without modifying the sampler;
(iii) CTR-Reg improves clean-token accuracy by 21.5--29.3 points per model and corpus (\Cref{fig:clean-corrupt-bars}; 26.5 on average over six benchmarks) with corrupted-token accuracy virtually unchanged, and cuts the positions changed at the first greedy-tail step to about a third of the \emph{NELBO-only} baseline's, plateauing at 3--11 (Figure~\ref{fig:token-change-vs-greedy-steps}); and
(iv) across DUO, UDLM, and SEDD, this yields substantially higher-quality generations with diversity above the training-data entropy (Figure~\ref{fig:greedy-tail-lowstep}): with 5 greedy-tail steps, perplexity drops from 105.86 to 49.57 (DUO), 93.28 to 44.31 (UDLM), and 194.59 to 87.19 (SEDD), with gains holding across all step budgets, including ultra-low ones (Figure~\ref{fig:step_ablation}), and under MAUVE \citep{pillutla2021mauve} (Figure~\ref{fig:mauve-greedy-tail}).

%% file: figures/motivation-figure.tex
\begin{figure}[t]
\centering
\includegraphics[width=\linewidth]{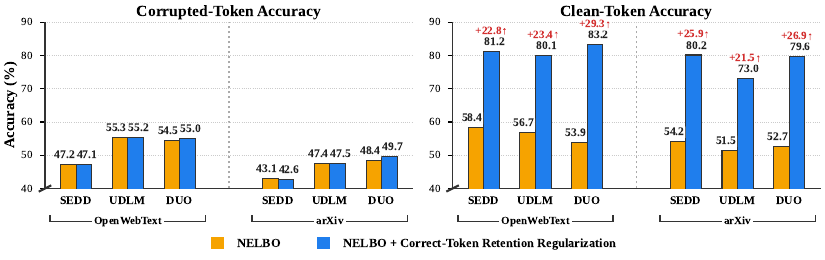}
\caption{\textbf{Effect of CTR-Reg on corrupted- vs.\ clean-token reconstruction accuracy under random-token perturbation.} For SEDD, UDLM, and DUO, we compare base NELBO training (orange) to NELBO + Correct-Token Retention Regularization (CTR-Reg; blue), measuring reconstruction accuracy on corrupted tokens (left) and clean tokens (right), averaged over corruption rates $p \in \{10,20,30,40\}\%$. Each panel shows OpenWebText and arXiv results, separated by a dotted line. CTR-Reg leaves corrupted-token accuracy essentially unchanged while improving clean-token accuracy by 21.5--29.3 percentage points, across all models and both corpora. All models are trained on 20B OpenWebText tokens (169M parameters, sequence length 512); evaluation uses the OpenWebText validation set ($\sim$110M tokens) and Scientific Papers (arXiv) validation set ($\sim$55M tokens).}
\label{fig:clean-corrupt-bars}
\end{figure}
% \vspace{-10pt}

%% file: sections/3_preliminary.tex
\input{figures/token-change-vs-greedy-steps}

\section{Background and Related Work}
\label{sec:background}

\paragraph{Notation.} Let $V:=\{v\in\{0,1\}^K:\sum_{i=1}^K v_i=1\}$ denote the set of one-hot encodings of discrete random variables over $K$ categories (the vocabulary size), and let $\Delta^{K}$ denote the corresponding $K$-simplex. For $v\in\Delta^{K}$, $\mathrm{Cat}(\cdot\,;v)$ denotes the categorical distribution with $\mathbb{P}(u_i=1)=v_i$ for $u\sim\mathrm{Cat}(\cdot\,;v)$, $u\in V$. For vectors $a,b$, let $\langle a,b\rangle$ and $a\odot b$ denote their dot and Hadamard products, and let $\mathbf{1}$ denote the all-ones vector. We write $x\in V$ for a single token and $x^{(1:L)}\in V^{L}$ for a sequence of $L$ tokens, with $x^{(\ell)}$ its $\ell$-th entry; the same convention applies to the noisy latents $z_t$. Let $\pi\in\Delta^{K}$ denote a designated prior distribution.

\subsection{Discrete Diffusion Models}
\label{subsec:discrete_diffusion}

Diffusion models~\citep{pmlr-v37-sohl-dickstein15,ho2020denoising} learn a parameterized reverse process $p_\theta$ that inverts a fixed forward corruption process $q$, which gradually maps clean data $x$ to increasingly noisy latent variables $z_t$ over time $t\in[0,1]$. Discrete diffusion models extend this framework to categorical data such as text~\citep{hoogeboom2021argmax,austin2021structured,lou2024discrete,sahoo2024simple}. Following the interpolating formulation of \citet{sahoo2024simple}, the forward marginal at time $t$ is
\begin{equation}
q_t(z_t \mid x)=\mathrm{Cat}\big(z_t;\,\alpha_t x+(1-\alpha_t)\pi\big),
\label{eq:forward}
\end{equation}
where the noise schedule $\alpha_t\in[0,1]$ decreases monotonically in $t$, with $\alpha_0\approx1$ and $\alpha_1\approx0$, and noise is applied independently to each position of a length-$L$ sequence. With $T$ denoising steps, we follow \citet{schiff2024discreteguidance} and set $t(i)=(i+1)/T$ and $s(i)=i/T$ for $i=0,\ldots,T-1$, and write $q_{s|t}(z_s\mid z_t,x)$ for the forward posterior between consecutive times $s<t$. A denoising network $x_\theta(z_t^{(1:L)},t)\in(\Delta^{K})^{L}$ predicts the clean sequence from the noisy one, and the reverse transition substitutes this prediction into the forward posterior~\citep{sahoo2024simple,schiff2024discreteguidance}:
\begin{equation}
p_\theta\big(z_s^{(1:L)}\mid z_t^{(1:L)}\big)=\prod_{\ell=1}^{L} q_{s|t}\Big(z_s^{(\ell)}\,\Big|\,z_t^{(\ell)},\,x=x_\theta^{(\ell)}\big(z_t^{(1:L)},t\big)\Big).
\label{eq:reverse}
\end{equation}
The reverse process therefore factorizes across positions given $z_t^{(1:L)}$: each position is updated independently of the others at every step. Following \citet{austin2021structured} and \citet{schiff2024discreteguidance}, the network is trained by minimizing the negative evidence lower bound (NELBO), writing $q^{(i)}_{s|t}:=q_{s|t}(z_{s(i)}\mid z_{t(i)},x)$ and $p^{(i)}_\theta:=p_\theta(z_{s(i)}\mid z_{t(i)})$, and $q_{t(T)},p_{t(T)}$ for $q(z_{t(T)}\mid x),p(z_{t(T)})$:
\begin{equation}\small
\mathcal{L}_{\mathrm{vb}} = \mathbb{E}_{q}\Big[\underbrace{-\log p_\theta(x\mid z_{t(0)})}_{\mathcal{L}_{\mathrm{recons}}} + \underbrace{\textstyle\sum_{i=1}^{T} D_{\mathrm{KL}}\!\big(q^{(i)}_{s|t}\,\|\,p^{(i)}_\theta\big)}_{\mathcal{L}_{\mathrm{diffusion}}} + \underbrace{D_{\mathrm{KL}}\!\big(q_{t(T)}\,\|\,p_{t(T)}\big)}_{\mathcal{L}_{\mathrm{prior}}}\Big]
\label{eq:nelbo_discrete}
\end{equation}
whose continuous-time limit $T\to\infty$ gives a tighter bound~\citep{sahoo2024simple,schiff2024discreteguidance}.

\subsection{Uniform-State Diffusion Language Models}
\label{subsec:usdm}

Uniform-state diffusion models (USDMs) set the prior to the uniform distribution over the vocabulary, $\pi=\mathbf{1}/K$, so a corrupted token is replaced by a token drawn uniformly at random~\citep{hoogeboom2021argmax,austin2021structured}. In the discrete-time D3PM framework of \citet{austin2021structured}, the $n$-th forward step applies $Q_n=(1-\beta_n)I+\beta_n\mathbf{1}\mathbf{1}^\top/K$ with $\beta_n\in[0,1]$ to every token; products of such matrices keep the same form, so the marginal after $n$ steps is \eqref{eq:forward} with $\pi=\mathbf{1}/K$ and $\alpha=\prod_{n'\le n}(1-\beta_{n'})$, and the forward posterior has the closed form~\citep{austin2021structured,schiff2024discreteguidance}
\begin{equation}\small
q_{s|t}(z_s\mid z_t,x)=\mathrm{Cat}\!\left(z_s;\,\frac{K\alpha_t\,z_t\odot x+(\alpha_{t|s}-\alpha_t)\,z_t+(\alpha_s-\alpha_t)\,x+(1-\alpha_{t|s})(1-\alpha_s)\,\mathbf{1}/K}{K\alpha_t\langle z_t,x\rangle+1-\alpha_t}\right),
\label{eq:usdm_posterior}
\end{equation}
with $\alpha_{t|s}=\alpha_t/\alpha_s$. Unlike absorbing-state diffusion, where a decoded token is kept deterministically~\citep{sahoo2024simple}, \eqref{eq:usdm_posterior} assigns non-zero probability both to keeping $z_t$ and to moving to any other token, so every position can be revised at every step; this is the source of the self-correction ability of USDMs~\citep{sahoo2025the,deschenaux2026the}.

\paragraph{Continuous-time objective (NELBO).} Current state-of-the-art USDMs adopt different
parameterizations, but all optimize a bound on the negative log-likelihood (NELBO), and up to
parameterization these bounds coincide with a single continuous-time objective, which we state here
for $\pi=\mathbf{1}/K$. UDLM~\citep{schiff2024discreteguidance} reaches it from the variational perspective, by
bounding a discrete-time chain and taking that bound to the continuous-time limit, as
MDLM~\citep{sahoo2024simple} does for the absorbing state. DUO~\citep{sahoo2025the} trains with the
same bound but rewrites its per-token term to avoid materializing an intermediate $K$-dimensional
quantity, and computes certain expectations analytically via Rao-Blackwellization, lowering memory,
training time and gradient variance. SEDD~\citep{lou2024discrete} instead formulates the process
directly as a continuous-time Markov chain~\citep{campbell2022continuous} and predicts ratios of
marginals (the concrete score~\citep{meng2022concrete}) rather than the clean data, training
with the diffusion-weighted denoising score entropy; \citet{schiff2024discreteguidance} show that
under a mean parameterization of the concrete score this objective is equivalent to the same bound,
and report that CTMC-based models underperform MDLM and UDLM.

Concretely, the bound is the $T\to\infty$ limit of \eqref{eq:nelbo_discrete}:
$\mathcal{L}_{\mathrm{prior}}$ and $\mathcal{L}_{\mathrm{recons}}$ vanish, and the diffusion term
converges to
\begin{equation}
\mathcal{L}^{\infty}=\mathbb{E}_{t\sim\mathcal{U}[0,1]}\,\mathbb{E}_{z_t\sim q_t(\cdot\mid x)}\textstyle\sum_{\ell=1}^{L} f\big(z_t^{(\ell)},x_\theta^{(\ell)}(z_t^{(1:L)},t),\alpha_t;x^{(\ell)}\big),
\label{eq:udlm_nelbo}
\end{equation}
\begin{equation}\small
f\big(z_t,x_\theta(z_t,t),\alpha_t;x\big)=\frac{\alpha_t'}{K\alpha_t}\Bigg[\frac{K}{\bar{x}_i}-\frac{K}{(\bar{x}_\theta)_i}-\sum_{j\neq i}\frac{\bar{x}_j}{\bar{x}_i}\log\frac{(\bar{x}_\theta)_i\,\bar{x}_j}{(\bar{x}_\theta)_j\,\bar{x}_i}\Bigg],
\label{eq:udlm_f}
\end{equation}
where $i$ indexes the non-zero entry of $z_t$, $\bar{x}=K\alpha_t x+(1-\alpha_t)\mathbf{1}$,
$\bar{x}_\theta=K\alpha_t x_\theta(z_t,t)+(1-\alpha_t)\mathbf{1}$, and $\alpha_t'=d\alpha_t/dt$.
The explicit per-token objectives of DUO and SEDD, and their equivalence to this bound, are given
in \Cref{appendix:usdm-objectives}. \eqref{eq:udlm_nelbo} is the objective to which CTR-Reg is
added in \Cref{sec:method-ccreg}.

%% file: figures/token-change-vs-greedy-steps.tex
\begin{figure}[t]
\centering
\begin{minipage}[t]{0.49\linewidth}
    \centering
    \includegraphics[width=0.9\linewidth]{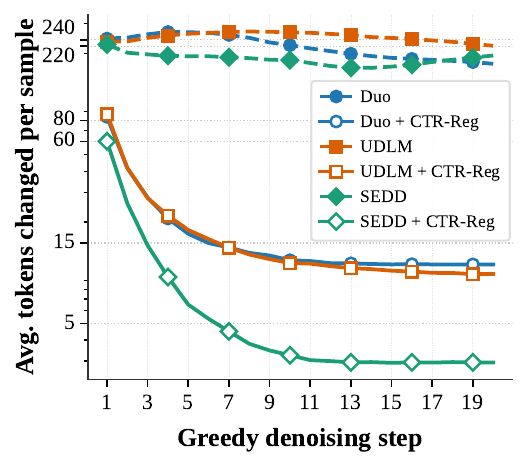}
\caption{\textbf{Token changes per greedy-tail step.} Mean positions (of 512) changed at each of $g=20$ greedy-tail steps after 500 ancestral steps (512 samples; broken $y$-axis) for DUO, UDLM, and SEDD, NELBO-only (dashed) vs.\ CTR-Reg (solid). NELBO-only models never converge, still revising 173--270 positions at step 20 and driving the diversity collapse in \Cref{fig:greedy-tail-lowstep}; CTR-Reg plateaus at 3--11 positions within ${\sim}10$ steps.}\label{fig:token-change-vs-greedy-steps}
\end{minipage}
\hfill
\begin{minipage}[t]{0.49\linewidth}
    \centering
    \includegraphics[width=0.9\linewidth]{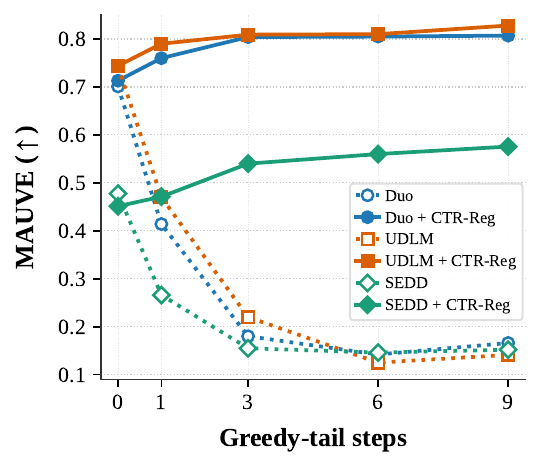}
\caption{\textbf{MAUVE ($\uparrow$) vs.\ greedy-tail steps.} DUO, UDLM, and SEDD with (solid) and without (dotted) CTR-Reg; $g \in \{0,1,3,6,9\}$ greedy-tail steps follow 500 ancestral steps; 5{,}000 samples vs.\ 5{,}000 OWT sequences. NELBO-only models degrade sharply as $g$ grows, consistent with the diversity collapse in \Cref{fig:greedy-tail-lowstep}; CTR-Reg models instead improve with $g$.}
% \caption{MAUVE score as a function of Greedy-Tail steps for DUO, UDLM, and SEDD, with (solid) and without (dotted) CTR-Reg. NELBO-only models degrade sharply as greedy-tail steps increase, consistent with the diversity collapse above; CTR-Reg instead improves MAUVE across all three models.}
    \label{fig:mauve-greedy-tail}
\end{minipage}
\end{figure}

%% file: sections/4_methods.tex
\input{figures/main_figure}

\section{Methodology}
\label{sec:method}

This section develops both the diagnosis and the remedy. We first show that current USDMs revise
positions without regard to whether the token already there is correct, that this deficit
surfaces as a collapse in sample diversity under greedy-tail decoding, and that the NELBO training
objective barely penalizes it (\Cref{sec:token_change_mechanism_method}). We then propose
Correct-Token Retention Regularization (CTR-Reg), a single auxiliary term that targets
correct-token retention directly (\Cref{sec:method-ccreg}).

\subsection{Current USDMs Lack Correct-Token Retention}
\label{sec:token_change_mechanism_method}

\paragraph{Multi-step greedy-tail sampling.}\label{multi-step-greedy-step-sampling}
We study three state-of-the-art USDMs, DUO~\citep{sahoo2025the}, UDLM~\citep{schiff2024discreteguidance}, and SEDD~\citep{lou2024discrete}, under a multi-step extension of the \emph{Greedy-Tail Sampler} of \citet{sahoo2025the}. Given a total step budget $T=n+g$, the sampler takes $n$ ancestral steps (\eqref{eq:reverse} with $(s,t)=(s(i),t(i))$) followed by $g$ greedy-tail steps, each a single forward pass at the terminal diffusion time $\delta$ followed by a per-position $\arg\max$, with no stochastic sampling:
\begin{equation}
z^{(\ell)}\;\leftarrow\;\mathrm{onehot}\Big(\arg\max_{j}\big[x_\theta^{(\ell)}\big(z^{(1:L)},\delta\big)\big]_j\Big)\qquad\forall\,\ell\in\{1,\ldots,L\}.
\label{eq:greedy_tail}
\end{equation}
Setting $g=0$ recovers pure ancestral sampling and $g=1$ the sampler of \citet{sahoo2025the}, whose final greedy step lowers sample entropy much as nucleus sampling does~\citep{holtzman2020curious}; fixing $T$ while sweeping $g$ keeps comparisons compute-matched. Under this sampler, all three USDMs suffer a substantial collapse in sample diversity, measured by entropy (\Cref{fig:greedy-tail-lowstep}), even the official DUO checkpoint trained on 26$\times$ more data, whose entropy falls from 5.54 to 2.14 over 20 greedy-tail steps and stays below the real-text level at its sequence length of 1024 (5.40) for every $g \ge 1$ (\Cref{sec:appendix-scale-does-not-fix}).

\paragraph{Where the collapse comes from.}
To identify the mechanism, we generate 512 sequences of length 512 per USDM with $n=500$ ancestral and $g=20$ greedy-tail steps, and track the number of positions whose decoded token changes at each greedy-tail step, averaged over samples (\Cref{fig:token-change-vs-greedy-steps}). For all three NELBO-only models, this number starts near half the sequence and shows no sign of convergence: even at the twentieth step, 173--270 positions still change. Because all positions are updated independently and simultaneously (\eqref{eq:reverse}), each step rewrites a large fraction of the sequence, and even locally plausible edits, made by the hundreds at once, compound into globally inconsistent sequences~\citep{gu2018nonautoregressive}, producing the low-diversity outputs in \Cref{fig:greedy-tail-lowstep,fig:step_ablation}. This persistent churn stems from \textbf{the models' inability to recognize and retain tokens that are already correct}: without such a signal, a greedy step has no basis for leaving a position unchanged.

\paragraph{Ancestral sampling masks the correct-token retention deficit.}
The retention seen under ancestral sampling comes from the noise schedule, not from the model. Under \eqref{eq:reverse}, a position that disagrees with the model's prediction adopts it with probability only $\tfrac{\alpha_s-\alpha_t}{1-\alpha_t}$ (\Cref{appendix:ancestral-suppress-token-change}), which is $O(1/T)$ for any smooth schedule (\Cref{app:scheduler_bound}): decoded tokens persist largely because the sampler rarely acts on the model's prediction, not because the model would keep them. This masking is weakest near the terminal region, where the greedy tail operates, yet even there at most roughly one in five disagreeing positions adopts the prediction (under the log-linear schedule of DUO and UDLM, for which $\alpha_t=1-t$; \Cref{tab:appendix-transition}). Absent genuine correct-token retention, which positions get revised is thus decided by sampling stochasticity rather than by the model; removing this bound, as the greedy tail does, exposes the deficit directly.

\paragraph{The NELBO barely rewards retention.}
To quantify how much the training objective penalizes this deficit, we split validation positions into clean ($z_t=x$) and corrupted ($z_t\neq x$), and by whether the model's prediction $\hat{x}_\theta=\mathrm{onehot}(\arg\max_j[x_\theta]_j)$ equals the clean token, and average the per-token NELBO of NELBO-only models within each group and noise level (\Cref{fig:nelbo_four_cases}). At corrupted positions, an incorrect prediction is heavily penalized: for $1-\alpha_t<0.4$ it costs 13--100 nats more than a correct one. At clean positions, the same error costs only 0.09--1.32 nats more; overall, mispredicted clean positions make up 23--31\% of all positions yet only 3.5--5.2\% of the NELBO, versus 90--92\% for mispredicted corrupted ones (\Cref{app:nelbo_decomposition}). This holds for all three models on all six corpora (\Cref{fig:nelbo_four_cases,fig:nelbo_four_cases_appendix}). The NELBO thus provides almost no signal for retaining correct tokens, consistent with their near-identical clean- and corrupted-token accuracies (\Cref{fig:clean-corrupt-bars}).

\input{figures/NELBO_Loss}

\subsection{Correct-Token Retention Regularization (CTR-Reg)}
\label{sec:method-ccreg}

Since the NELBO provides almost no signal at clean positions (\Cref{fig:nelbo_four_cases}), we add
one explicitly. CTR-Reg applies only at positions left unperturbed by the forward process
($z_t=x$), where the displayed token is already correct, and rewards the model for predicting that
token with high probability:
\begin{equation}
f_{\mathrm{CTR}}\big(z_t,x_\theta(z_t,t);x\big)
\;=\;
\mathbb{I}\big[z_t=x\big]\;
\Big(-\log\big\langle x_\theta(z_t,t),\,x\big\rangle\Big),
\label{eq:ccreg}
\end{equation}
where $\langle x_\theta,x\rangle=(x_\theta)_m$ is the probability the denoiser assigns to the
clean token and $m$ indexes its non-zero entry; the indicator restricts the term to correct
tokens and directly reinforces their retention. The full objective adds this term
to the model's NELBO $\mathcal{L}^{\infty}$ (\eqref{eq:udlm_nelbo}, whose per-token form
differs across USDMs) through a single scalar $\lambda\ge0$:
\begin{equation}
\mathcal{L}
\;=\;
\mathcal{L}^{\infty}
\;+\;
\lambda\,\mathbb{E}_{t\sim\mathcal{U}[0,1]}\,\mathbb{E}_{z_t\sim q_t(\cdot\mid x)}
\textstyle\sum_{\ell=1}^{L}
f_{\mathrm{CTR}}\big(z_t^{(\ell)},x_\theta^{(\ell)}(z_t^{(1:L)},t);x^{(\ell)}\big).
\label{eq:ccreg_full}
\end{equation}

\paragraph{CTR-Reg preserves the likelihood bound.}
Since $\langle x_\theta,x\rangle\in(0,1]$, we have $-\log\langle x_\theta,x\rangle\ge0$; with
$\lambda\ge0$ and the indicator non-negative, the CTR-Reg term is non-negative pointwise. Hence
\begin{equation}
\mathcal{L}\;=\;\mathcal{L}^{\infty}+\lambda\,\mathbb{E}\big[f_{\mathrm{CTR}}\big]
\;\ge\;\mathcal{L}^{\infty}\;\ge\;-\log p_\theta(x).
\label{eq:bound}
\end{equation}
Adding a non-negative quantity to an upper bound leaves an upper bound: training with CTR-Reg still
minimizes a valid bound on the negative log-likelihood, merely a looser one. Empirically, the likelihood cost is small: validation perplexity changes by $+1.5\%$, $+1.2\%$, and $-2.4\%$ on average for DUO, UDLM, and SEDD (\Cref{app:val_ppl}), even though clean-token accuracy rises by 26.5 points (\Cref{sec:exp-retention}), further indicating that the likelihood is largely insensitive to retention.

% Empirically it loosens little: validation perplexity changes by $+1.5\%$,
% $+1.2\%$, and $-2.4\%$ on average for DUO, UDLM, and SEDD, even though clean-token accuracy rises by
% 26.5 points (\Cref{app:val_ppl}), further indicating that the likelihood is largely insensitive to
% retention.

\paragraph{Cost and hyperparameter.}
CTR-Reg acts only at training time and reuses the same forward pass as the NELBO, so it adds no
computational overhead and leaves the sampler unchanged. We use $\lambda=0.02$ for every USDM, and
ablate it in \Cref{sec:ablation_study}.

\subsection{Applying CTR-Reg to Score-Parameterized Models}
\label{sec:method-sedd}

UDLM and DUO predict the clean data directly, so \eqref{eq:ccreg} applies verbatim. SEDD, in
contrast, outputs the log concrete score $\log s_\theta(z_t,t)$, whose $j$-th entry estimates
$\log\big(q_t(j)/q_t(i)\big)$ relative to the displayed token $i$. These ratios determine the reverse
transition rates from $i$ to $j$~\citep{lou2024discrete}; because the uniform forward rate is
identical across all $j\neq i$, $s_\theta(z_t,t)_j$ directly measures how strongly the model favors
moving from $i$ to $j$. The displayed token has a fixed self-score $s_\theta(z_t,t)_i=1$, so the
preference for retaining $i$ is encoded jointly by the off-diagonal scores and increases as they
decrease. Consequently, \eqref{eq:ccreg} does not apply directly to SEDD. We instead formulate
the penalty over the off-diagonal scores, penalizing scores above SEDD's training target with a
smooth hinge:
\begin{equation}
f_{\mathrm{CTR}}^{\mathrm{score}}
\;=\;
\mathbb{I}\big[z_t=x\big]\,
\frac{1}{K-1}\sum_{j\neq i}
\mathrm{softplus}\!\left(\log\frac{s_\theta(z_t,t)_j}{\rho_t}\right),
\qquad
\rho_t=\frac{1-\alpha_t}{K\alpha_t+1-\alpha_t}.
\label{eq:ccreg_score}
\end{equation}
Here, $\rho_t=q_t(j\mid x)/q_t(m\mid x)$ is the target value of every off-diagonal score at an
unperturbed position ($i=m$) (\Cref{app:sedd-target}). We use a softplus hinge to obtain a non-negative penalty that vanishes
smoothly below the target; adding $\lambda\,\mathbb{E}\big[f_{\mathrm{CTR}}^{\mathrm{score}}\big]$ to
SEDD's objective therefore preserves the bound argument of \Cref{sec:method-ccreg}. Dividing by
$\rho_t$ places the hinge at the training target: without this normalization,
$\mathrm{softplus}(\log s)$ becomes negligible as soon as $s<1$, which can be substantially above
$\rho_t$. With it, the penalty pushes scores above $\rho_t$ downward and fades below it, where SEDD's own
objective already holds them (\Cref{app:sedd-penalty}).

%% file: figures/main_figure.tex
\begin{figure}[t]
\centering
\includegraphics[width=\linewidth]{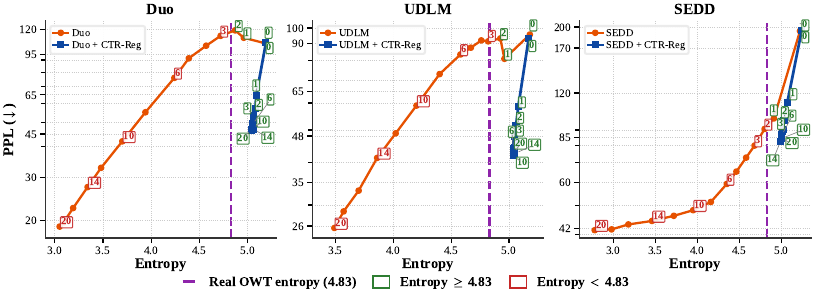}
\caption{\textbf{PPL--entropy frontier vs.\ greedy-tail steps.} Unconditional generation with DUO, UDLM, and SEDD (one per panel) at a fixed budget of 512 steps, sweeping the number of terminal greedy-tail steps $g$ from 0 (fully ancestral) to 20 (annotated at each point); PPL is measured with GPT2-Large over 1024 samples. The dashed line marks real OpenWebText entropy (4.83), below which lower PPL reflects degenerate repetition rather than genuine quality. Greedy-tail steps collapse diversity in NELBO-only models, which fail to preserve correct tokens, whereas CTR-Reg shifts the frontier toward lower PPL at higher entropy and reaches high-quality generations within a few steps.}% \caption{\textbf{PPL–entropy frontier as a function of greedy-tail step count.} We evaluate three state-of-the-art uniform-state diffusion language models (DUO, UDLM, and SEDD), one per panel, using standard ancestral sampling with a greedy-tail extension: some number of terminal denoising steps (argmax, non-stochastic) are appended to the end of ancestral sampling, with the total step budget fixed at 512 (ancestral steps + greedy-tail steps = 512). The greedy-tail step count is swept from $0$ (fully ancestral) to $20$ and annotated at each point. For each model, we plot perplexity (quality) against entropy (diversity) of unconditional text generation, comparing the NELBO-only baseline against the model trained with NELBO plus Correct-Token Retention Regularization (CTR-Reg). Perplexity (PPL) is computed with GPT2-Large over 1024 generated samples per model. The dashed vertical line marks the entropy of real OpenWebText, the minimum diversity a point must reach for its PPL to reflect genuine quality rather than degenerate, repetitive text. This figure demonstrates that (i) CTR-Reg shifts the CTR-Reg-trained models' curves toward the lower right, reaching substantially lower PPL at higher entropy than the baseline; (ii) greedy-tail steps in NELBO-only trained models always reduce diversity (entropy), since without sampling or randomness these models are never confident enough to keep the correct tokens and instead converge to low-diversity output; and (iii) CTR-Reg-trained models perform well after only a few greedy-tail steps, converging to confident, high-quality generations.}
\label{fig:greedy-tail-lowstep}
\end{figure}

% \caption{\textbf{PPL–Entropy Frontier as a function of greedy-tail step count.} 

% In this figure We used Standard Ancestral Sampling with 512 step to evaluate the state of the art uniform-state diffusion langue models (DUO, UDLM and SEDD) in three panel. 

% Each panel the number of terminal denoising steps replaced by a greedy-tail (argmax, non-stochastic) update, from $0$ (fully ancestral) to $20$, Numbers annotate the greedy-tail step count at each point (Ancestral Step + greedy-tail step = 512); Perpelexity (PPL) evaluated using GPT2-Large using 1024 generated samples.

% The dashed vertical line marks the entropy of real OpenWebText, the minimum diversity a point must reach for its PPL to reflect genuine quality rather than degenerate, repetitive text;

% for each panel of DUO, UDLM, and SEDD we compare the base NELBO training and NELBO plus Clean-Token Confidence Regualrization trained model.

% Across all three models, (i) CTR-Reg shifts the trade-off curve toward the lower-right, reaching substantially lower PPL at higher entropy than the baseline, (ii) greedy-tail steps in NEBLO-only trained models alway reduce the diversity (entropy)

%% file: figures/NELBO_Loss.tex
\begin{figure}[t]
\centering
\includegraphics[width=\linewidth]{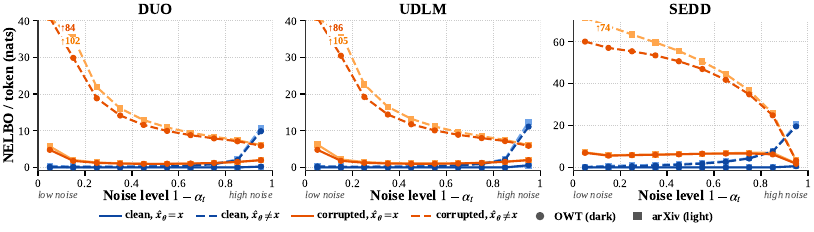}
\caption{\textbf{The NELBO barely penalizes failures of retention.} Mean validation NELBO per token of NELBO-only models on OpenWebText (dark) and arXiv (light) versus the noise level $1-\alpha_t$ (fraction of corrupted tokens), for clean (blue) and corrupted (orange) positions where the argmax prediction $\hat{x}_\theta$ equals the clean token $x$ (solid) or not (dashed). The solid–dashed gap is the likelihood penalty for an incorrect prediction: large at corrupted positions, near zero at clean positions except at the highest noise. Magnitudes are comparable within, not across, panels.}
\label{fig:nelbo_four_cases}\label{fig:nelbo_four_cases}
\end{figure}
% \vspace{-10pt}

%% file: sections/5_experiments.tex
\section{Experiments}
\label{sec:experiments}
We evaluate the effectiveness of CTR-Reg on three USDMs, namely DUO~\citep{sahoo2025the}, UDLM~\citep{schiff2024discreteguidance}, and SEDD with uniform noise~\citep{lou2024discrete}, each trained with the standard NELBO objective (\emph{NELBO-only}) and with NELBO~+~CTR-Reg. We address three questions: (i)~does CTR-Reg improve correct-token retention without impairing the revision of incorrect tokens (\Cref{sec:exp-retention}); (ii)~does better retention translate into better unconditional generation (\Cref{sec:ccreg_generation});    and (iii)~how does CTR-Reg prevent the diversity collapse under greedy-tail decoding (\Cref{sec:token_change_mechanism})?

% \paragraph{Implementation Details.}\label{implementation_detail}
% %We train DUO~\citep{sahoo2025the}, UDLM~\citep{schiff2024discreteguidance}, and SEDD~\citep{lou2024discrete} using either the standard NELBO objective or NELBO~+~CTR-Reg. 
% We train the three models with the two objectives above. All models use the same 169M-parameter transformer backbone and are trained on 20B tokens from OpenWebText~\citep{Gokaslan2019OpenWeb} with a sequence length of $L=512$. The optimizer, batch size, and learning-rate schedule follow the respective model's official implementation and paper. Unless otherwise stated, sampling uses a budget of 512 denoising steps. For each unconditional text-generation configuration, we generate 1{,}024 unconditional samples of length $L$ and report generative perplexity (Gen PPL) under GPT-2 Large~\citep{radford2019language} and the average per-sequence unigram entropy. For MAUVE~\citep{pillutla2021mauve}, we generate 5{,}000 unconditional samples and compare them with 5{,}000 human-written OpenWebText sequences.

\paragraph{Implementation Details.}\label{implementation_detail}
We train the three models with the two objectives above. All models use the same 169M-parameter transformer and are trained from scratch on OpenWebText~\citep{Gokaslan2019OpenWeb} ($L=512$, 77K steps, batch size 512, ${\approx}$20B tokens), with the optimizer and remaining settings following each model's official implementation: log-linear noise for DUO and UDLM, geometric noise for SEDD, and DUO's Gaussian curriculum~\citep{sahoo2025the} over the first half of training. CTR-Reg uses a constant $\lambda=0.02$ for all models; NELBO-only baselines use identical settings with $\lambda=0$. Unless stated otherwise, we sample with 512 steps and report Gen PPL under GPT-2 Large~\citep{radford2019language} and mean unigram entropy over 1,024 samples, and MAUVE~\citep{pillutla2021mauve} between 5,000 generated and 5,000 OpenWebText sequences.

\subsection{CTR-Reg Improves Retention Without Hurting Revision}
\label{sec:exp-retention}

\paragraph{Protocol.}
We evaluate retention using a controlled random-token corruption experiment. We split each evaluation corpus into sequences of $L=512$ tokens and randomly replace $p\%$ of the positions with tokens drawn uniformly from the vocabulary. Each model then reconstructs the corrupted sequence in a single forward pass, taking the argmax of $x_\theta^{(\ell)}$ at each position, with the time input set to the corruption level corresponding to $p$. We report accuracy, i.e., the fraction of positions whose prediction matches the original token, separately on the corrupted positions (\emph{Corrupted-Token Accuracy}, measuring revision) and the remaining $(100-p)\%$ of positions (\emph{Clean-Token Accuracy}, measuring retention), averaged over $p\in\{10,20,30,40\}\%$. For a model that recognizes and preserves correct tokens, clean positions are the easier target: the correct prediction is simply the observed token.

\paragraph{Results.}
\Cref{fig:clean-corrupt-bars} reports both accuracies on the OpenWebText validation split (last 100K documents) and arXiv~\citep{cohan-etal-2018-discourse}; \Cref{tab:corruption_exp_generalization} (\Cref{sec:appendix-generalization-random_corruption}) extends the evaluation to six corpora spanning web, news, scientific, biomedical, and encyclopedic text by adding LM1B~\citep{chelba2014billionwordbenchmarkmeasuring}, AG News~\citep{zhang2016characterlevelconvolutionalnetworkstext}, PubMed~\citep{cohan-etal-2018-discourse}, and WikiText-103~\citep{merity2017pointer}. NELBO-only models achieve nearly identical accuracy on clean and corrupted positions despite clean positions being the easier target, indicating that they revise tokens indiscriminately. This is not an artifact of undertraining: the official DUO checkpoint, trained on 26$\times$ more data, raises both accuracies by 5--6 points but leaves the clean-minus-corrupted gap at 1.1 points (vs.\ 2.0 at 20B), and CTR-Reg at 20B exceeds the checkpoint's clean-token accuracy by 24.7 points (77.4 vs.\ 52.7; \Cref{sec:appendix-scale-does-not-fix}). CTR-Reg improves Clean-Token Accuracy for every model and corpus, by 26.5 points on average, while changing Corrupted-Token Accuracy by only 0.7 points per cell on average (\Cref{sec:appendix-generalization-random_corruption}). CTR-Reg thus substantially improves retention without compromising revision.
% this is not an artifact of undertraining, as the official DUO checkpoint, trained on 26$\times$ more data, shows the same near-zero gap (\Cref{sec:appendix-scale-does-not-fix}). 

% CTR-Reg improves Clean-Token Accuracy for every model and corpus, by 26.5 percentage points on average, while changing Corrupted-Token Accuracy by less than 1.5 percentage points in every cell, a pattern that holds at every corruption rate, not only in aggregate (\Cref{sec:appendix-generalization-random_corruption}). CTR-Reg thus substantially improves retention without compromising revision.

% \input{figures/Step-Ablation}

\subsection{Better Retention Yields Better Generations}
\label{sec:ccreg_generation}
We next test whether the single-step retention gains from CTR-Reg translate into better full generations. We sample with the multi-step greedy-tail sampler (\Cref{multi-step-greedy-step-sampling}), which appends a few deterministic denoising steps (a forward pass followed by a per-position argmax, with no stochastic sampling) after standard ancestral sampling. Because no sampling randomness is involved, whether a position is kept or rewritten depends only on the model's prediction. A model that cannot recognize already-correct tokens therefore has no basis for leaving them untouched, so greedy-tail decoding directly exposes the retention deficit identified in \Cref{sec:token_change_mechanism_method}. We evaluate the quality and diversity of the resulting unconditionally generated samples to assess the effectiveness of CTR-Reg.

\paragraph{Generative Perplexity vs.\ Entropy.} \Cref{fig:greedy-tail-lowstep} plots Gen PPL against entropy as the number of greedy-tail steps grows from $g=0$ to $g=20$ at $T=512$ (exact values in \Cref{sec:appendix-greedy-tail}). For NELBO-only models, greedy-tail steps lower Gen PPL but push entropy below that of real text, so the lower Gen PPL reflects a collapse toward repetitive output rather than better text, consistent with the retention deficit measured in \Cref{sec:exp-retention}. CTR-Reg instead moves all three models toward lower Gen PPL at higher entropy, reaching substantially lower Gen PPL while keeping entropy at or above that of real OpenWebText. A few steps suffice: from $g = 0$ to $g = 5$, CTR-Reg lowers Gen PPL from 105.86 to 49.57 for DUO, from 93.28 to 44.31 for UDLM, and from 194.59 to 87.19 for SEDD. The benefit already appears under DUO's default sampler ($g=1$), with no change to sampling: CTR-Reg reduces DUO's Gen PPL from 110.88 (NELBO-only) to 64.67 while raising entropy from 4.96 to 5.09. These gains hold across step budgets from 16 to 1024, including the ultra-low-step regime (\Cref{sec:appendix-step-budget}).

% These gains hold across step budgets from 16 to 1024, including the ultra-low-step regime (\Cref{sec:appendix-step-budget}). 
% Even DUO's default sampler ($g = 1$) benefits without modification: CTR-Reg lowers its Gen PPL from 110.9 (NELBO-only) to 64.7 at $T = 512$ and from 147.5 to 80.5 at $T = 16$, while raising entropy (\Cref{sec:appendix-step-budget}).

% A few steps suffice: from $g=0$ to $g=5$, CTR-Reg lowers Gen PPL from 105.86 to 49.57 for DUO, from 93.28 to 44.31 for UDLM, and from 194.59 to 87.19 for SEDD. These gains hold across step budgets from 16 to 1024, including the ultra-low-step regime (\Cref{sec:appendix-step-budget}). 

\paragraph{MAUVE.}\label{sec:mauve_greedy_tail} To confirm the Gen PPL and entropy trends, we additionally evaluate MAUVE~\citep{pillutla2021mauve}. Whereas Gen PPL and entropy measure quality and diversity separately, MAUVE captures both jointly by comparing the distributions of generated and human-written text. \Cref{fig:mauve-greedy-tail} reports MAUVE as a function of $g$ (exact values in \Cref{sec:appendix-mauve-greedy-tail}). At $g=0$, NELBO-only and CTR-Reg models score comparably, so CTR-Reg does not degrade standard ancestral sampling. As $g$ grows, NELBO-only models collapse: within 6--9 steps, MAUVE falls from $\sim$0.70--0.75 to as low as 0.13--0.17 for DUO and UDLM, and from 0.47 to 0.15 for SEDD. CTR-Reg models instead improve with additional steps, reaching 0.81 (DUO), 0.83 (UDLM), and 0.58 (SEDD). 

% \paragraph{Step-Budget Analysis.} \Cref{fig:step_ablation} shows that this pattern holds across the full range of step budgets from 16 to 1024 (exact values in \Cref{sec:appendix-step-ablation}). Under each model's default sampler (for DUO, with one trailing greedy step), CTR-Reg matches or outperforms the NELBO-only baseline at every budget, including the ultra-low-step regime. Adding greedy-tail steps widens the gap: CTR-Reg substantially lowers Gen PPL while preserving diversity, whereas the apparent gains of NELBO-only models come at the cost of diversity, consistent with the collapse in \Cref{fig:greedy-tail-lowstep}. The benefit of improved retention therefore holds across sampling budgets.

\subsection{How CTR-Reg Resolves the Diversity Collapse}
\label{sec:token_change_mechanism}

We repeat the token-change analysis of \Cref{sec:token_change_mechanism_method} with the same setup (\Cref{fig:token-change-vs-greedy-steps}). There, NELBO-only models still revise 173--270 of the 512 positions at the twentieth greedy-tail step, and these independent, simultaneous edits compound into the entropy and MAUVE collapse of \Cref{sec:ccreg_generation}. CTR-Reg removes this churn: for all three models, the number of changed positions at the first greedy-tail step is already about one third of the NELBO-only count and keeps decreasing to a stable plateau of 3--11 positions, so greedy decoding converges to a near-fixed point instead of repeatedly rewriting the sequence. The corruption experiment (\Cref{sec:exp-retention}), where ground truth is available, confirms that this selectivity is well placed: CTR-Reg models retain clean tokens while still correcting corrupted ones as often as NELBO-only models. Bounded, targeted revision thus preserves sequence structure, whereas indiscriminate revision destroys it, which accounts for the Gen PPL, entropy, and MAUVE gaps in \Cref{sec:ccreg_generation}.

\definecolor{purplereg}{HTML}{7A4EAB}  % Deep purple for primary text accents
\definecolor{purplerow}{HTML}{F3EEF9}  % Soft purple tint for CTR-Reg rows

%% file: sections/6_ablation_study.tex
\input{figures/lambda_ablation_ppl_entropy}

% \section{Ablation Study}
% \label{sec:ablation_study}
\subsection{Regularization Strength Analysis}
\label{sec:ablation_study}
We vary $\lambda\in\{0,0.02,0.1\}$ on DUO and UDLM ($\lambda=0$: NELBO-only; main experiments use $\lambda=0.02$). Raising $\lambda$ from 0.02 to 0.1 further improves Clean-Token Accuracy (DUO: 77.4\%$\to$87.2\%; UDLM: 72.8\%$\to$83.6\%), while Corrupted-Token Accuracy stays largely unchanged (\Cref{tab:appendix-lambda-corruption}; per-rate results in \Cref{sec:appendix-lambda-percorruption}). To verify that CTR-Reg does not simply learn to copy its input, we also report the copy rate: the percentage of corrupted positions where the model keeps the random replacement token instead of recovering the original one. It is near zero for NELBO-only models (0.003--0.007\%) and grows with $\lambda$ (DUO: 0.05\%/0.22\%; UDLM: 0.03\%/0.11\% at $\lambda=0.02/0.1$), yet stays below 0.25\% throughout: CTR-Reg retains tokens selectively rather than copying them. Stronger retention, however, does not yield better generation (\Cref{fig:lambda-greedy-tail}): $\lambda=0.1$ matches smaller $\lambda$ under ancestral sampling ($g=0$) and avoids the diversity collapse under greedy-tail decoding, but gains less from additional greedy-tail steps than $\lambda=0.02$. Consistent with its higher copy rate, excessive retention can thus hinder the revision of earlier sampling mistakes, whereas $\lambda=0.02$ better balances retaining correct tokens and revising uncertain ones, yielding stronger generation for both models.
% Our main experiments use $\lambda=0.02$. We ablate $\lambda\in\{0,0.02,0.1\}$ on DUO and UDLM, where $\lambda=0$ corresponds to the NELBO-only baseline. As shown in \Cref{tab:appendix-lambda-corruption}, increasing $\lambda$ from 0.02 to 0.1 further improves Clean-Token Accuracy (DUO: 77.4\%$\to$87.2\%; UDLM: 72.8\%$\to$83.6\%), while Corrupted-Token Accuracy remains largely unchanged (per-rate results in \Cref{sec:appendix-lambda-percorruption}). However, stronger retention does not translate into better generation. As shown in \Cref{fig:lambda-greedy-tail}, $\lambda=0.1$ performs comparably to smaller $\lambda$ under pure ancestral sampling ($g=0$) and avoids the diversity collapse under greedy-tail decoding, but gains less from the additional greedy-tail steps than $\lambda=0.02$. This suggests that excessive retention can hinder the revision of earlier sampling mistakes. In contrast, $\lambda=0.02$ better balances retaining correct tokens and revising uncertain ones, yielding stronger generation performance for both models.

%% file: figures/lambda_ablation_ppl_entropy.tex
\makeatletter
\@ifundefined{bluereg}{\definecolor{bluereg}{HTML}{1F5FA6}}{}
\@ifundefined{bluerow}{\definecolor{bluerow}{HTML}{EAF1FB}}{}
\makeatother
\newcolumntype{G}{>{\columncolor{purplerow}\color{purplereg}}c} % lambda = 0.02
\newcolumntype{H}{>{\columncolor{bluerow}\color{bluereg}}c}     % lambda = 0.1

\begin{table}[t]
\centering
\begin{minipage}[t]{0.49\textwidth}
\vspace{0pt}\centering
\caption{\textbf{Effect of CTR-Reg coefficient $\lambda$ on the corruption probe.} Corrupted- and clean-token accuracy and copy rate (corrupted positions that keep the random replacement token instead of recovering the original), in \%, avg.\ over $p \in \{10,20,30,40\}\%$, for DUO and UDLM at $\lambda=0$ (NELBO-only), $0.02$ (\textcolor{purplereg}{\textbf{our default}}), and $0.1$ (\textcolor{bluereg}{\textbf{blue}}). \textbf{Avg}: mean over six corpora; \textbf{bold}: best accuracy across $\lambda$ per model and corpus. Clean-token accuracy rises monotonically with $\lambda$; corrupted-token accuracy does not; copy rate stays below $0.25\%$.}\label{tab:appendix-lambda-corruption}
\renewcommand{\arraystretch}{1.0}\setlength{\tabcolsep}{3pt}
\resizebox{\linewidth}{!}{%
\begin{tabular}{ll cGH cGH cGH}
\specialrule{1.1pt}{2pt}{2pt}
& & \multicolumn{3}{c}{\textbf{Corrupted-Token Acc.} $\uparrow$} & \multicolumn{3}{c}{\textbf{Clean-Token Acc.} $\uparrow$} & \multicolumn{3}{c}{\textbf{Copy Rate} $\downarrow$} \\
\cmidrule(lr){3-5}\cmidrule(lr){6-8}\cmidrule(lr){9-11}
\textbf{Model} & \textbf{Dataset} & 0 & \textbf{0.02} & \textbf{0.1} & 0 & \textbf{0.02} & \textbf{0.1} & 0 & \textbf{0.02} & \textbf{0.1} \\
\specialrule{1.1pt}{2pt}{2pt}
\multirow{7}{*}{\rotatebox{90}{\textbf{DUO}}}
 & OWT     & 54.5 & \textbf{55.0} & 54.6 & 53.9 & 83.2 & \textbf{91.6} & 0.003 & 0.04 & 0.16 \\
 & AG News & 41.5 & \textbf{41.8} & 41.7 & 40.7 & 72.5 & \textbf{84.3} & 0.004 & 0.05 & 0.22 \\
 & arXiv   & 48.4 & \textbf{49.7} & 48.7 & 52.7 & 79.6 & \textbf{88.2} & 0.005 & 0.06 & 0.24 \\
 & LM1B    & 35.2 & \textbf{36.5} & 35.5 & 36.3 & 69.9 & \textbf{82.3} & 0.003 & 0.05 & 0.24 \\
 & PubMed  & 46.3 & \textbf{47.7} & 46.8 & 50.2 & 78.4 & \textbf{87.1} & 0.005 & 0.06 & 0.23 \\
 & WT-103  & 45.5 & \textbf{47.2} & 46.2 & 49.1 & 80.8 & \textbf{89.6} & 0.006 & 0.05 & 0.24 \\
\cmidrule(lr){2-11}
 & \textbf{Avg} & 45.2 & \textbf{46.3} & 45.6 & 47.2 & 77.4 & \textbf{87.2} & 0.004 & 0.05 & 0.22 \\
\specialrule{1.1pt}{2pt}{2pt}
\multirow{7}{*}{\rotatebox{90}{\textbf{UDLM}}}
 & OWT     & 55.3 & 55.2 & \textbf{55.5} & 56.7 & 80.1 & \textbf{89.6} & 0.003 & 0.02 & 0.09 \\
 & AG News & 42.8 & 42.6 & \textbf{43.7} & 46.0 & 68.2 & \textbf{81.0} & 0.007 & 0.03 & 0.09 \\
 & arXiv   & 47.4 & \textbf{47.6} & 47.4 & 51.5 & 73.0 & \textbf{83.0} & 0.006 & 0.03 & 0.11 \\
 & LM1B    & \textbf{38.8} & 38.0 & 37.7 & 40.9 & 66.2 & \textbf{78.7} & 0.005 & 0.03 & 0.12 \\
 & PubMed  & 45.4 & \textbf{45.9} & 45.6 & 49.6 & 72.8 & \textbf{82.3} & 0.006 & 0.04 & 0.11 \\
 & WT-103  & \textbf{47.9} & 47.4 & 47.7 & 51.8 & 76.4 & \textbf{86.8} & 0.007 & 0.03 & 0.11 \\
\cmidrule(lr){2-11}
 & \textbf{Avg} & \textbf{46.3} & 46.1 & \textbf{46.3} & 49.4 & 72.8 & \textbf{83.6} & 0.006 & 0.03 & 0.11 \\
\specialrule{1.1pt}{2pt}{2pt}
\end{tabular}}
\end{minipage}\hfill
\begin{minipage}[t]{0.49\textwidth}
\vspace{0pt}\centering
\includegraphics[width=\linewidth]{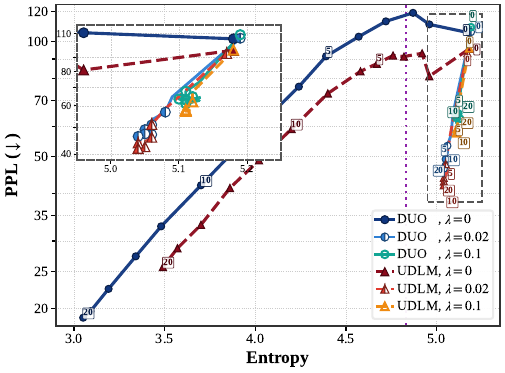}
\captionof{figure}{\textbf{Effect of CTR-Reg coefficient $\lambda$ on greedy-tail generation.} PPL (GPT2-Large, 1024 samples) vs.\ entropy over greedy-tail steps $g$ (budget 512) for DUO (blue, solid, circles) and UDLM (red/orange, dashed, triangles); marker fill encodes $\lambda$ (filled: 0, half: 0.02, hollow: 0.1). Dotted line: real-OWT entropy (4.83); inset: zoom on the converging high-entropy cluster.}\label{fig:lambda-greedy-tail}
\end{minipage}
\end{table}

% \begin{wrapfigure}{r}{0.5\textwidth}
% \vspace{-30pt}
%     \centering
%     \includegraphics[width=\linewidth]{images/lambda_ablation_ppl_entropy_single.pdf}
% \caption{\textbf{Effect of the CTR-Reg coefficient $\lambda$ on greedy-tail generation quality.} PPL (GPT2-Large, 1024 samples) vs.\ entropy across greedy-tail steps $g$ (budget fixed at 512) for DUO (solid, circles) and UDLM (dashed, triangles) at $\lambda \in \{0, 0.02, 0.1\}$. Color encodes model (blue = DUO, red/orange = UDLM); fill style encodes $\lambda$ (filled = 0, half-filled = 0.02, hollow = 0.1). Dotted line: real-OWT reference entropy ($4.83$); inset: zoom on the converging high-entropy cluster.}
% \label{fig:lambda-greedy-tail}
% \vspace{-20pt}
% \end{wrapfigure}

%% file: sections/7_conclusion.tex
\section{Conclusion}
Self-correction in USDMs requires both revising incorrect tokens and retaining correct ones; current USDMs lack retention, and their NELBO barely rewards it. CTR-Reg, a simple auxiliary loss on unperturbed tokens that leaves the sampler unchanged, raises clean-token accuracy by 26.5 points on average without hurting revision and more than halves generative perplexity within five greedy-tail steps while preserving diversity, across DUO, UDLM, and SEDD. Correct-token retention is thus a key ingredient for self-correcting diffusion language models.

%% file: sections/7_5_ICLR-Policy.tex
\subsection*{AI Use Statement.} 
Generative AI tools were used as assistants during manuscript preparation, primarily for language refinement, organization, \LaTeX{} and table formatting, reference checking, numerical consistency checks between the main text and appendix tables, and feedback on clarity and presentation, as well as for syntax and debugging assistance during coding. The research problem, methodology, experimental design, implementation, experiments, analyses, and scientific conclusions were developed and carried out by the authors. All AI-assisted edits, code suggestions, and checks were reviewed and verified by the authors, and the authors take full responsibility for the content of the paper.

\subsection*{Ethics Statement} 
This work involves no human subjects or personal data collection. All models are trained and evaluated on publicly available research corpora, which as web-derived text may contain biases that trained models can reproduce. Our models are small (169M parameters) and intended solely for research, and CTR-Reg, as a training objective, does not introduce new risks or misuse potential beyond those of existing diffusion language models.

\subsection*{Reproducibility statement}
    Section~\ref{implementation_detail} specifies the datasets, backbone, optimization, and evaluation protocols, and Section~\ref{sec:ablation_study} reports the CTR-Reg coefficient used for training. The CTR-Reg objective and its score-space variant for SEDD are fully specified in Sections~\ref{sec:method-ccreg}--\ref{sec:method-sedd}, with derivations in Appendices~\ref{appendix:ancestral-suppress-token-change} and~\ref{app:sedd}, and exact values for all reported results are given in Appendix~\ref{sec:appendix}. The supplementary material contains, for each USDM (DUO, UDLM, and SEDD), the full training code and the code for the random-token corruption experiment and the MAUVE and Gen PPL/entropy evaluations.

\subsubsection*{Acknowledgments}
This work was funded in part by the Swiss National Science Foundation under the ERC-Advanced grant BALM, Swiss National Science Foundation Agreement \#MB25.00024, and supported as part of the Swiss AI Initiative by a grant from the Swiss National Supercomputing Centre (CSCS) under project ID a0236 on Alps.

%% file: sections/8_appendix.tex
% \appendix

\input{figures/Step-Ablation}

\section{Step-Budget Analysis}
\label{sec:appendix-step-budget}

\Cref{fig:step_ablation} extends the greedy-tail analysis of \Cref{sec:ccreg_generation} across total sampling budgets from 16 to 1024 steps, comparing each model's default sampler with greedy-tail sampling (final three steps greedy); exact values are given in \Cref{sec:appendix-step-ablation}. Under each model's default sampler (for DUO, with one trailing greedy step), CTR-Reg matches or outperforms the NELBO-only baseline at every budget, including the ultra-low-step regime. Adding greedy-tail steps widens the gap: CTR-Reg substantially lowers Gen PPL while preserving diversity, whereas the apparent gains of NELBO-only models come at the cost of diversity, consistent with the collapse in \Cref{fig:greedy-tail-lowstep}. The benefit of improved retention therefore holds across sampling budgets.

\input{sections/NELBO-appendix}

\input{sections/theory_ancestral_samling_not_helping}

\input{sections/appendix_likelihood_analysis}

\section{Training on a Larger Corpus Does Not Fix the Problem}
\label{sec:appendix-scale-does-not-fix}

Owing to computational cost, our main experiments train DUO, UDLM, and SEDD from scratch on 20B OpenWebText tokens (169M parameters, sequence length 512). A natural question is whether the correct-token retention deficit we identify is simply an artifact of undertraining, and would disappear if the model were trained on substantially more data. To test this, we repeat both the random-token corruption analysis and the greedy-tail sampling analysis on the official, publicly released DUO checkpoint \citep{sahoo2025the}, which is trained on approximately 524B tokens ($\sim$26$\times$ more than our 20B-token models) with a native sequence length of 1024 (rather than 512). For the corruption experiment, sequences are chunked to this checkpoint's native length of 1024 tokens rather than 512, so that the model is evaluated in-distribution; because of this sequence-length mismatch, we do not directly compare PPL or entropy magnitudes between the 524B-token checkpoint and our 512-length models (see below).

\paragraph{Random-token corruption at scale.} Table~\ref{tab:appendix-524b-corruption} reports corrupted-token, clean-token, and overall reconstruction accuracy for the 524B-token DUO checkpoint alongside our 20B-token NELBO-only and NELBO~+~CTR-Reg models, following the same protocol as Table~\ref{tab:corruption_exp_generalization}. Scaling to 26$\times$ more training tokens raises corrupted-token accuracy substantially (avg.\ 45.2\%~$\rightarrow$~51.6\%), as expected: more data simply makes the model a better denoiser. However, clean-token accuracy improves by only a comparable amount (avg.\ 47.2\%~$\rightarrow$~52.7\%), so the model still cannot tell already-correct tokens apart from corrupted ones: the average clean-minus-corrupted gap is 2.0 points at 20B tokens and only 1.1 points at 524B tokens, i.e., scaling by 26$\times$ leaves this gap unchanged within noise. In contrast, CTR-Reg opens a 31.1-point average gap using only the 20B-token budget, and its clean-token accuracy exceeds the 524B-token checkpoint's by nearly 25 percentage points on average (77.4\% vs.\ 52.7\%) despite seeing 26$\times$ fewer training tokens. This indicates that correct-token retention is not a data-scale bottleneck: training on $26\times$ more data does not close the gap, whereas CTR-Reg closes it within the 20B-token budget.

\definecolor{bluereg}{HTML}{1F5FA6}
\definecolor{bluerow}{HTML}{EAF1FB}
\input{tables/duo-524B-rand-coruption-exp}

\paragraph{Greedy-tail sampling at scale.} Table~\ref{tab:appendix-524b-greedy-tail} reports PPL and generation entropy for the 524B-token DUO checkpoint under greedy-tail sampling, following the protocol of Table~\ref{tab:appendix-greedy-tail-ppl}. The real-OpenWebText reference entropy at sequence length 1024 is $5.40$ (vs.\ $4.83$ at length 512, Table~\ref{tab:appendix-greedy-tail-ppl}), and we use it to color-code entropy in Table~\ref{tab:appendix-524b-greedy-tail} in the same way. As with our 20B-token NELBO-only baselines, PPL falls monotonically as the number of greedy-tail steps increases (77.77 at $g=0$ down to 9.90 at $g=20$), but this is accompanied by a steep, monotonic collapse in entropy (5.54 down to 2.14) with no sign of stabilizing. Only the $g=0$ setting (no greedy-tail steps at all) reaches the real-data entropy of 5.40; every greedy-tail step count $g \geq 1$ falls below it, so the PPL improvements from $g=1$ onward are, by the same criterion as Table~\ref{tab:appendix-greedy-tail-ppl}, achieved through degenerate, low-diversity generation rather than genuine quality gains. Because this checkpoint uses a native sequence length of 1024 rather than 512, we do not compare PPL or entropy magnitudes directly against Table~\ref{tab:appendix-greedy-tail-ppl}; each model is instead judged against its own sequence-length-matched reference entropy. Together with Table~\ref{tab:appendix-524b-corruption}, this shows that training on substantially more data improves corrupted-token denoising but does not improve the model's ability to leave correct tokens unchanged, and the diversity collapse under greedy-tail sampling persists at scale.

\input{tables/duo-524-ppl-entropy}

These results indicate that the correct-token retention deficit in USDMs is not a symptom of insufficient training data. Scaling training data by $26\times$ leaves the clean-vs-corrupted discrimination gap essentially unchanged and does not prevent diversity collapse under greedy-tail sampling, whereas CTR-Reg induces this discrimination directly, at a fraction of the training cost.

\section{CTR-Reg for Score-Parameterized Models: Details}
\label{app:sedd}

This appendix supports \Cref{sec:method-sedd}. Throughout, $i$ indexes the displayed token, $m$ the
clean token, and we write $s_v$ for $s_\theta(z_t,t)_v$.

\subsection{Reading predictions from scores}
\label{app:sedd-inverse}

The network emits $\log s_\theta(z_t,t)$, so the scores $s_v$ are positive. Under the mean
parameterization $s_v=(\bar{x}_\theta)_v/(\bar{x}_\theta)_i$ they determine $\bar{x}_\theta$ only up to
its common factor $(\bar{x}_\theta)_i$, which is positive since $\bar{x}_\theta\ge1-\alpha_t>0$
entrywise. Inverting $\bar{x}_\theta=K\alpha_t x_\theta+(1-\alpha_t)\mathbf{1}$ then gives
\begin{equation}
(x_\theta)_v=\frac{(\bar{x}_\theta)_i\,s_v-(1-\alpha_t)}{K\alpha_t},
\label{eq:score_inverse}
\end{equation}
which is strictly increasing in $s_v$ because $(\bar{x}_\theta)_i>0$ is shared by all $v$. Ranking
tokens by score therefore ranks them by $x_\theta$, whatever the value of the common factor, and
argmax predictions, in greedy-tail decoding (\eqref{eq:greedy_tail}) and in the random-token
corruption experiment (\Cref{sec:exp-retention}), are taken directly over the scores. With $s_i=1$,
the rule reads directly: $s_j<1\iff(x_\theta)_j<(x_\theta)_i$, so the displayed token is kept exactly
when the model ranks it above every alternative.

\subsection{The training target $\rho_t$}
\label{app:sedd-target}

\paragraph{SEDD's objective is a per-coordinate regression onto marginal ratios.}
In \eqref{eq:sedd_nelbo}, each off-diagonal score is trained through a term
$s_j-r_t(j)\log s_j$, minimized at $s_j=r_t(j)$, where $r_t(j)=q_t(j\mid x)/q_t(i\mid x)$; the
remaining factor $g(r_t(j))$ does not involve $\theta$, and for uniform noise the weight
$R_t(z_t,z')$ is identical across all $j\neq i$. The sum runs over $j\neq i$ only, so the displayed
token's own score is never trained: it equals $q_t(i\mid x)/q_t(i\mid x)=1$ by definition, which the
network enforces by fixing the corresponding log-score to zero. Writing the forward marginal
\eqref{eq:forward} with $\pi=\mathbf{1}/K$ as $q_t(v\mid x)=\alpha_t[v=m]+(1-\alpha_t)/K$, and
recalling $\bar{x}=K\,q_t(\cdot\mid x)$, the targets are $r_t(j)=\bar{x}_j/\bar{x}_i$, that is
\begin{equation}
r_t(j)=
\begin{cases}
\rho_t=\dfrac{1-\alpha_t}{K\alpha_t+1-\alpha_t}, & \text{clean position } (i=m),\ \forall j\neq i,\\[8pt]
1/\rho_t, & \text{corrupted position } (i\neq m),\ j=m,\\[4pt]
1, & \text{corrupted position},\ j\notin\{i,m\}.
\end{cases}
\label{eq:sedd_targets}
\end{equation}
Whenever $K\alpha_t\gg1-\alpha_t$, that is for all but the noisiest $t$, $\rho_t\ll1$: at a clean
position the objective pulls every off-diagonal score well below one, so that the displayed token is
retained, while at a corrupted one it pulls $s_m$ far above one, so that the position is revised to
$m$, leaving the remaining wrong tokens at one. Substituting the parameterization
$s_j=(\bar{x}_\theta)_j/(\bar{x}_\theta)_i$ into \eqref{eq:sedd_nelbo} reproduces
\eqref{eq:udlm_nelbo} term by term, which is the sense in which the two objectives are
equivalent~\citep{schiff2024discreteguidance}.

\eqref{eq:sedd_targets} is what \eqref{eq:ccreg_score} builds on: at an unperturbed position
every off-diagonal score has the same target $\rho_t$, and the hinge is placed there
(\Cref{app:sedd-penalty}).

\subsection{Placing the hinge at the target}
\label{app:sedd-penalty}

The derivative of $\mathrm{softplus}\big(\log(s_j/c)\big)$ with respect to $\log s_j$ is
$s_j/(s_j+c)$: close to one for $s_j\gg c$, one half at $s_j=c$, and vanishing for $s_j\ll c$. The
threshold $c$ therefore sets how far the penalty pushes. With $c=1$ the push fades as soon as a score
falls below one, which by \eqref{eq:sedd_targets} is the target for tokens at a corrupted
position, far above $\rho_t$: at $K=50257$ and $\alpha_t=0.9$, where $\rho_t\approx2.2\times10^{-6}$,
the gradient at $s_j=10^{-3}$ is about $10^{-3}$ with $c=1$ but close to one with $c=\rho_t$. With
$c=\rho_t$ the push persists until the score reaches the clean-position target and fades below it,
since $\rho_t$ is already the value SEDD's own objective trains toward. Note that $\log\rho_t$ is
constant in $\theta$; it matters only because it lies inside the nonlinearity.

\section{Random-Token Corruption on Additional Corpora}
\label{sec:appendix-generalization-random_corruption}
To test whether our findings generalize across domains, we repeat the random-token corruption experiment of \Cref{sec:exp-retention} on six corpora: OpenWebText (web text), LM1B and AG News (news), arXiv (scientific papers), PubMed (biomedical abstracts), and WikiText-103 (encyclopedic text). The protocol is unchanged: each corpus is split into sequences of $L=512$ tokens, $p\%$ of positions are replaced with tokens drawn uniformly from the vocabulary, and each model reconstructs the sequence in a single forward pass.

\Cref{tab:corruption_exp_generalization} reports Clean- and Corrupted-Token Accuracy, averaged over corruption rates $p\in\{10,20,30,40\}\%$. Two patterns hold on every corpus. First, NELBO-only models reach nearly identical accuracy on clean and corrupted positions, even though clean positions are the easier target, confirming that they revise tokens indiscriminately regardless of domain. Second, CTR-Reg raises Clean-Token Accuracy for all three models on all six corpora, by 26.5 percentage points on average, while changing Corrupted-Token Accuracy by less than 1.7 percentage points in every cell. The retention gain therefore generalizes across domains and does not come at the expense of revision.

Tables~\ref{tab:appendix-agnews}, \ref{tab:appendix-arxiv}, \ref{tab:appendix-lm1b}, \ref{tab:appendix-pubmed}, and \ref{tab:appendix-wt103} give the full per-corruption-rate breakdown underlying these averages for AG News, arXiv, LM1B, PubMed, and WikiText-103. The same two patterns hold at every individual corruption rate, so the averaged results are not driven by any single value of $p$.

\input{tables/rand_perturb_exp}

\section{Additional Experimental Results}
\label{sec:appendix}

This appendix reports the exact numerical values underlying the results aggregated in the main text.

\subsection{Exact Values for the Greedy-Tail Step Sweep}
\label{sec:appendix-greedy-tail}
Table~\ref{tab:appendix-greedy-tail-ppl} reports the exact perplexity and entropy values underlying Figure~\ref{fig:greedy-tail-lowstep}, for every greedy-tail step count $g$ and each of the three USDMs.

\subsection{Exact Values Across Sampling-Step Budgets}
\label{sec:appendix-step-ablation}
Table~\ref{tab:appendix-step-ablation} reports the exact perplexity and entropy values underlying Figure~\ref{fig:step_ablation}, for sampling-step budgets from 16 to 1024 under both standard ancestral sampling and greedy-tail sampling.

\subsection{Exact Values for the MAUVE Greedy-Tail Sweep}
\label{sec:appendix-mauve-greedy-tail}
Table~\ref{tab:appendix-greedy-tail-full} reports the exact PPL, entropy, and MAUVE values underlying Figure~\ref{fig:mauve-greedy-tail}, for every greedy-tail step count $g \in \{0,1,3,6,9\}$ and each of the three USDMs, with and without CTR-Reg.

\subsection{Per-Corruption-Rate Breakdown for the 524B-Token Checkpoint}
\label{sec:appendix-scale-percorruption}
Section~\ref{sec:appendix-scale-does-not-fix} reports accuracy for the 524B-token DUO checkpoint averaged over corruption rates $p\in\{10,20,30,40\}\%$ across six evaluation corpora. Table~\ref{tab:appendix-524b-percorruption} gives the full per-corruption-rate breakdown underlying these averages, confirming that corrupted-token and clean-token accuracy remain close to each other at every individual corruption rate, not only on average.

\subsection{Exact Values for the CTR-Reg Coefficient Ablation}
\label{sec:appendix-lambda-greedy-tail-values}
Table~\ref{tab:appendix-lambda-greedy-tail-values} reports the exact PPL and entropy values underlying Figure~\ref{fig:lambda-greedy-tail}, for DUO and UDLM at $\lambda \in \{0, 0.02, 0.1\}$ across every greedy-tail step count $g$.

\subsection{Per-Corruption-Rate Breakdown for the CTR-Reg Coefficient Ablation}
\label{sec:appendix-lambda-percorruption}
Table~\ref{tab:appendix-lambda-corruption} reports corruption-probe accuracy averaged over corruption rates $p\in\{10,20,30,40\}\%$ for $\lambda \in \{0, 0.02, 0.1\}$. Table~\ref{tab:appendix-lambda01-percorruption} gives the full per-corruption-rate breakdown underlying the $\lambda=0.1$ row across all six evaluation corpora, for DUO and UDLM.

Table~\ref{tab:appendix-lambda-copyrate} gives the per-corruption-rate breakdown of the copy rate for all three $\lambda$. The copy rate stays below 0.35\% at every corruption rate, so the averages in Table~\ref{tab:appendix-lambda-corruption} are not hiding a high-copy regime. For DUO it grows with $p$ (e.g., 0.15\%$\to$0.30\% on AG News at $\lambda=0.1$), whereas for UDLM it is flat or slightly decreasing.

% \input{sections/temp_val_ppl}

\input{tables/ppl-entropy-greedy-step}

\input{tables/step-ablation-full}

\input{tables/rand_perturb_exp_full_coruption_rate}

\input{tables/mauve-detail-results}

\input{tables/duo-524-corruption-rate-detail}

\input{tables/lambda_ablation_ppl_entropy_detail}

\input{tables/lamda-rand-corruption-detail}

\input{tables/copy-rate-detail}

%% file: figures/Step-Ablation.tex
\begin{figure}[h]
\centering
\includegraphics[width=\linewidth]{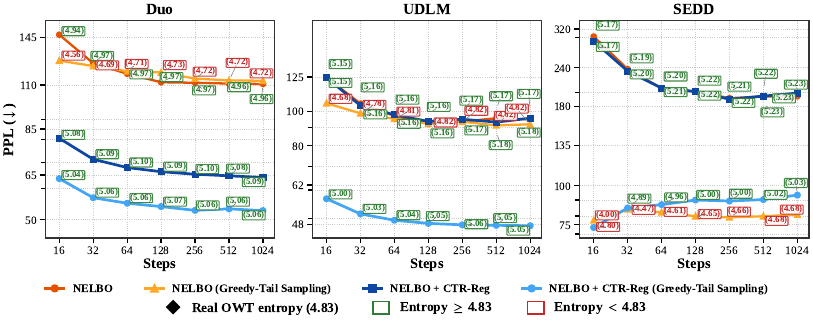}
\caption{\textbf{Generation quality across sampling budgets (16--1024 steps).} PPL ($\downarrow$) vs.\ total sampling steps for DUO, UDLM, and SEDD trained with NELBO or NELBO + CTR-Reg, under default (DUO: +1 greedy step) and greedy-tail sampling (final three steps greedy); each point is annotated with its generation entropy. Under the default sampler, CTR-Reg matches or outperforms NELBO-only training; under greedy-tail sampling, CTR-Reg substantially lowers PPL while preserving diversity, whereas NELBO-only models lower PPL only at the cost of diversity. These gains hold across the full step-budget range, including the ultra-low-step regime.}\label{fig:step_ablation}
% \caption{\textbf{Generation quality across sampling budgets from 16 to 1024 steps, with and without greedy-tail sampling.} Each panel plots PPL ($\downarrow$; y-axis) against the total number of diffusion sampling steps for DUO, UDLM, and SEDD, trained with either the base NELBO loss or NELBO plus our Correct-Token Retention Regularization (CTR-Reg). For each model, we compare standard ancestral sampling against greedy-tail ancestral sampling, which fixes the final three steps to be greedy. Each point is annotated with its generation entropy in parentheses. The results show: (i) under standard ancestral sampling (no greedy-tail steps), CTR-Reg matches or outperforms NELBO-only training; (ii) under greedy-tail ancestral sampling, CTR-Reg substantially improves quality (lower PPL) while preserving diversity; (iii) under greedy-tail sampling, NELBO-only training achieves lower perplexity but at the cost of diversity (entropy), which undermines quality and makes the PPL improvement uninformative; and (iv) the improvement from CTR-Reg remains consistent across the full range of step budgets, including the ultra-low-step regime.}\label{fig:step_ablation}
\end{figure}
% \vspace{-10pt}

%% file: sections/NELBO-appendix.tex
% ===========================================================================
% APPENDIX --- place in the appendix, after the main text
% ===========================================================================

\section{Per-Model USDM Objectives}
\label{appendix:usdm-objectives}

This appendix gives the explicit training objectives of DUO and SEDD, both of which are equivalent to the continuous-time bound of \eqref{eq:udlm_nelbo} that UDLM optimizes directly. We reuse the notation of \Cref{subsec:usdm}: $i$ indexes the non-zero entry of $z_t$, $\bar{x}=K\alpha_t x+(1-\alpha_t)\mathbf{1}$, $\bar{x}_\theta=K\alpha_t x_\theta(z_t,t)+(1-\alpha_t)\mathbf{1}$, and $\alpha_t'=d\alpha_t/dt$.

\paragraph{DUO.} \citet{sahoo2025the} rewrite \eqref{eq:udlm_f} so that $\bar{x}$ never has to be materialized, and compute certain expectations analytically via Rao-Blackwellization; the authors describe the result as an efficient, low-variance formulation with minimal memory overhead. With $m$ the index of the non-zero entry of $x$ and $\rho_t=(1-\alpha_t)/(K\alpha_t+1-\alpha_t)$, the per-token loss becomes
\begin{equation}\small
\begin{split}
f_{\mathrm{DUO}}={}&\frac{\alpha_t'}{K\alpha_t}\Bigg[\frac{K}{\bar{x}_i}-\frac{K}{(\bar{x}_\theta)_i}-\big(\rho_t\,\mathbb{I}[z_t=x]+\mathbb{I}[z_t\neq x]\big)\sum_{j=1}^{K}\log\frac{(\bar{x}_\theta)_i}{(\bar{x}_\theta)_j}\\
&-\frac{K\alpha_t}{1-\alpha_t}\,\mathbb{I}[z_t\neq x]\log\frac{(\bar{x}_\theta)_i}{(\bar{x}_\theta)_m}-\Big((K-1)\rho_t\,\mathbb{I}[z_t=x]-\frac{\mathbb{I}[z_t\neq x]}{\rho_t}\Big)\log\rho_t\Bigg].
\end{split}
\label{eq:duo_f}
\end{equation}
DUO further accelerates training with a curriculum that relaxes the discrete latents through an underlying Gaussian diffusion~\citep{sahoo2025the}, which \citet{deschenaux2026the} later make more memory-efficient.

\paragraph{SEDD and CTMC-based formulations.} Rather than bounding a discrete-time chain and taking that bound to the continuous-time limit, \citet{campbell2022continuous} and SEDD~\citep{lou2024discrete} formulate discrete diffusion directly with continuous-time Markov chains (CTMCs); both still optimize a bound on the likelihood. The forward process is defined by a rate matrix $R_t\in\mathbb{R}^{K\times K}$ through $q(z_t=z'\mid z_s=z)=\delta_{z',z}+R_t(z,z')/T+o(1/T)$ with $s=t-1/T$. For the uniform interpolating process, $R_t=-\frac{\alpha_t'}{K\alpha_t}(\mathbf{1}\mathbf{1}^\top-KI)$~\citep{schiff2024discreteguidance}, which coincides with SEDD-Uniform's rate $\sigma(t)(\mathbf{1}\mathbf{1}^\top/K-I)$ for $\alpha_t=e^{-\bar{\sigma}(t)}$ and $\bar{\sigma}(t)=\int_0^t\sigma(u)\,du$. Reversing this chain requires ratios of marginals: for $z'\neq z$, the reverse rate is $\hat{R}_t(z,z')=\frac{q_t(z')}{q_t(z)}R_t(z',z)$. These ratios, the concrete score~\citep{meng2022concrete}, replace the score $\nabla\log q_t$ used by continuous score-based models~\citep{song2021score}, since discrete data admit no gradient. Because tokens are corrupted independently, only sequences that differ from $z_t$ in a single position have non-zero rate. Whereas UDLM, DUO and \citet{campbell2022continuous} predict the clean data, SEDD's network outputs these ratios directly, one per position and candidate token: $s_\theta(z_t,t)\in\mathbb{R}^{L\times K}$ with $s_\theta(z_t,t)_{z'}\approx q_t(z')/q_t(z_t)$. It is trained with the diffusion-weighted denoising score entropy, which forms a NELBO~\citep{lou2024discrete}:
\begin{equation}\small
\mathcal{L}_{\mathrm{SEDD}}=\mathbb{E}_{t\sim\mathcal{U}[0,1]}\,\mathbb{E}_{z_t\sim q_t(\cdot\mid x)}\sum_{z'\neq z_t}R_t(z',z_t)\Big[s_\theta(z_t,t)_{z'}-r_t(z')\log s_\theta(z_t,t)_{z'}+g\big(r_t(z')\big)\Big],
\label{eq:sedd_nelbo}
\end{equation}
where $r_t(z')=q_t(z'\mid x)/q_t(z_t\mid x)$ and $g(a)=a(\log a-1)$. Under the mean parameterization $s_\theta(z_t,t)_{z'}=(\bar{x}_\theta)_j/(\bar{x}_\theta)_i$, with $j$ indexing the non-zero entry of $z'$, \citet{schiff2024discreteguidance} show that \eqref{eq:sedd_nelbo} is equivalent to \eqref{eq:udlm_nelbo}.

%% file: sections/theory_ancestral_samling_not_helping.tex
\section{Ancestral Sampling Suppresses Indiscriminate Token Changes}
\label{appendix:ancestral-suppress-token-change}

We show that the ancestral sampler's noise schedule, rather than the
denoiser's predictions, is what prevents the sampler from adopting the
model's predictions en masse. The schedule assigns most of its
probability mass to retaining the currently decoded token, which bounds
the number of positions that can be revised toward the model's prediction
at any single step.

\subsection{Single-position transition probabilities}

Consider a position $\ell$ with $z_t^\ell \neq x_\theta^\ell$ (both
one-hot), so that we are asking how readily the sampler incorporates a
change the model has already decided on. The cross term in the reverse
posterior (\eqref{eq:reverse}) then vanishes, since
$z_t^\ell\odot x_\theta^\ell=\mathbf{0}$ and
$\langle z_t^\ell,x_\theta^\ell\rangle=0$, leaving
\[
q_{s|t}(z_s^\ell\mid z_t^\ell,x_\theta^\ell) = \mathrm{Cat}\left(\cdot\,;\,
\frac{(\alpha_{t|s}-\alpha_t)z_t^\ell + (\alpha_s-\alpha_t)x_\theta^\ell +
(1-\alpha_{t|s})(1-\alpha_s)\mathbf{1}/K}{1-\alpha_t}\right).
\]
Collecting the weight assigned to each of the $K$ categories and dropping
$O(1/K)$ terms (negligible for $K\approx 50\text{k}$) yields three
outcomes:
\begin{equation}
P(\text{stay}) \approx \frac{\alpha_{t|s}-\alpha_t}{1-\alpha_t},
\label{eq:stay}
\end{equation}
\begin{equation}
P(\text{jump-to-prediction}) \approx \frac{\alpha_s-\alpha_t}{1-\alpha_t},
\label{eq:jump}
\end{equation}
\begin{equation}
P(\text{jump-elsewhere}) \approx \frac{(1-\alpha_{t|s})(1-\alpha_s)}{1-\alpha_t}.
\label{eq:elsewhere}
\end{equation}
Summing the exact, un-truncated weights and applying
$\alpha_{t|s}\alpha_s=\alpha_t$ confirms that these normalize to one.
Note that \eqref{eq:stay}--\eqref{eq:elsewhere} assume no functional form
for $\alpha_t$ and therefore hold for any interpolating noise schedule.

\subsection{The schedule bounds the number of changes per step}
\label{app:scheduler_bound}
Under the log-linear noise schedule used in our experiments, $\bar{\sigma}(t) = -\log\big(1 - (1-\epsilon)t\big)$, so $\alpha_t = e^{-\bar{\sigma}(t)} = 1 - (1-\epsilon)t$ is linear in $t$; with $\epsilon = 10^{-3}$ we write $\alpha_t \approx 1 - t$.
With step size $h = 1/T$, we have $\alpha_s - \alpha_t = (1-\epsilon)h \approx h$ at every step, so \eqref{eq:jump} simplifies to
\begin{equation}
P(\text{jump-to-prediction}) \approx \frac{h}{1-\alpha_t}.
\label{eq:jump-linear}
\end{equation}
This is $O(1/T)$ for any $\alpha_t$ bounded away from one, and approaches
$O(1)$ only as the remaining noise budget $1-\alpha_t$ shrinks toward $h$.
For a general smooth schedule, $\alpha_s-\alpha_t\approx|\alpha'_t|h$, so
\eqref{eq:jump-linear} becomes $|\alpha'_t|h/(1-\alpha_t)$: the $O(1/T)$
per-step suppression holds for any such schedule, which enters only
through $|\alpha'_t|$. We specialize to the linear case for concreteness,
matching our experimental setup.

Consequently, if $m$ positions disagree with the denoiser's prediction at
step $t$, the expected number that adopt that prediction on the step is
\begin{equation}
\mathbb{E}[\#\text{changed}] \approx \frac{mh}{1-\alpha_t},
\label{eq:expected-changed}
\end{equation}
which remains small whenever $h\ll 1-\alpha_t$. Through most of the
trajectory the sampler therefore revises only a small fraction of the
disagreeing positions per step, never all $m$ simultaneously; the bound
loosens only near the terminal region.

\begin{table}[h]
\centering
\caption{Single-step transition probabilities at a disagreeing position under the log-linear schedule ($\alpha_t \approx 1 - t$) with $T = 512$ ($h \approx 1.95 \times 10^{-3}$).}
\label{tab:appendix-transition}
\begin{tabular}{lccc}
\toprule
$\alpha_t$ & stay & jump-to-prediction & jump-elsewhere \\
\midrule
0.01 & 0.835 & 0.002 & 0.163 \\
0.50 & 0.992 & 0.004 & 0.004 \\
0.99 & 0.803 & 0.195 & 0.002 \\
\bottomrule
\end{tabular}
\end{table}

\Cref{tab:appendix-transition} evaluates
\eqref{eq:stay}--\eqref{eq:elsewhere} at three points along the
trajectory. Even at $\alpha_t=0.99$, where the bound is loosest, roughly
four out of five disagreeing positions still fail to update in a single
step. The bound constrains \emph{model-driven} revisions specifically:
early in generation ($\alpha_t=0.01$) many positions do change, but
almost all of these are draws from the uniform prior
(\eqref{eq:elsewhere}) rather than adoptions of the denoiser's
prediction, which remains suppressed at $0.002$ throughout that regime.

\subsection{Why the greedy tail reveals what ancestral sampling hides}

The greedy tail sets $z_s^\ell=x_\theta^\ell$ deterministically, so
$P_g(\text{jump-to-prediction})=1$ for every $\alpha_t$ and $h$. This
removes the suppression term in \eqref{eq:jump-linear} and lifts the
bound in \eqref{eq:expected-changed} entirely: all $m$ disagreeing
positions update at once, rather than the $O(mh)$ fraction permitted by
ancestral sampling. A denoiser whose argmax is unstable across steps
therefore exhibits little visible instability under ancestral sampling,
while the same instability is fully exposed under the greedy tail. The
near-complete, non-converging rewrites in
\Cref{fig:token-change-vs-greedy-steps} indicate that this instability
was present throughout ancestral sampling and was simply not surfaced.

%% file: sections/appendix_likelihood_analysis.tex
\section{Likelihood Analysis}
\label{app:likelihood}

This appendix supports the likelihood analysis of \Cref{sec:token_change_mechanism_method} (\Cref{fig:nelbo_four_cases}). All quantities are computed from the validation NELBO of each model \emph{without} the CTR-Reg term, using the EMA weights and the six evaluation corpora of \Cref{sec:appendix-generalization-random_corruption} at sequence length 512. For each sequence, we draw one diffusion time $t$ (stratified within each batch) and one corruption $z_t \sim q_t(\cdot \mid x)$, and evaluate the per-token NELBO integrand of each model: \eqref{eq:udlm_nelbo} for UDLM, and the equivalent per-token objectives of DUO and SEDD given in \Cref{appendix:usdm-objectives} (for SEDD, the diffusion-weighted score entropy of \eqref{eq:sedd_nelbo}). NELBO-only and CTR-Reg models of the same USDM are evaluated with the same seed and batch size, so both see identical $t$ and corruptions. DUO and SEDD are evaluated on identical token sequences, whereas UDLM uses its own data pipeline over the same corpora, which yields slightly different chunks.

\subsection{Validation Perplexity}
\label{app:val_ppl}

\Cref{tab:val_ppl} reports the resulting perplexity upper bound, $\exp(\text{NELBO per token})$. CTR-Reg increases perplexity by 1.5\% on average for DUO and 1.2\% for UDLM, with a maximum of 4.4\% (DUO on PubMed), and decreases it by 2.4\% on average for SEDD, on five of six corpora. The in-distribution (OWT) change is below 1\% for DUO and UDLM.

\subsection{Decomposing the NELBO by Position Type and Prediction}
\label{app:nelbo_decomposition}

To measure how much the likelihood rewards retention, we split every validation position into four groups: \emph{clean} ($z_t = x$) or \emph{corrupted} ($z_t \neq x$), and whether the model's argmax prediction $\hat{x}_\theta$ equals the clean token ($\hat{x}_\theta = x$) or not ($\hat{x}_\theta \neq x$). For SEDD, $\hat{x}_\theta$ is taken directly over the scores, which yields the same ranking as $x_\theta$ (\Cref{app:sedd-inverse}). Positions where a uniform draw happens to coincide with the original token count as clean, consistent with the NELBO. We average the per-token NELBO within each group and within ten bins of the noise level $1-\alpha_t$, the expected fraction of corrupted tokens. For DUO and UDLM, whose log-linear schedule gives $1-\alpha_t \approx t$, these bins are bins of $t$. SEDD uses a geometric schedule, so its bins are bins of $1-\alpha_t = 1-e^{-\sigma(t)}$, and its per-token magnitudes are not directly comparable to those of the other two models.

\Cref{fig:nelbo_four_cases_appendix} extends \Cref{fig:nelbo_four_cases} to the remaining four corpora, and \Cref{tab:nelbo_four_cases_duo,tab:nelbo_four_cases_udlm,tab:nelbo_four_cases_sedd} give the exact values underlying both figures, together with the number of tokens behind each mean. The pattern is consistent across all models and corpora. Clean positions where the prediction is incorrect ($\hat{x}_\theta\neq x$) make up 23--31\% of all validation positions yet contribute only 3.5--5.2\% of the NELBO, whereas corrupted positions with incorrect predictions make up 25--41\% of positions and contribute 90--92\%; correctly predicted clean positions contribute almost nothing. At noise levels $1-\alpha_t < 0.4$, an incorrect prediction at a corrupted position costs 13--100 nats more than a correct one, whereas at a clean position it costs only 0.09--1.32 nats more. The penalty at clean positions grows only near $1-\alpha_t = 1$, where almost every displayed token is noise and replacing it is reasonable. Means at the highest noise levels rest on few clean tokens and are correspondingly noisier; these cells are grayed out in the tables.

\input{tables/validation_ppl}

\input{figures/nelbo_loss_4_additional_dataset}

% \clearpage
\input{tables/NELBO-figrue-detail}
% \clearpage

% \input{tables/validation_ppl}

% \input{figures/nelbo_loss_4_additional_dataset}

% \input{tables/NELBO-figrue-detail}

%% file: tables/validation_ppl.tex
% requires \usepackage{booktabs,multirow,xcolor}
\begin{table}[t]
\centering
\newcommand{\tabCTR}[1]{\textcolor{purple}{#1}}
\newcommand{\tabDelta}[1]{{\scriptsize\textcolor{gray}{#1}}}
\caption{\textbf{Validation perplexity ($\downarrow$) of NELBO-only and CTR-Reg models.} Perplexity upper bound computed from the validation NELBO alone; the CTR-Reg term is excluded at evaluation. All models have 169M parameters and are trained on 20B OpenWebText (OWT) tokens with sequence length 512. $\Delta$: relative change of \tabCTR{NELBO + CTR-Reg} with respect to NELBO-only. CTR-Reg changes perplexity by at most a few percent, and lowers it on average for SEDD.}
\label{tab:val_ppl}
\small
\setlength{\tabcolsep}{4.5pt}
\renewcommand{\arraystretch}{1.1}
\begin{tabular}{ll c cccccc}
\toprule
& & \multicolumn{1}{c}{In-dist.} & \multicolumn{5}{c}{Zero-shot} & \\
\cmidrule(lr){3-3} \cmidrule(lr){4-8}
Model & Loss & OWT & LM1B & AG News & arXiv & PubMed & WT-103 & Avg. \\
\midrule
\multirow{3}{*}{DUO}  & NELBO              & 36.59 & 93.21 & 114.21 & 62.53 & 74.81 & 51.31 & 72.11 \\
                      & \tabCTR{+ CTR-Reg} & \tabCTR{36.74} & \tabCTR{93.66} & \tabCTR{114.47} & \tabCTR{64.21} & \tabCTR{78.09} & \tabCTR{51.78} & \tabCTR{73.16} \\
                      & \tabDelta{$\Delta$} & \tabDelta{+0.4\%} & \tabDelta{+0.5\%} & \tabDelta{+0.2\%} & \tabDelta{+2.7\%} & \tabDelta{+4.4\%} & \tabDelta{+0.9\%} & \tabDelta{+1.5\%} \\
\midrule
\multirow{3}{*}{UDLM} & NELBO              & 40.31 & 128.92 & 139.37 & 77.24 & 91.03 & 62.80 & 89.95 \\
                      & \tabCTR{+ CTR-Reg} & \tabCTR{40.65} & \tabCTR{129.84} & \tabCTR{143.68} & \tabCTR{77.37} & \tabCTR{91.07} & \tabCTR{63.69} & \tabCTR{91.05} \\
                      & \tabDelta{$\Delta$} & \tabDelta{+0.8\%} & \tabDelta{+0.7\%} & \tabDelta{+3.1\%} & \tabDelta{+0.2\%} & \tabDelta{+0.0\%} & \tabDelta{+1.4\%} & \tabDelta{+1.2\%} \\
\midrule
\multirow{3}{*}{SEDD} & NELBO              & 68.07 & 155.47 & 229.39 & 115.25 & 154.18 & 98.33 & 136.78 \\
                      & \tabCTR{+ CTR-Reg} & \tabCTR{66.87} & \tabCTR{152.93} & \tabCTR{228.08} & \tabCTR{109.38} & \tabCTR{144.97} & \tabCTR{98.79} & \tabCTR{133.50} \\
                      & \tabDelta{$\Delta$} & \tabDelta{$-$1.8\%} & \tabDelta{$-$1.6\%} & \tabDelta{$-$0.6\%} & \tabDelta{$-$5.1\%} & \tabDelta{$-$6.0\%} & \tabDelta{+0.5\%} & \tabDelta{$-$2.4\%} \\
\bottomrule
\end{tabular}
\end{table}

%% file: figures/nelbo_loss_4_additional_dataset.tex
\begin{figure}[h]
\centering
\includegraphics[width=\linewidth]{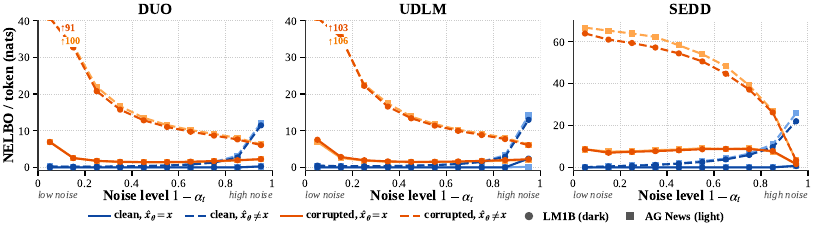}\\[4pt]
\includegraphics[width=\linewidth]{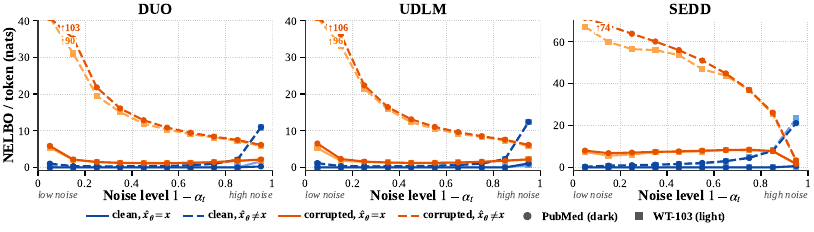}
\caption{\textbf{The NELBO barely penalizes failures of retention on the remaining corpora.} Mean validation NELBO per token of NELBO-only DUO, UDLM, and SEDD on LM1B and AG News (top) and PubMed and WikiText-103 (bottom), with the first corpus of each pair dark and the second light, versus the noise level $1-\alpha_t$ (fraction of corrupted tokens). Lines show clean (blue) and corrupted (orange) positions where the argmax prediction $\hat{x}_\theta$ equals the clean token $x$ (solid) or not (dashed), so the solid–dashed gap is the likelihood penalty for an incorrect prediction. As on OpenWebText and arXiv (Figure~\ref{fig:nelbo_four_cases}), this penalty is large at corrupted positions and near zero at clean ones, except at the highest noise. Magnitudes are comparable within, not across, panels.}
\label{fig:nelbo_four_cases_appendix}\label{fig:nelbo_four_cases_appendix}
\end{figure}

%% file: tables/NELBO-figrue-detail.tex
% requires \usepackage{booktabs,multirow,xcolor,colortbl}
\begin{table}[p]
\centering
\definecolor{nClean}{HTML}{0D47A1}\definecolor{nCorrupt}{HTML}{E65100}\definecolor{nCleanBg}{HTML}{EAF1FB}\definecolor{nCorruptBg}{HTML}{FDF0E6}\definecolor{nLow}{HTML}{9E9E9E}
\caption{\textbf{Exact values underlying Figures~\ref{fig:nelbo_four_cases} and~\ref{fig:nelbo_four_cases_appendix} for DUO.} Mean validation NELBO per token (nats) of the NELBO-only model and the number of tokens behind each mean, by corpus and noise-level bin $1-\alpha_t$ (fraction of corrupted tokens), for \textcolor{nClean}{\textbf{clean}} and \textcolor{nCorrupt}{\textbf{corrupted}} positions where the argmax prediction $\hat{x}_\theta$ equals the clean token $x$ or not. $^{\dagger}$In units of $10^{-3}$ nats. \textcolor{nLow}{Gray}: fewer than 1{,}000 tokens (noisier estimate).}
\label{tab:nelbo_four_cases_duo}
\scriptsize
\setlength{\tabcolsep}{3.5pt}
\renewcommand{\arraystretch}{1.05}
\begin{tabular}{ll >{\columncolor{nCleanBg}}r>{\columncolor{nCleanBg}}r>{\columncolor{nCorruptBg}}r>{\columncolor{nCorruptBg}}r >{\columncolor{nCleanBg}}r>{\columncolor{nCleanBg}}r>{\columncolor{nCorruptBg}}r>{\columncolor{nCorruptBg}}r}
\toprule
\rowcolor{white}& & \multicolumn{4}{c}{\textbf{Mean NELBO per token (nats)}} & \multicolumn{4}{c}{\textbf{Number of tokens}} \\
\cmidrule(lr){3-6} \cmidrule(lr){7-10}
\rowcolor{white}& & \multicolumn{2}{c}{\textcolor{nClean}{\textbf{Clean}}} & \multicolumn{2}{c}{\textcolor{nCorrupt}{\textbf{Corrupted}}} & \multicolumn{2}{c}{\textcolor{nClean}{\textbf{Clean}}} & \multicolumn{2}{c}{\textcolor{nCorrupt}{\textbf{Corrupted}}} \\
\cmidrule(lr){3-4} \cmidrule(lr){5-6} \cmidrule(lr){7-8} \cmidrule(lr){9-10}
\rowcolor{white}\textbf{Corpus} & $1-\alpha_t$ & $\hat{x}_\theta\!=\!x$$^{\dagger}$ & $\hat{x}_\theta\!\neq\!x$ & $\hat{x}_\theta\!=\!x$ & $\hat{x}_\theta\!\neq\!x$ & $\hat{x}_\theta\!=\!x$ & $\hat{x}_\theta\!\neq\!x$ & $\hat{x}_\theta\!=\!x$ & $\hat{x}_\theta\!\neq\!x$ \\
\midrule
\multirow{10}{*}{\textbf{OpenWebText}} & $[0.0, 0.1)$ & 0.019 & 0.1587 & 4.7935 & 84.0081 & 7{,}062{,}434 & 3{,}572{,}467 & 378{,}832 & 186{,}779 \\
 & $[0.1, 0.2)$ & 0.027 & 0.1111 & 1.7759 & 29.8928 & 6{,}139{,}616 & 3{,}545{,}775 & 1{,}074{,}039 & 632{,}058 \\
 & $[0.2, 0.3)$ & 0.040 & 0.1255 & 1.2435 & 18.8710 & 5{,}059{,}097 & 3{,}420{,}197 & 1{,}627{,}315 & 1{,}195{,}791 \\
 & $[0.3, 0.4)$ & 0.063 & 0.1586 & 1.0497 & 14.1688 & 4{,}087{,}763 & 3{,}298{,}748 & 2{,}052{,}332 & 1{,}915{,}781 \\
 & $[0.4, 0.5)$ & 0.110 & 0.2154 & 0.9840 & 11.5886 & 3{,}121{,}303 & 3{,}108{,}428 & 2{,}283{,}455 & 2{,}808{,}670 \\
 & $[0.5, 0.6)$ & 0.213 & 0.3119 & 0.9903 & 9.9526 & 2{,}253{,}591 & 2{,}872{,}115 & 2{,}341{,}637 & 3{,}914{,}417 \\
 & $[0.6, 0.7)$ & 0.480 & 0.4894 & 1.0615 & 8.7982 & 1{,}454{,}797 & 2{,}506{,}955 & 2{,}157{,}867 & 5{,}178{,}685 \\
 & $[0.7, 0.8)$ & 1.358 & 0.8690 & 1.2128 & 7.9157 & 805{,}436 & 2{,}039{,}583 & 1{,}804{,}974 & 6{,}700{,}535 \\
 & $[0.8, 0.9)$ & 6.010 & 1.9837 & 1.4987 & 7.1404 & 327{,}824 & 1{,}384{,}040 & 1{,}294{,}731 & 8{,}358{,}269 \\
 & $[0.9, 1.0]$ & 222.624 & 9.8801 & 1.9753 & 5.9091 & 66{,}583 & 507{,}961 & 732{,}809 & 9{,}994{,}535 \\
\midrule
\multirow{10}{*}{\textbf{LM1B}} & $[0.0, 0.1)$ & 0.025 & 0.2502 & 6.9276 & 91.0346 & 452{,}913 & 344{,}306 & 23{,}380 & 19{,}081 \\
 & $[0.1, 0.2)$ & 0.036 & 0.1799 & 2.5497 & 32.7434 & 393{,}046 & 350{,}682 & 66{,}084 & 65{,}196 \\
 & $[0.2, 0.3)$ & 0.055 & 0.2066 & 1.7987 & 20.7934 & 304{,}654 & 329{,}088 & 94{,}251 & 115{,}783 \\
 & $[0.3, 0.4)$ & 0.089 & 0.2638 & 1.5314 & 15.7570 & 242{,}774 & 320{,}485 & 115{,}453 & 185{,}544 \\
 & $[0.4, 0.5)$ & 0.158 & 0.3617 & 1.4475 & 12.8561 & 176{,}240 & 298{,}517 & 123{,}104 & 266{,}907 \\
 & $[0.5, 0.6)$ & 0.307 & 0.5154 & 1.4576 & 11.0393 & 120{,}936 & 267{,}303 & 118{,}772 & 355{,}197 \\
 & $[0.6, 0.7)$ & 0.690 & 0.7823 & 1.5611 & 9.7381 & 73{,}915 & 227{,}277 & 103{,}354 & 452{,}542 \\
 & $[0.7, 0.8)$ & 1.908 & 1.3204 & 1.7287 & 8.6864 & 38{,}960 & 176{,}301 & 83{,}444 & 559{,}919 \\
 & $[0.8, 0.9)$ & 8.310 & 2.7153 & 1.9655 & 7.6393 & 16{,}485 & 115{,}337 & 60{,}845 & 681{,}829 \\
 & $[0.9, 1.0]$ & 275.655 & 11.4829 & 2.2610 & 6.0831 & 3{,}642 & 38{,}471 & 40{,}216 & 761{,}959 \\
\midrule
\multirow{10}{*}{\textbf{AG News}} & $[0.0, 0.1)$ & 0.027 & 0.2943 & \textcolor{nLow}{7.0253} & \textcolor{nLow}{100.4483} & 20{,}614 & 14{,}506 & \textcolor{nLow}{985} & \textcolor{nLow}{759} \\
 & $[0.1, 0.2)$ & 0.039 & 0.1826 & 2.5778 & 33.2478 & 20{,}138 & 17{,}230 & 3{,}513 & 3{,}151 \\
 & $[0.2, 0.3)$ & 0.058 & 0.2181 & 1.7661 & 21.9576 & 12{,}619 & 12{,}733 & 3{,}943 & 4{,}497 \\
 & $[0.3, 0.4)$ & 0.092 & 0.2715 & 1.4993 & 16.6985 & 12{,}103 & 14{,}451 & 5{,}671 & 8{,}223 \\
 & $[0.4, 0.5)$ & 0.175 & 0.4011 & 1.4325 & 13.5510 & 8{,}656 & 13{,}967 & 5{,}759 & 12{,}578 \\
 & $[0.5, 0.6)$ & 0.331 & 0.5495 & 1.3784 & 11.5594 & 5{,}886 & 13{,}036 & 5{,}624 & 17{,}438 \\
 & $[0.6, 0.7)$ & 0.768 & 0.8787 & 1.4398 & 10.1959 & 2{,}967 & 9{,}720 & 4{,}146 & 20{,}031 \\
 & $[0.7, 0.8)$ & 1.905 & 1.4776 & 1.5987 & 9.1441 & 1{,}688 & 8{,}262 & 3{,}473 & 26{,}513 \\
 & $[0.8, 0.9)$ & \textcolor{nLow}{7.810} & 3.2152 & 1.9279 & 8.0340 & \textcolor{nLow}{550} & 5{,}176 & 2{,}142 & 30{,}532 \\
 & $[0.9, 1.0]$ & \textcolor{nLow}{234.666} & 12.1551 & 2.2062 & 6.2942 & \textcolor{nLow}{103} & 1{,}920 & 1{,}411 & 35{,}478 \\
\midrule
\multirow{10}{*}{\textbf{arXiv}} & $[0.0, 0.1)$ & 0.025 & 0.4159 & 5.7355 & 102.1255 & 3{,}152{,}387 & 1{,}900{,}869 & 160{,}269 & 109{,}227 \\
 & $[0.1, 0.2)$ & 0.035 & 0.2448 & 2.0284 & 35.4463 & 2{,}728{,}095 & 1{,}831{,}867 & 449{,}252 & 353{,}986 \\
 & $[0.2, 0.3)$ & 0.051 & 0.2377 & 1.3667 & 21.9201 & 2{,}269{,}589 & 1{,}745{,}305 & 685{,}259 & 650{,}247 \\
 & $[0.3, 0.4)$ & 0.079 & 0.2625 & 1.1066 & 16.1067 & 1{,}845{,}448 & 1{,}639{,}599 & 866{,}826 & 1{,}007{,}743 \\
 & $[0.4, 0.5)$ & 0.131 & 0.3176 & 0.9943 & 12.9259 & 1{,}433{,}287 & 1{,}511{,}996 & 978{,}438 & 1{,}429{,}239 \\
 & $[0.5, 0.6)$ & 0.243 & 0.4165 & 0.9703 & 10.9009 & 1{,}055{,}985 & 1{,}367{,}986 & 1{,}013{,}330 & 1{,}944{,}331 \\
 & $[0.6, 0.7)$ & 0.523 & 0.6040 & 1.0185 & 9.4821 & 704{,}371 & 1{,}173{,}911 & 945{,}656 & 2{,}531{,}070 \\
 & $[0.7, 0.8)$ & 1.399 & 1.0083 & 1.1704 & 8.3863 & 403{,}960 & 935{,}107 & 780{,}469 & 3{,}219{,}600 \\
 & $[0.8, 0.9)$ & 5.846 & 2.1731 & 1.4860 & 7.4202 & 177{,}842 & 636{,}076 & 558{,}903 & 4{,}022{,}635 \\
 & $[0.9, 1.0]$ & 182.380 & 10.7864 & 2.0129 & 6.1090 & 32{,}987 & 237{,}425 & 315{,}807 & 4{,}752{,}917 \\
\midrule
\multirow{10}{*}{\textbf{PubMed}} & $[0.0, 0.1)$ & 0.024 & 1.0386 & 5.8592 & 102.7561 & 1{,}545{,}685 & 983{,}268 & 80{,}294 & 54{,}689 \\
 & $[0.1, 0.2)$ & 0.034 & 0.3510 & 2.1517 & 35.1877 & 1{,}323{,}358 & 967{,}203 & 221{,}774 & 182{,}321 \\
 & $[0.2, 0.3)$ & 0.050 & 0.3024 & 1.4764 & 21.8310 & 1{,}089{,}726 & 913{,}998 & 332{,}266 & 333{,}066 \\
 & $[0.3, 0.4)$ & 0.079 & 0.3072 & 1.2327 & 16.0828 & 878{,}997 & 872{,}799 & 414{,}117 & 525{,}159 \\
 & $[0.4, 0.5)$ & 0.137 & 0.3471 & 1.1381 & 12.8692 & 671{,}605 & 811{,}982 & 456{,}947 & 756{,}170 \\
 & $[0.5, 0.6)$ & 0.264 & 0.4392 & 1.1460 & 10.8607 & 478{,}143 & 734{,}833 & 455{,}013 & 1{,}024{,}619 \\
 & $[0.6, 0.7)$ & 0.593 & 0.6074 & 1.2299 & 9.4668 & 308{,}135 & 637{,}871 & 410{,}483 & 1{,}340{,}727 \\
 & $[0.7, 0.8)$ & 1.695 & 0.9848 & 1.4401 & 8.4028 & 162{,}859 & 506{,}431 & 317{,}414 & 1{,}682{,}864 \\
 & $[0.8, 0.9)$ & 7.300 & 2.1245 & 1.7702 & 7.4685 & 65{,}295 & 344{,}824 & 222{,}004 & 2{,}086{,}597 \\
 & $[0.9, 1.0]$ & 264.126 & 10.9695 & 2.1681 & 6.1455 & 12{,}086 & 122{,}576 & 129{,}028 & 2{,}403{,}318 \\
\midrule
\multirow{10}{*}{\textbf{WikiText-103}} & $[0.0, 0.1)$ & 0.022 & 0.5311 & \textcolor{nLow}{5.4428} & \textcolor{nLow}{89.7271} & 12{,}942 & 8{,}491 & \textcolor{nLow}{677} & \textcolor{nLow}{418} \\
 & $[0.1, 0.2)$ & 0.031 & 0.1856 & 2.0839 & 31.1121 & 12{,}921 & 8{,}849 & 2{,}213 & 1{,}617 \\
 & $[0.2, 0.3)$ & 0.047 & 0.2213 & 1.4747 & 19.5526 & 8{,}929 & 7{,}163 & 2{,}845 & 2{,}567 \\
 & $[0.3, 0.4)$ & 0.073 & 0.2350 & 1.2366 & 15.0690 & 7{,}891 & 7{,}451 & 3{,}889 & 4{,}321 \\
 & $[0.4, 0.5)$ & 0.144 & 0.2694 & 1.2133 & 11.9555 & 6{,}032 & 7{,}534 & 4{,}427 & 6{,}583 \\
 & $[0.5, 0.6)$ & 0.256 & 0.3623 & 1.2105 & 10.3137 & 4{,}477 & 6{,}851 & 4{,}595 & 9{,}165 \\
 & $[0.6, 0.7)$ & 0.589 & 0.5344 & 1.2951 & 9.0539 & 2{,}404 & 5{,}255 & 3{,}556 & 10{,}801 \\
 & $[0.7, 0.8)$ & 1.543 & 0.9669 & 1.4541 & 8.1912 & 1{,}527 & 4{,}458 & 3{,}286 & 14{,}793 \\
 & $[0.8, 0.9)$ & \textcolor{nLow}{7.320} & 2.1099 & 1.8352 & 7.2770 & \textcolor{nLow}{565} & 2{,}903 & 2{,}257 & 17{,}827 \\
 & $[0.9, 1.0]$ & \textcolor{nLow}{1658.877} & \textcolor{nLow}{10.9192} & 2.1069 & 5.8171 & \textcolor{nLow}{113} & \textcolor{nLow}{985} & 1{,}403 & 20{,}539 \\
\bottomrule
\end{tabular}
\end{table}

\begin{table}[p]
\centering
\definecolor{nClean}{HTML}{0D47A1}\definecolor{nCorrupt}{HTML}{E65100}\definecolor{nCleanBg}{HTML}{EAF1FB}\definecolor{nCorruptBg}{HTML}{FDF0E6}\definecolor{nLow}{HTML}{9E9E9E}
\caption{\textbf{Exact values underlying Figures~\ref{fig:nelbo_four_cases} and~\ref{fig:nelbo_four_cases_appendix} for UDLM.} Mean validation NELBO per token (nats) of the NELBO-only model and the number of tokens behind each mean, by corpus and noise-level bin $1-\alpha_t$ (fraction of corrupted tokens), for \textcolor{nClean}{\textbf{clean}} and \textcolor{nCorrupt}{\textbf{corrupted}} positions where the argmax prediction $\hat{x}_\theta$ equals the clean token $x$ or not. $^{\dagger}$In units of $10^{-3}$ nats. \textcolor{nLow}{Gray}: fewer than 1{,}000 tokens (noisier estimate).}
\label{tab:nelbo_four_cases_udlm}
\scriptsize
\setlength{\tabcolsep}{3.5pt}
\renewcommand{\arraystretch}{1.05}
\begin{tabular}{ll >{\columncolor{nCleanBg}}r>{\columncolor{nCleanBg}}r>{\columncolor{nCorruptBg}}r>{\columncolor{nCorruptBg}}r >{\columncolor{nCleanBg}}r>{\columncolor{nCleanBg}}r>{\columncolor{nCorruptBg}}r>{\columncolor{nCorruptBg}}r}
\toprule
\rowcolor{white}& & \multicolumn{4}{c}{\textbf{Mean NELBO per token (nats)}} & \multicolumn{4}{c}{\textbf{Number of tokens}} \\
\cmidrule(lr){3-6} \cmidrule(lr){7-10}
\rowcolor{white}& & \multicolumn{2}{c}{\textcolor{nClean}{\textbf{Clean}}} & \multicolumn{2}{c}{\textcolor{nCorrupt}{\textbf{Corrupted}}} & \multicolumn{2}{c}{\textcolor{nClean}{\textbf{Clean}}} & \multicolumn{2}{c}{\textcolor{nCorrupt}{\textbf{Corrupted}}} \\
\cmidrule(lr){3-4} \cmidrule(lr){5-6} \cmidrule(lr){7-8} \cmidrule(lr){9-10}
\rowcolor{white}\textbf{Corpus} & $1-\alpha_t$ & $\hat{x}_\theta\!=\!x$$^{\dagger}$ & $\hat{x}_\theta\!\neq\!x$ & $\hat{x}_\theta\!=\!x$ & $\hat{x}_\theta\!\neq\!x$ & $\hat{x}_\theta\!=\!x$ & $\hat{x}_\theta\!\neq\!x$ & $\hat{x}_\theta\!=\!x$ & $\hat{x}_\theta\!\neq\!x$ \\
\midrule
\multirow{10}{*}{\textbf{OpenWebText}} & $[0.0, 0.1)$ & 0.018 & 0.2514 & 4.8082 & 86.1054 & 7{,}101{,}661 & 3{,}543{,}893 & 374{,}566 & 191{,}656 \\
 & $[0.1, 0.2)$ & 0.025 & 0.1482 & 1.8055 & 30.4774 & 6{,}149{,}687 & 3{,}543{,}007 & 1{,}056{,}380 & 651{,}118 \\
 & $[0.2, 0.3)$ & 0.038 & 0.1566 & 1.2567 & 19.2184 & 5{,}063{,}872 & 3{,}423{,}035 & 1{,}601{,}837 & 1{,}223{,}896 \\
 & $[0.3, 0.4)$ & 0.060 & 0.1894 & 1.0628 & 14.4402 & 4{,}078{,}135 & 3{,}314{,}353 & 2{,}011{,}194 & 1{,}960{,}158 \\
 & $[0.4, 0.5)$ & 0.107 & 0.2455 & 0.9950 & 11.7728 & 3{,}103{,}788 & 3{,}131{,}260 & 2{,}238{,}431 & 2{,}858{,}105 \\
 & $[0.5, 0.6)$ & 0.210 & 0.3488 & 1.0089 & 10.0972 & 2{,}221{,}855 & 2{,}908{,}232 & 2{,}271{,}244 & 3{,}990{,}157 \\
 & $[0.6, 0.7)$ & 0.473 & 0.5339 & 1.0817 & 8.9078 & 1{,}424{,}470 & 2{,}541{,}104 & 2{,}084{,}878 & 5{,}258{,}604 \\
 & $[0.7, 0.8)$ & 1.353 & 0.9326 & 1.2354 & 8.0044 & 783{,}849 & 2{,}063{,}379 & 1{,}736{,}773 & 6{,}775{,}231 \\
 & $[0.8, 0.9)$ & 6.160 & 2.0812 & 1.5205 & 7.1903 & 319{,}177 & 1{,}394{,}077 & 1{,}243{,}837 & 8{,}416{,}989 \\
 & $[0.9, 1.0]$ & 604.616 & 11.1240 & 1.9855 & 5.8919 & 68{,}777 & 506{,}349 & 714{,}534 & 10{,}022{,}980 \\
\midrule
\multirow{10}{*}{\textbf{LM1B}} & $[0.0, 0.1)$ & 0.029 & 0.4898 & 7.5498 & 102.6641 & 440{,}923 & 383{,}999 & 22{,}789 & 21{,}153 \\
 & $[0.1, 0.2)$ & 0.042 & 0.3293 & 2.8552 & 35.6002 & 381{,}491 & 387{,}979 & 63{,}564 & 72{,}182 \\
 & $[0.2, 0.3)$ & 0.062 & 0.3297 & 1.9539 & 22.2561 & 297{,}758 & 359{,}300 & 90{,}095 & 127{,}855 \\
 & $[0.3, 0.4)$ & 0.101 & 0.3760 & 1.6471 & 16.5993 & 235{,}784 & 346{,}085 & 109{,}475 & 201{,}072 \\
 & $[0.4, 0.5)$ & 0.179 & 0.4685 & 1.5275 & 13.4078 & 171{,}836 & 321{,}355 & 117{,}685 & 287{,}172 \\
 & $[0.5, 0.6)$ & 0.342 & 0.6299 & 1.5377 & 11.4153 & 116{,}833 & 285{,}343 & 112{,}320 & 378{,}432 \\
 & $[0.6, 0.7)$ & 0.747 & 0.9177 & 1.6264 & 9.9845 & 72{,}231 & 239{,}601 & 97{,}880 & 477{,}584 \\
 & $[0.7, 0.8)$ & 2.023 & 1.4827 & 1.7817 & 8.8357 & 38{,}251 & 184{,}311 & 79{,}517 & 585{,}729 \\
 & $[0.8, 0.9)$ & 8.333 & 2.9082 & 2.0005 & 7.7406 & 15{,}953 & 120{,}730 & 59{,}727 & 710{,}342 \\
 & $[0.9, 1.0]$ & 2399.443 & 12.9781 & 2.2236 & 6.0835 & 3{,}680 & 39{,}901 & 38{,}159 & 792{,}244 \\
\midrule
\multirow{10}{*}{\textbf{AG News}} & $[0.0, 0.1)$ & 0.028 & 0.4651 & 6.9650 & \textcolor{nLow}{106.3791} & 20{,}936 & 15{,}130 & 1{,}025 & \textcolor{nLow}{797} \\
 & $[0.1, 0.2)$ & 0.042 & 0.2893 & 2.5737 & 35.4581 & 20{,}493 & 17{,}257 & 3{,}353 & 3{,}441 \\
 & $[0.2, 0.3)$ & 0.062 & 0.2840 & 1.8510 & 22.5358 & 12{,}831 & 13{,}311 & 3{,}861 & 4{,}813 \\
 & $[0.3, 0.4)$ & 0.104 & 0.3552 & 1.5563 & 17.4600 & 12{,}153 & 14{,}700 & 5{,}420 & 8{,}687 \\
 & $[0.4, 0.5)$ & 0.190 & 0.4509 & 1.4576 & 13.8635 & 8{,}643 & 14{,}521 & 5{,}686 & 13{,}134 \\
 & $[0.5, 0.6)$ & 0.376 & 0.6376 & 1.4925 & 11.8373 & 5{,}894 & 13{,}237 & 5{,}423 & 17{,}942 \\
 & $[0.6, 0.7)$ & 0.847 & 0.9601 & 1.5335 & 10.2963 & 2{,}955 & 10{,}135 & 3{,}859 & 20{,}939 \\
 & $[0.7, 0.8)$ & 2.207 & 1.5690 & 1.6712 & 9.2171 & 1{,}613 & 8{,}456 & 3{,}447 & 26{,}932 \\
 & $[0.8, 0.9)$ & \textcolor{nLow}{7.840} & 3.3168 & 1.8626 & 8.1033 & \textcolor{nLow}{567} & 5{,}359 & 2{,}154 & 31{,}856 \\
 & $[0.9, 1.0]$ & \textcolor{nLow}{173.881} & 14.1470 & 2.0505 & 6.2160 & \textcolor{nLow}{116} & 1{,}852 & 1{,}258 & 35{,}686 \\
\midrule
\multirow{10}{*}{\textbf{arXiv}} & $[0.0, 0.1)$ & 0.031 & 0.3746 & 6.2288 & 105.4303 & 3{,}072{,}423 & 1{,}981{,}318 & 153{,}449 & 116{,}074 \\
 & $[0.1, 0.2)$ & 0.044 & 0.2311 & 2.1870 & 36.6228 & 2{,}627{,}959 & 1{,}932{,}454 & 425{,}086 & 378{,}213 \\
 & $[0.2, 0.3)$ & 0.064 & 0.2361 & 1.4674 & 22.6117 & 2{,}182{,}498 & 1{,}833{,}190 & 649{,}660 & 686{,}076 \\
 & $[0.3, 0.4)$ & 0.100 & 0.2674 & 1.1919 & 16.5478 & 1{,}758{,}235 & 1{,}727{,}158 & 819{,}365 & 1{,}055{,}370 \\
 & $[0.4, 0.5)$ & 0.170 & 0.3318 & 1.0840 & 13.2530 & 1{,}353{,}026 & 1{,}592{,}558 & 917{,}289 & 1{,}490{,}599 \\
 & $[0.5, 0.6)$ & 0.317 & 0.4437 & 1.0612 & 11.1607 & 984{,}752 & 1{,}439{,}683 & 939{,}615 & 2{,}018{,}606 \\
 & $[0.6, 0.7)$ & 0.667 & 0.6482 & 1.1078 & 9.6734 & 653{,}254 & 1{,}225{,}230 & 876{,}791 & 2{,}600{,}245 \\
 & $[0.7, 0.8)$ & 1.782 & 1.0788 & 1.2697 & 8.5224 & 374{,}800 & 964{,}388 & 718{,}299 & 3{,}282{,}161 \\
 & $[0.8, 0.9)$ & 7.503 & 2.2944 & 1.6030 & 7.5019 & 162{,}872 & 651{,}203 & 505{,}555 & 4{,}076{,}850 \\
 & $[0.9, 1.0]$ & 359.492 & 12.4100 & 2.0410 & 6.1071 & 32{,}247 & 238{,}181 & 300{,}010 & 4{,}769{,}210 \\
\midrule
\multirow{10}{*}{\textbf{PubMed}} & $[0.0, 0.1)$ & 0.028 & 1.1884 & 6.5149 & 106.0867 & 1{,}511{,}100 & 1{,}018{,}355 & 76{,}518 & 58{,}475 \\
 & $[0.1, 0.2)$ & 0.040 & 0.3343 & 2.3100 & 36.2489 & 1{,}292{,}426 & 998{,}600 & 212{,}358 & 191{,}784 \\
 & $[0.2, 0.3)$ & 0.060 & 0.2941 & 1.5958 & 22.3527 & 1{,}053{,}738 & 950{,}382 & 315{,}722 & 349{,}726 \\
 & $[0.3, 0.4)$ & 0.097 & 0.3104 & 1.3390 & 16.4529 & 843{,}782 & 909{,}010 & 392{,}601 & 547{,}215 \\
 & $[0.4, 0.5)$ & 0.171 & 0.3555 & 1.2293 & 13.1226 & 638{,}614 & 845{,}229 & 429{,}737 & 783{,}636 \\
 & $[0.5, 0.6)$ & 0.331 & 0.4562 & 1.2390 & 11.0649 & 451{,}682 & 761{,}286 & 423{,}499 & 1{,}056{,}141 \\
 & $[0.6, 0.7)$ & 0.736 & 0.6458 & 1.3387 & 9.6304 & 285{,}946 & 660{,}783 & 373{,}888 & 1{,}378{,}647 \\
 & $[0.7, 0.8)$ & 2.045 & 1.0518 & 1.5358 & 8.5216 & 151{,}044 & 518{,}149 & 291{,}492 & 1{,}708{,}371 \\
 & $[0.8, 0.9)$ & 8.441 & 2.2659 & 1.8391 & 7.5513 & 60{,}669 & 349{,}768 & 205{,}718 & 2{,}104{,}613 \\
 & $[0.9, 1.0]$ & 1073.970 & 12.4168 & 2.2015 & 6.1451 & 11{,}573 & 123{,}055 & 123{,}553 & 2{,}408{,}827 \\
\midrule
\multirow{10}{*}{\textbf{WikiText-103}} & $[0.0, 0.1)$ & 0.024 & 1.0118 & \textcolor{nLow}{5.2412} & \textcolor{nLow}{96.3114} & 13{,}569 & 8{,}338 & \textcolor{nLow}{672} & \textcolor{nLow}{461} \\
 & $[0.1, 0.2)$ & 0.031 & 0.4319 & 1.8248 & 33.2287 & 13{,}189 & 8{,}989 & 2{,}293 & 1{,}641 \\
 & $[0.2, 0.3)$ & 0.052 & 0.3447 & 1.4828 & 21.4422 & 9{,}027 & 7{,}433 & 2{,}874 & 2{,}682 \\
 & $[0.3, 0.4)$ & 0.077 & 0.3383 & 1.3226 & 15.8585 & 7{,}867 & 7{,}516 & 3{,}728 & 4{,}441 \\
 & $[0.4, 0.5)$ & 0.138 & 0.3656 & 1.1833 & 12.3010 & 6{,}283 & 7{,}541 & 4{,}464 & 6{,}800 \\
 & $[0.5, 0.6)$ & 0.261 & 0.4640 & 1.2436 & 10.5837 & 4{,}399 & 6{,}955 & 4{,}379 & 9{,}355 \\
 & $[0.6, 0.7)$ & 0.605 & 0.6751 & 1.3156 & 9.2074 & 2{,}421 & 5{,}446 & 3{,}445 & 11{,}216 \\
 & $[0.7, 0.8)$ & 1.705 & 1.0182 & 1.5106 & 8.3065 & 1{,}403 & 4{,}737 & 3{,}031 & 15{,}405 \\
 & $[0.8, 0.9)$ & \textcolor{nLow}{7.846} & 2.2968 & 1.8117 & 7.3334 & \textcolor{nLow}{590} & 2{,}775 & 2{,}065 & 17{,}610 \\
 & $[0.9, 1.0]$ & \textcolor{nLow}{567.680} & 12.2931 & 2.1372 & 5.8313 & \textcolor{nLow}{117} & 1{,}074 & 1{,}470 & 21{,}403 \\
\bottomrule
\end{tabular}
\end{table}

\begin{table}[p]
\centering
\definecolor{nClean}{HTML}{0D47A1}\definecolor{nCorrupt}{HTML}{E65100}\definecolor{nCleanBg}{HTML}{EAF1FB}\definecolor{nCorruptBg}{HTML}{FDF0E6}\definecolor{nLow}{HTML}{9E9E9E}
\caption{\textbf{Exact values underlying Figures~\ref{fig:nelbo_four_cases} and~\ref{fig:nelbo_four_cases_appendix} for SEDD.} Mean validation NELBO per token (nats) of the NELBO-only model and the number of tokens behind each mean, by corpus and noise-level bin $1-\alpha_t$ (fraction of corrupted tokens), for \textcolor{nClean}{\textbf{clean}} and \textcolor{nCorrupt}{\textbf{corrupted}} positions where the argmax prediction $\hat{x}_\theta$ equals the clean token $x$ or not. $^{\dagger}$In units of $10^{-3}$ nats. \textcolor{nLow}{Gray}: fewer than 1{,}000 tokens (noisier estimate).}
\label{tab:nelbo_four_cases_sedd}
\scriptsize
\setlength{\tabcolsep}{3.5pt}
\renewcommand{\arraystretch}{1.05}
\begin{tabular}{ll >{\columncolor{nCleanBg}}r>{\columncolor{nCleanBg}}r>{\columncolor{nCorruptBg}}r>{\columncolor{nCorruptBg}}r >{\columncolor{nCleanBg}}r>{\columncolor{nCleanBg}}r>{\columncolor{nCorruptBg}}r>{\columncolor{nCorruptBg}}r}
\toprule
\rowcolor{white}& & \multicolumn{4}{c}{\textbf{Mean NELBO per token (nats)}} & \multicolumn{4}{c}{\textbf{Number of tokens}} \\
\cmidrule(lr){3-6} \cmidrule(lr){7-10}
\rowcolor{white}& & \multicolumn{2}{c}{\textcolor{nClean}{\textbf{Clean}}} & \multicolumn{2}{c}{\textcolor{nCorrupt}{\textbf{Corrupted}}} & \multicolumn{2}{c}{\textcolor{nClean}{\textbf{Clean}}} & \multicolumn{2}{c}{\textcolor{nCorrupt}{\textbf{Corrupted}}} \\
\cmidrule(lr){3-4} \cmidrule(lr){5-6} \cmidrule(lr){7-8} \cmidrule(lr){9-10}
\rowcolor{white}\textbf{Corpus} & $1-\alpha_t$ & $\hat{x}_\theta\!=\!x$$^{\dagger}$ & $\hat{x}_\theta\!\neq\!x$ & $\hat{x}_\theta\!=\!x$ & $\hat{x}_\theta\!\neq\!x$ & $\hat{x}_\theta\!=\!x$ & $\hat{x}_\theta\!\neq\!x$ & $\hat{x}_\theta\!=\!x$ & $\hat{x}_\theta\!\neq\!x$ \\
\midrule
\multirow{10}{*}{\textbf{OpenWebText}} & $[0.0, 0.1)$ & 0.006 & 0.1122 & 6.9664 & 60.0927 & 44{,}975{,}769 & 18{,}586{,}155 & 570{,}010 & 382{,}626 \\
 & $[0.1, 0.2)$ & 0.094 & 0.4139 & 5.6180 & 57.0556 & 3{,}958{,}369 & 2{,}032{,}886 & 559{,}789 & 454{,}652 \\
 & $[0.2, 0.3)$ & 0.224 & 0.6183 & 5.8430 & 55.4429 & 2{,}015{,}786 & 1{,}270{,}128 & 539{,}883 & 536{,}443 \\
 & $[0.3, 0.4)$ & 0.441 & 0.8860 & 6.0644 & 53.4810 & 1{,}204{,}759 & 953{,}960 & 513{,}454 & 640{,}979 \\
 & $[0.4, 0.5)$ & 0.823 & 1.2522 & 6.2645 & 50.6722 & 767{,}219 & 773{,}081 & 480{,}113 & 773{,}571 \\
 & $[0.5, 0.6)$ & 1.516 & 1.8145 & 6.5382 & 46.9844 & 510{,}867 & 679{,}067 & 460{,}118 & 990{,}332 \\
 & $[0.6, 0.7)$ & 2.976 & 2.6879 & 6.6226 & 41.8275 & 312{,}820 & 567{,}255 & 405{,}832 & 1{,}228{,}525 \\
 & $[0.7, 0.8)$ & 6.859 & 4.2608 & 6.6664 & 34.8671 & 190{,}432 & 494{,}755 & 372{,}513 & 1{,}691{,}740 \\
 & $[0.8, 0.9)$ & 21.992 & 7.5930 & 6.2161 & 24.8689 & 98{,}726 & 386{,}897 & 324{,}732 & 2{,}500{,}749 \\
 & $[0.9, 1.0]$ & 589.139 & 19.5619 & 1.7468 & 3.1931 & 43{,}632 & 255{,}949 & 866{,}816 & 18{,}896{,}835 \\
\midrule
\multirow{10}{*}{\textbf{LM1B}} & $[0.0, 0.1)$ & 0.008 & 0.0920 & 8.6592 & 63.9557 & 2{,}903{,}919 & 1{,}910{,}925 & 34{,}754 & 37{,}954 \\
 & $[0.1, 0.2)$ & 0.113 & 0.4848 & 7.0151 & 61.0286 & 247{,}147 & 205{,}419 & 32{,}877 & 43{,}453 \\
 & $[0.2, 0.3)$ & 0.263 & 0.7810 & 7.4407 & 59.3964 & 126{,}499 & 130{,}755 & 31{,}949 & 52{,}301 \\
 & $[0.3, 0.4)$ & 0.519 & 1.1916 & 7.7749 & 57.2406 & 69{,}556 & 90{,}792 & 27{,}981 & 57{,}943 \\
 & $[0.4, 0.5)$ & 0.949 & 1.7878 & 8.2122 & 54.4366 & 44{,}642 & 75{,}249 & 25{,}875 & 71{,}322 \\
 & $[0.5, 0.6)$ & 1.751 & 2.5869 & 8.6842 & 50.6192 & 27{,}966 & 62{,}018 & 23{,}535 & 86{,}161 \\
 & $[0.6, 0.7)$ & 3.530 & 3.8091 & 8.8262 & 44.7385 & 15{,}524 & 47{,}176 & 18{,}788 & 98{,}736 \\
 & $[0.7, 0.8)$ & 7.961 & 5.9508 & 8.7397 & 37.1627 & 9{,}609 & 41{,}732 & 17{,}699 & 137{,}808 \\
 & $[0.8, 0.9)$ & 25.439 & 9.6504 & 7.8313 & 26.2638 & 5{,}305 & 33{,}093 & 16{,}183 & 205{,}003 \\
 & $[0.9, 1.0]$ & 697.746 & 22.0292 & 1.5896 & 3.3037 & 2{,}665 & 19{,}929 & 61{,}284 & 1{,}432{,}666 \\
\midrule
\multirow{10}{*}{\textbf{AG News}} & $[0.0, 0.1)$ & 0.010 & 0.1004 & 8.6059 & 66.7374 & 134{,}034 & 87{,}786 & 1{,}657 & 1{,}803 \\
 & $[0.1, 0.2)$ & 0.133 & 0.5284 & 7.5544 & 65.3083 & 9{,}845 & 8{,}651 & 1{,}286 & 1{,}722 \\
 & $[0.2, 0.3)$ & 0.344 & 0.9272 & 7.6416 & 63.9260 & 5{,}775 & 6{,}188 & 1{,}452 & 2{,}457 \\
 & $[0.3, 0.4)$ & 0.689 & 1.3195 & 8.2262 & 62.3019 & 2{,}843 & 4{,}133 & 1{,}190 & 2{,}586 \\
 & $[0.4, 0.5)$ & 1.136 & 1.9244 & 8.5453 & 58.3105 & 2{,}456 & 4{,}718 & 1{,}376 & 4{,}250 \\
 & $[0.5, 0.6)$ & \textcolor{nLow}{2.172} & 3.0250 & \textcolor{nLow}{9.1466} & 54.0694 & \textcolor{nLow}{889} & 2{,}160 & \textcolor{nLow}{662} & 2{,}945 \\
 & $[0.6, 0.7)$ & \textcolor{nLow}{4.283} & 4.2171 & \textcolor{nLow}{8.7848} & 48.3291 & \textcolor{nLow}{833} & 2{,}955 & \textcolor{nLow}{938} & 6{,}026 \\
 & $[0.7, 0.8)$ & \textcolor{nLow}{9.560} & 6.5947 & \textcolor{nLow}{9.1107} & 39.1040 & \textcolor{nLow}{302} & 1{,}809 & \textcolor{nLow}{570} & 6{,}023 \\
 & $[0.8, 0.9)$ & \textcolor{nLow}{27.113} & 11.0171 & \textcolor{nLow}{7.5178} & 27.0388 & \textcolor{nLow}{176} & 1{,}470 & \textcolor{nLow}{523} & 9{,}095 \\
 & $[0.9, 1.0]$ & \textcolor{nLow}{865.570} & \textcolor{nLow}{25.9089} & 1.6462 & 3.2496 & \textcolor{nLow}{72} & \textcolor{nLow}{898} & 1{,}820 & 65{,}818 \\
\midrule
\multirow{10}{*}{\textbf{arXiv}} & $[0.0, 0.1)$ & 0.011 & 0.2776 & 7.1886 & 73.7853 & 19{,}128{,}699 & 10{,}928{,}086 & 246{,}368 & 203{,}855 \\
 & $[0.1, 0.2)$ & 0.174 & 0.7912 & 5.8121 & 67.4313 & 1{,}659{,}101 & 1{,}170{,}999 & 243{,}408 & 236{,}060 \\
 & $[0.2, 0.3)$ & 0.408 & 0.9846 & 5.9938 & 63.4647 & 859{,}485 & 713{,}369 & 238{,}386 & 277{,}208 \\
 & $[0.3, 0.4)$ & 0.784 & 1.2472 & 6.1180 & 59.6908 & 516{,}468 & 498{,}321 & 229{,}582 & 314{,}157 \\
 & $[0.4, 0.5)$ & 1.386 & 1.5850 & 6.2932 & 55.5943 & 339{,}169 & 400{,}115 & 219{,}514 & 381{,}106 \\
 & $[0.5, 0.6)$ & 2.424 & 2.1210 & 6.5232 & 50.5840 & 224{,}746 & 326{,}927 & 206{,}187 & 468{,}380 \\
 & $[0.6, 0.7)$ & 4.390 & 2.9880 & 6.7732 & 44.5527 & 142{,}722 & 267{,}698 & 183{,}365 & 579{,}207 \\
 & $[0.7, 0.8)$ & 9.079 & 4.4954 & 7.0221 & 36.7034 & 91{,}322 & 232{,}920 & 168{,}081 & 807{,}645 \\
 & $[0.8, 0.9)$ & 27.609 & 7.6246 & 6.9697 & 25.8699 & 45{,}721 & 186{,}298 & 138{,}457 & 1{,}213{,}652 \\
 & $[0.9, 1.0]$ & 585.046 & 20.4808 & 1.9786 & 3.3328 & 16{,}723 & 125{,}233 & 345{,}527 & 8{,}985{,}029 \\
\midrule
\multirow{10}{*}{\textbf{PubMed}} & $[0.0, 0.1)$ & 0.010 & 0.3319 & 8.0136 & 74.1592 & 9{,}411{,}118 & 5{,}659{,}156 & 119{,}174 & 107{,}064 \\
 & $[0.1, 0.2)$ & 0.154 & 0.8721 & 6.8068 & 67.6566 & 815{,}313 & 600{,}642 & 115{,}061 & 124{,}792 \\
 & $[0.2, 0.3)$ & 0.368 & 1.0510 & 7.0889 & 63.8145 & 421{,}854 & 373{,}939 & 111{,}687 & 149{,}288 \\
 & $[0.3, 0.4)$ & 0.718 & 1.3161 & 7.4342 & 60.0986 & 245{,}672 & 265{,}166 & 102{,}650 & 171{,}408 \\
 & $[0.4, 0.5)$ & 1.304 & 1.6583 & 7.6368 & 56.0094 & 156{,}019 & 212{,}201 & 95{,}114 & 205{,}338 \\
 & $[0.5, 0.6)$ & 2.378 & 2.1629 & 8.0167 & 51.0325 & 100{,}770 & 175{,}732 & 85{,}598 & 252{,}812 \\
 & $[0.6, 0.7)$ & 4.654 & 3.0120 & 8.2824 & 44.8822 & 62{,}556 & 143{,}790 & 73{,}123 & 311{,}379 \\
 & $[0.7, 0.8)$ & 9.997 & 4.5093 & 8.4930 & 36.9475 & 38{,}043 & 122{,}512 & 63{,}110 & 422{,}479 \\
 & $[0.8, 0.9)$ & 29.820 & 7.9068 & 7.8295 & 26.1446 & 19{,}206 & 97{,}746 & 53{,}520 & 626{,}200 \\
 & $[0.9, 1.0]$ & 645.013 & 21.1488 & 1.6938 & 3.3757 & 7{,}597 & 63{,}835 & 173{,}982 & 4{,}503{,}898 \\
\midrule
\multirow{10}{*}{\textbf{WikiText-103}} & $[0.0, 0.1)$ & 0.008 & 0.1784 & 7.4300 & \textcolor{nLow}{67.1533} & 88{,}611 & 44{,}509 & 1{,}153 & \textcolor{nLow}{895} \\
 & $[0.1, 0.2)$ & 0.100 & 0.6166 & \textcolor{nLow}{5.5952} & \textcolor{nLow}{59.9009} & 7{,}273 & 4{,}184 & \textcolor{nLow}{913} & \textcolor{nLow}{942} \\
 & $[0.2, 0.3)$ & 0.252 & 0.8249 & 6.2021 & 56.4863 & 4{,}077 & 2{,}852 & 1{,}033 & 1{,}254 \\
 & $[0.3, 0.4)$ & 0.584 & 1.1212 & \textcolor{nLow}{7.2267} & 56.0038 & 2{,}007 & 2{,}019 & \textcolor{nLow}{796} & 1{,}322 \\
 & $[0.4, 0.5)$ & 0.909 & 1.5419 & 7.4784 & 53.6313 & 1{,}873 & 2{,}145 & 1{,}027 & 2{,}123 \\
 & $[0.5, 0.6)$ & \textcolor{nLow}{1.816} & 1.7696 & \textcolor{nLow}{7.6558} & 46.8919 & \textcolor{nLow}{741} & 1{,}108 & \textcolor{nLow}{672} & 1{,}575 \\
 & $[0.6, 0.7)$ & \textcolor{nLow}{3.421} & 2.8474 & 8.3333 & 43.8387 & \textcolor{nLow}{921} & 2{,}123 & 1{,}098 & 4{,}562 \\
 & $[0.7, 0.8)$ & \textcolor{nLow}{6.716} & \textcolor{nLow}{4.8043} & \textcolor{nLow}{8.1447} & 36.9778 & \textcolor{nLow}{139} & \textcolor{nLow}{598} & \textcolor{nLow}{358} & 1{,}977 \\
 & $[0.8, 0.9)$ & \textcolor{nLow}{25.539} & \textcolor{nLow}{8.2892} & \textcolor{nLow}{7.7705} & 25.4684 & \textcolor{nLow}{186} & \textcolor{nLow}{835} & \textcolor{nLow}{539} & 5{,}096 \\
 & $[0.9, 1.0]$ & \textcolor{nLow}{414.697} & \textcolor{nLow}{23.3882} & 1.6172 & 3.1123 & \textcolor{nLow}{82} & \textcolor{nLow}{487} & 1{,}903 & 39{,}512 \\
\bottomrule
\end{tabular}
\end{table}

%% file: tables/duo-524B-rand-coruption-exp.tex
% Colors purplereg/purplerow are assumed already defined as in Table~\ref{tab:corruption_exp_generalization}.
\definecolor{bluereg}{HTML}{1F5FA6}
\definecolor{bluerow}{HTML}{EAF1FB}

\begin{table}[t]
\centering
\small
\renewcommand{\arraystretch}{1.05}
\setlength{\tabcolsep}{5pt}
\caption{\textbf{Scaling training data alone does not close the retention gap.} Corrupted-token, clean-token, and overall reconstruction accuracy (\%, averaged over corruption rates $p \in \{10,20,30,40\}\%$) on six evaluation corpora, comparing our from-scratch DUO trained on 20B tokens (NELBO-only, black), the same 20B-token model with \textcolor{purplereg}{\textbf{NELBO + CTR-Reg}} (purple), and the officially released DUO checkpoint trained on \textcolor{bluereg}{\textbf{524B tokens}} (blue, $\sim$26$\times$ more data, native sequence length 1024). The \textbf{bolded} value in each column is the highest of the three settings. Scaling to 524B tokens substantially improves corrupted-token accuracy but leaves clean-token accuracy nearly as low as the 20B baseline, so the clean-minus-corrupted gap does not close with scale; CTR-Reg opens a large gap using $26\times$ less data. See Table~\ref{tab:appendix-524b-percorruption} for the per-corruption-rate breakdown underlying the 524B-token row.}\label{tab:appendix-524b-corruption}
\resizebox{\linewidth}{!}{%
\begin{tabular}{llcccccccc}
\specialrule{1.1pt}{2pt}{2pt}
& & \multicolumn{6}{c}{\textbf{Accuracy (\%) by Evaluation Dataset}} & \\
\cmidrule(lr){3-8}
\textbf{Evaluated On} & \textbf{Setting} & \textbf{OWT} & \textbf{AG News} & \textbf{arXiv} & \textbf{LM1B} & \textbf{PubMed} & \textbf{WT-103} & \textbf{Avg} \\
\specialrule{1.1pt}{2pt}{2pt}

\multirow{3}{*}{\textbf{Corrupted-Tokens}}
 & DUO (20B) & 54.5 & 41.5 & 48.4 & 35.2 & 46.3 & 45.5 & 45.2 \\
 & \cellcolor{purplerow}\textcolor{purplereg}{\textbf{+ CTR-Reg (20B)}} & \cellcolor{purplerow}55.0 & \cellcolor{purplerow}41.8 & \cellcolor{purplerow}49.7 & \cellcolor{purplerow}36.5 & \cellcolor{purplerow}47.7 & \cellcolor{purplerow}47.2 & \cellcolor{purplerow}46.3 \\
 & \cellcolor{bluerow}\textcolor{bluereg}{\textbf{DUO (524B)}} & \cellcolor{bluerow}\textbf{60.7} & \cellcolor{bluerow}\textbf{48.8} & \cellcolor{bluerow}\textbf{54.4} & \cellcolor{bluerow}\textbf{39.9} & \cellcolor{bluerow}\textbf{53.6} & \cellcolor{bluerow}\textbf{52.3} & \cellcolor{bluerow}\textbf{51.6} \\

\specialrule{1.1pt}{2pt}{2pt}

\multirow{3}{*}{\textbf{Clean-Tokens}}
 & DUO (20B) & 53.9 & 40.7 & 52.7 & 36.3 & 50.2 & 49.1 & 47.2 \\
 & \cellcolor{purplerow}\textcolor{purplereg}{\textbf{+ CTR-Reg (20B)}} & \cellcolor{purplerow}\textbf{83.2} & \cellcolor{purplerow}\textbf{72.5} & \cellcolor{purplerow}\textbf{79.6} & \cellcolor{purplerow}\textbf{69.9} & \cellcolor{purplerow}\textbf{78.4} & \cellcolor{purplerow}\textbf{80.8} & \cellcolor{purplerow}\textbf{77.4} \\
 & \cellcolor{bluerow}\textcolor{bluereg}{\textbf{DUO (524B)}} & \cellcolor{bluerow}61.5 & \cellcolor{bluerow}49.3 & \cellcolor{bluerow}55.8 & \cellcolor{bluerow}41.5 & \cellcolor{bluerow}53.7 & \cellcolor{bluerow}54.7 & \cellcolor{bluerow}52.7 \\

\specialrule{1.1pt}{2pt}{2pt}

\multirow{3}{*}{\textbf{Overall}}
 & DUO (20B) & 53.9 & 40.7 & 51.5 & 35.8 & 49.0 & 48.0 & 46.5 \\
 & \cellcolor{purplerow}\textcolor{purplereg}{\textbf{+ CTR-Reg (20B)}} & \cellcolor{purplerow}\textbf{75.9} & \cellcolor{purplerow}\textbf{64.6} & \cellcolor{purplerow}\textbf{71.8} & \cellcolor{purplerow}\textbf{61.4} & \cellcolor{purplerow}\textbf{70.4} & \cellcolor{purplerow}\textbf{72.0} & \cellcolor{purplerow}\textbf{69.3} \\
 & \cellcolor{bluerow}\textcolor{bluereg}{\textbf{DUO (524B)}} & \cellcolor{bluerow}61.1 & \cellcolor{bluerow}48.9 & \cellcolor{bluerow}55.2 & \cellcolor{bluerow}40.9 & \cellcolor{bluerow}53.4 & \cellcolor{bluerow}53.9 & \cellcolor{bluerow}52.2 \\

\specialrule{1.1pt}{2pt}{2pt}
\end{tabular}%
}
\end{table}

%% file: tables/duo-524-ppl-entropy.tex
\providecommand{\entokB}[1]{\textcolor{entokB}{#1}}
\providecommand{\enbadB}[1]{\textcolor{enbadB}{#1}}
\definecolor{entokB}{HTML}{2E7D32}
\definecolor{enbadB}{HTML}{C62828}
\definecolor{labelgray}{HTML}{F2F2F2}

\begin{table}[t]
\centering
\small
\renewcommand{\arraystretch}{1.3}
\setlength{\tabcolsep}{5pt}
\caption{\textbf{Greedy-tail sampling collapses diversity at scale, too.} PPL (GPT2-Large, 1024 generated samples) and generation entropy for the officially released DUO checkpoint (524B training tokens, sequence length 1024, total step budget 512) across greedy-tail step counts $g$. Entropy is colored \entokB{green} when it meets or exceeds the real-OpenWebText reference entropy at sequence length 1024 ($5.40$) and \enbadB{red} when it falls short, following the same convention as Table~\ref{tab:appendix-greedy-tail-ppl} (whose reference entropy, $4.83$, is computed at sequence length 512 and is therefore not the value used here). A red entropy means the corresponding PPL was achieved via degenerate, low-diversity generation rather than genuine quality. The \textbf{bolded} PPL is the lowest across all $g$. Because this checkpoint's native sequence length (1024) differs from our 20B-token models (512), PPL and entropy values here are \emph{not} directly comparable in magnitude to Table~\ref{tab:appendix-greedy-tail-ppl}; only the qualitative trend, and each model's standing relative to its own reference entropy, should be compared.}
\label{tab:appendix-524b-greedy-tail}
\resizebox{\linewidth}{!}{%
\begin{tabular}{l|cccccccccccccc}
\specialrule{1.1pt}{2pt}{2pt}
\cellcolor{labelgray}\textbf{Greedy-Tail Steps ($g$)} & \textbf{0} & \textbf{1} & \textbf{2} & \textbf{3} & \textbf{4} & \textbf{5} & \textbf{6} & \textbf{8} & \textbf{10} & \textbf{12} & \textbf{14} & \textbf{16} & \textbf{18} & \textbf{20} \\
\specialrule{0.9pt}{1pt}{1pt}
\cellcolor{labelgray}\textbf{PPL ($\downarrow$)} & 77.77 & 71.14 & 71.91 & 68.67 & 57.55 & 47.61 & 36.01 & 22.76 & 16.19 & 12.56 & 11.22 & 10.44 & 9.97 & \textbf{9.90} \\
\cmidrule(lr){1-15}
\cellcolor{labelgray}\textbf{Entropy} & \entokB{5.54} & \enbadB{5.21} & \enbadB{4.89} & \enbadB{4.58} & \enbadB{4.26} & \enbadB{3.93} & \enbadB{3.63} & \enbadB{3.07} & \enbadB{2.83} & \enbadB{2.61} & \enbadB{2.49} & \enbadB{2.34} & \enbadB{2.21} & \enbadB{2.14} \\
\specialrule{1.1pt}{2pt}{2pt}
\end{tabular}%
}
\end{table}

%% file: tables/rand_perturb_exp.tex
% Required packages in preamble:
% \usepackage{booktabs}
% \usepackage{multirow}
% \usepackage[table]{xcolor}

\definecolor{purplereg}{HTML}{7A4EAB}  % Deep purple for primary text accents
\definecolor{purplerow}{HTML}{F3EEF9}  % Soft purple tint for CTR-Reg rows

\begin{table}[t]
\centering
\small
\renewcommand{\arraystretch}{1.05}
\setlength{\tabcolsep}{5pt}
\caption{\textbf{Generalization of CTR-Reg across evaluation corpora.} For SEDD, UDLM, and DUO, we compare NELBO-only training (black) to \textcolor{purplereg}{\textbf{NELBO + CTR-Reg}} (purple) on six corpora: OWT, AG News, arXiv, LM1B, PubMed, and WT-103, reporting corrupted-token, clean-token, and overall reconstruction accuracy averaged over $p \in \{10,20,30,40\}\%$; \textbf{Avg} is the mean across all six corpora, and the \textbf{bolded} value in each pair is higher. CTR-Reg leaves corrupted-token accuracy essentially unchanged (0.7 points per cell on average) while substantially improving clean-token and overall accuracy across every dataset and model. All models are trained on 20B OpenWebText tokens (169M parameters, sequence length 512); evaluation uses each corpus's validation split.}\label{tab:corruption_exp_generalization}
\resizebox{\linewidth}{!}{%
\begin{tabular}{lllccccccc}
\specialrule{1.1pt}{2pt}{2pt}
& & & \multicolumn{6}{c}{\textbf{Accuracy (\%) by Evaluation Dataset}} & \\
\cmidrule(lr){4-9}
\textbf{Evaluated On} & \textbf{Model} & \textbf{Loss} & \textbf{OWT} & \textbf{AG News} & \textbf{arXiv} & \textbf{LM1B} & \textbf{PubMed} & \textbf{WT-103} & \textbf{Avg} \\
\specialrule{1.1pt}{2pt}{2pt}

\multirow{6}{*}{\textbf{Corrupted-Tokens}}
 & \multirow{2}{*}{\textbf{DUO}}
   & NELBO & 54.5 & 41.5 & 48.4 & 35.2 & 46.3 & 45.5 & 45.2 \\
 & & \cellcolor{purplerow}\textcolor{purplereg}{\textbf{NELBO + CTR-Reg}} & \cellcolor{purplerow}\textbf{55.0} & \cellcolor{purplerow}\textbf{41.8} & \cellcolor{purplerow}\textbf{49.7} & \cellcolor{purplerow}\textbf{36.5} & \cellcolor{purplerow}\textbf{47.7} & \cellcolor{purplerow}\textbf{47.2} & \cellcolor{purplerow}\textbf{46.3} \\
\cmidrule(lr){2-10}
 & \multirow{2}{*}{\textbf{SEDD}}
   & NELBO & \textbf{47.2} & \textbf{33.6} & \textbf{43.1} & 28.9 & \textbf{39.7} & \textbf{39.0} & \textbf{38.6} \\
 & & \cellcolor{purplerow}\textcolor{purplereg}{\textbf{NELBO + CTR-Reg}} & \cellcolor{purplerow}47.1 & \cellcolor{purplerow}32.6 & \cellcolor{purplerow}42.6 & \cellcolor{purplerow}28.9 & \cellcolor{purplerow}38.5 & \cellcolor{purplerow}38.1 & \cellcolor{purplerow}38.0 \\
\cmidrule(lr){2-10}
 & \multirow{2}{*}{\textbf{UDLM}}
   & NELBO & \textbf{55.3} & \textbf{42.8} & 47.4 & \textbf{38.8} & 45.4 & \textbf{47.9} & \textbf{46.3} \\
 & & \cellcolor{purplerow}\textcolor{purplereg}{\textbf{NELBO + CTR-Reg}} & \cellcolor{purplerow}55.2 & \cellcolor{purplerow}42.6 & \cellcolor{purplerow}\textbf{47.6} & \cellcolor{purplerow}38.0 & \cellcolor{purplerow}\textbf{45.9} & \cellcolor{purplerow}47.4 & \cellcolor{purplerow}46.1 \\

\specialrule{1.1pt}{2pt}{2pt}

\multirow{6}{*}{\textbf{Clean-Tokens}}
 & \multirow{2}{*}{\textbf{DUO}}
   & NELBO & 53.9 & 40.7 & 52.7 & 36.3 & 50.2 & 49.1 & 47.2 \\
 & & \cellcolor{purplerow}\textcolor{purplereg}{\textbf{NELBO + CTR-Reg}} & \cellcolor{purplerow}\textbf{83.2} & \cellcolor{purplerow}\textbf{72.5} & \cellcolor{purplerow}\textbf{79.6} & \cellcolor{purplerow}\textbf{69.9} & \cellcolor{purplerow}\textbf{78.4} & \cellcolor{purplerow}\textbf{80.8} & \cellcolor{purplerow}\textbf{77.4} \\
\cmidrule(lr){2-10}
 & \multirow{2}{*}{\textbf{SEDD}}
   & NELBO & 58.4 & 45.1 & 54.2 & 42.5 & 52.1 & 55.6 & 51.3 \\
 & & \cellcolor{purplerow}\textcolor{purplereg}{\textbf{NELBO + CTR-Reg}} & \cellcolor{purplerow}\textbf{81.2} & \cellcolor{purplerow}\textbf{71.9} & \cellcolor{purplerow}\textbf{80.2} & \cellcolor{purplerow}\textbf{72.5} & \cellcolor{purplerow}\textbf{78.0} & \cellcolor{purplerow}\textbf{79.5} & \cellcolor{purplerow}\textbf{77.2} \\
\cmidrule(lr){2-10}
 & \multirow{2}{*}{\textbf{UDLM}}
   & NELBO & 56.7 & 46.0 & 51.5 & 40.9 & 49.6 & 51.8 & 49.4 \\
 & & \cellcolor{purplerow}\textcolor{purplereg}{\textbf{NELBO + CTR-Reg}} & \cellcolor{purplerow}\textbf{80.1} & \cellcolor{purplerow}\textbf{68.2} & \cellcolor{purplerow}\textbf{73.0} & \cellcolor{purplerow}\textbf{66.2} & \cellcolor{purplerow}\textbf{72.8} & \cellcolor{purplerow}\textbf{76.4} & \cellcolor{purplerow}\textbf{72.8} \\

\specialrule{1.1pt}{2pt}{2pt}

\multirow{6}{*}{\textbf{Overall}}
 & \multirow{2}{*}{\textbf{DUO}}
   & NELBO & 53.9 & 40.7 & 51.5 & 35.8 & 49.0 & 48.0 & 46.5 \\
 & & \cellcolor{purplerow}\textcolor{purplereg}{\textbf{NELBO + CTR-Reg}} & \cellcolor{purplerow}\textbf{75.9} & \cellcolor{purplerow}\textbf{64.6} & \cellcolor{purplerow}\textbf{71.8} & \cellcolor{purplerow}\textbf{61.4} & \cellcolor{purplerow}\textbf{70.4} & \cellcolor{purplerow}\textbf{72.0} & \cellcolor{purplerow}\textbf{69.3} \\
\cmidrule(lr){2-10}
 & \multirow{2}{*}{\textbf{SEDD}}
   & NELBO & 55.6 & 42.3 & 51.3 & 39.3 & 48.9 & 51.4 & 48.1 \\
 & & \cellcolor{purplerow}\textcolor{purplereg}{\textbf{NELBO + CTR-Reg}} & \cellcolor{purplerow}\textbf{72.3} & \cellcolor{purplerow}\textbf{62.1} & \cellcolor{purplerow}\textbf{70.2} & \cellcolor{purplerow}\textbf{61.7} & \cellcolor{purplerow}\textbf{67.9} & \cellcolor{purplerow}\textbf{69.0} & \cellcolor{purplerow}\textbf{67.2} \\
\cmidrule(lr){2-10}
 & \multirow{2}{*}{\textbf{UDLM}}
   & NELBO & 56.1 & 45.0 & 50.3 & 40.3 & 48.4 & 50.7 & 48.5 \\
 & & \cellcolor{purplerow}\textcolor{purplereg}{\textbf{NELBO + CTR-Reg}} & \cellcolor{purplerow}\textbf{73.6} & \cellcolor{purplerow}\textbf{61.9} & \cellcolor{purplerow}\textbf{66.7} & \cellcolor{purplerow}\textbf{59.3} & \cellcolor{purplerow}\textbf{66.0} & \cellcolor{purplerow}\textbf{69.1} & \cellcolor{purplerow}\textbf{66.1} \\

\specialrule{1.1pt}{2pt}{2pt}
\end{tabular}%
}
\end{table}

%% file: tables/ppl-entropy-greedy-step.tex
% Requires: booktabs, multirow, xcolor[table], graphicx
% \newcommand{\entok}[1]{\textcolor{entok}{#1}}   -- entropy >= real-OWT entropy
% \newcommand{\enbad}[1]{\textcolor{enbad}{#1}}   -- entropy <  real-OWT entropy
\definecolor{entok}{HTML}{2E7D32}
\definecolor{enbad}{HTML}{C62828}
\newcommand{\entok}[1]{\textcolor{entok}{#1}}
\newcommand{\enbad}[1]{\textcolor{enbad}{#1}}

\begin{table}[t]
\centering
\small
\renewcommand{\arraystretch}{1.05}
\setlength{\tabcolsep}{5pt}
\caption{\textbf{Exact PPL and entropy values underlying Figure~\ref{fig:greedy-tail-lowstep}.} For each of three uniform-state diffusion language models and each greedy-tail step count $g \in \{0,1,2,3,4,5,6,8,10,12,14,16,18,20\}$ (total step budget fixed at 512), we report perplexity (PPL, computed with GPT2-Large over 1024 generated samples; \textbf{bolded} value is the lower, i.e.\ better, of the two) and generation entropy for the NELBO-only baseline and for NELBO + \textcolor{purplereg}{\textbf{CTR-Reg}}. Entropy values are colored \entok{green} when they meet or exceed the real-OpenWebText reference entropy ($4.83$) and \enbad{red} when they fall short, matching Figure~\ref{fig:greedy-tail-lowstep}; a red entropy means the corresponding PPL was achieved via degenerate, low-diversity generation rather than genuine quality.}
\label{tab:appendix-greedy-tail-ppl}
\resizebox{0.7\linewidth}{!}{%
\begin{tabular}{lccccc}
\specialrule{1.1pt}{2pt}{2pt}
 & & \multicolumn{2}{c}{\textbf{NELBO}} & \multicolumn{2}{c}{\textbf{NELBO + \textcolor{purplereg}{CTR-Reg}}} \\
\cmidrule(lr){3-4} \cmidrule(lr){5-6}
\textbf{Model} & \textbf{Greedy-Tail Steps ($g$)} & \textbf{PPL ($\downarrow$)} & \textbf{Entropy} & \textbf{PPL ($\downarrow$)} & \textbf{Entropy} \\
\specialrule{1.1pt}{2pt}{2pt}
\multirow{14}{*}{\textbf{DUO}}
 & 0 & \textbf{105.37} & \entok{5.18} & \cellcolor{purplerow}105.86 & \cellcolor{purplerow}\entok{5.19} \\
 & 1 & 110.88 & \entok{4.96} & \cellcolor{purplerow}\textbf{64.67} & \cellcolor{purplerow}\entok{5.09} \\
 & 2 & 118.86 & \entok{4.87} & \cellcolor{purplerow}\textbf{56.93} & \cellcolor{purplerow}\entok{5.08} \\
 & 3 & 112.91 & \enbad{4.72} & \cellcolor{purplerow}\textbf{53.32} & \cellcolor{purplerow}\entok{5.06} \\
 & 4 & 102.94 & \enbad{4.57} & \cellcolor{purplerow}\textbf{51.16} & \cellcolor{purplerow}\entok{5.06} \\
 & 5 & 91.46 & \enbad{4.39} & \cellcolor{purplerow}\textbf{49.57} & \cellcolor{purplerow}\entok{5.05} \\
 & 6 & 76.15 & \enbad{4.24} & \cellcolor{purplerow}\textbf{49.17} & \cellcolor{purplerow}\entok{5.05} \\
 & 8 & 55.18 & \enbad{3.94} & \cellcolor{purplerow}\textbf{47.49} & \cellcolor{purplerow}\entok{5.05} \\
 & 10 & \textbf{42.01} & \enbad{3.70} & \cellcolor{purplerow}47.41 & \cellcolor{purplerow}\entok{5.05} \\
 & 12 & \textbf{32.79} & \enbad{3.48} & \cellcolor{purplerow}47.37 & \cellcolor{purplerow}\entok{5.05} \\
 & 14 & \textbf{27.37} & \enbad{3.34} & \cellcolor{purplerow}47.15 & \cellcolor{purplerow}\entok{5.06} \\
 & 16 & \textbf{22.49} & \enbad{3.19} & \cellcolor{purplerow}46.78 & \cellcolor{purplerow}\entok{5.05} \\
 & 18 & \textbf{19.65} & \enbad{3.09} & \cellcolor{purplerow}46.44 & \cellcolor{purplerow}\entok{5.04} \\
 & 20 & \textbf{18.90} & \enbad{3.05} & \cellcolor{purplerow}47.07 & \cellcolor{purplerow}\entok{5.04} \\
\specialrule{1.1pt}{2pt}{2pt}
\multirow{14}{*}{\textbf{UDLM}}
 & 0 & 95.64 & \entok{5.18} & \cellcolor{purplerow}\textbf{93.28} & \cellcolor{purplerow}\entok{5.17} \\
 & 1 & 81.10 & \entok{4.96} & \cellcolor{purplerow}\textbf{58.58} & \cellcolor{purplerow}\entok{5.08} \\
 & 2 & 93.13 & \entok{4.92} & \cellcolor{purplerow}\textbf{51.39} & \cellcolor{purplerow}\entok{5.06} \\
 & 3 & 91.31 & \enbad{4.82} & \cellcolor{purplerow}\textbf{47.74} & \cellcolor{purplerow}\entok{5.05} \\
 & 4 & 91.90 & \enbad{4.76} & \cellcolor{purplerow}\textbf{46.15} & \cellcolor{purplerow}\entok{5.06} \\
 & 5 & 87.46 & \enbad{4.67} & \cellcolor{purplerow}\textbf{44.31} & \cellcolor{purplerow}\entok{5.05} \\
 & 6 & 83.52 & \enbad{4.58} & \cellcolor{purplerow}\textbf{43.93} & \cellcolor{purplerow}\entok{5.04} \\
 & 8 & 73.10 & \enbad{4.40} & \cellcolor{purplerow}\textbf{43.09} & \cellcolor{purplerow}\entok{5.04} \\
 & 10 & 59.01 & \enbad{4.20} & \cellcolor{purplerow}\textbf{41.91} & \cellcolor{purplerow}\entok{5.04} \\
 & 12 & 48.85 & \enbad{4.02} & \cellcolor{purplerow}\textbf{42.96} & \cellcolor{purplerow}\entok{5.04} \\
 & 14 & \textbf{41.34} & \enbad{3.86} & \cellcolor{purplerow}42.60 & \cellcolor{purplerow}\entok{5.05} \\
 & 16 & \textbf{33.10} & \enbad{3.70} & \cellcolor{purplerow}42.38 & \cellcolor{purplerow}\entok{5.04} \\
 & 18 & \textbf{28.73} & \enbad{3.57} & \cellcolor{purplerow}42.02 & \cellcolor{purplerow}\entok{5.04} \\
 & 20 & \textbf{25.67} & \enbad{3.49} & \cellcolor{purplerow}42.61 & \cellcolor{purplerow}\entok{5.04} \\
\specialrule{1.1pt}{2pt}{2pt}
\multirow{14}{*}{\textbf{SEDD}}
 & 0 & \textbf{193.99} & \entok{5.22} & \cellcolor{purplerow}194.59 & \cellcolor{purplerow}\entok{5.24} \\
 & 1 & \textbf{98.52} & \entok{4.91} & \cellcolor{purplerow}111.68 & \cellcolor{purplerow}\entok{5.07} \\
 & 2 & \textbf{90.80} & \enbad{4.80} & \cellcolor{purplerow}97.11 & \cellcolor{purplerow}\entok{5.03} \\
 & 3 & \textbf{79.82} & \enbad{4.68} & \cellcolor{purplerow}90.34 & \cellcolor{purplerow}\entok{5.02} \\
 & 4 & \textbf{72.69} & \enbad{4.58} & \cellcolor{purplerow}89.00 & \cellcolor{purplerow}\entok{5.01} \\
 & 5 & \textbf{65.24} & \enbad{4.46} & \cellcolor{purplerow}87.19 & \cellcolor{purplerow}\entok{5.00} \\
 & 6 & \textbf{59.26} & \enbad{4.35} & \cellcolor{purplerow}86.20 & \cellcolor{purplerow}\entok{5.00} \\
 & 8 & \textbf{51.52} & \enbad{4.16} & \cellcolor{purplerow}85.67 & \cellcolor{purplerow}\entok{5.00} \\
 & 10 & \textbf{48.34} & \enbad{3.95} & \cellcolor{purplerow}84.32 & \cellcolor{purplerow}\entok{5.00} \\
 & 12 & \textbf{46.28} & \enbad{3.72} & \cellcolor{purplerow}84.70 & \cellcolor{purplerow}\entok{5.00} \\
 & 14 & \textbf{44.47} & \enbad{3.46} & \cellcolor{purplerow}82.63 & \cellcolor{purplerow}\entok{4.99} \\
 & 16 & \textbf{43.29} & \enbad{3.18} & \cellcolor{purplerow}85.74 & \cellcolor{purplerow}\entok{5.00} \\
 & 18 & \textbf{41.66} & \enbad{2.98} & \cellcolor{purplerow}85.08 & \cellcolor{purplerow}\entok{5.00} \\
 & 20 & \textbf{41.40} & \enbad{2.78} & \cellcolor{purplerow}84.03 & \cellcolor{purplerow}\entok{5.00} \\
\specialrule{1.1pt}{2pt}{2pt}
\end{tabular}%
}
\end{table}

%% file: tables/step-ablation-full.tex
\begin{table}[t]
\centering
\footnotesize
\renewcommand{\arraystretch}{1.05}
\setlength{\tabcolsep}{3.5pt}
\caption{\textbf{Exact PPL and entropy values underlying Figure~\ref{fig:step_ablation}.} For each of three uniform-state diffusion language models and each total sampling-step budget from 16 to 1024, we report perplexity (PPL, computed with GPT2-Large; \textbf{bolded} value is the lower, i.e.\ better, of the two) and generation entropy for the NELBO-only baseline and for NELBO + \textcolor{purplereg}{\textbf{CTR-Reg}}, under standard ancestral sampling (DUO's default sampler includes one trailing greedy step) and under greedy-tail sampling (final three steps fixed to be greedy). Entropy values are colored \entok{green} when they meet or exceed the real-OpenWebText reference entropy ($4.83$) and \enbad{red} when they fall short. Under greedy-tail sampling, the NELBO-only baseline's lower PPL is consistently paired with red (sub-threshold) entropy, indicating the gain comes from reduced diversity rather than genuine quality; CTR-Reg keeps entropy green while matching or improving PPL.}
\label{tab:appendix-step-ablation}
\resizebox{0.9\linewidth}{!}{%
\begin{tabular}{llcccccccc}
\specialrule{1.1pt}{2pt}{2pt}
 & & \multicolumn{4}{c}{\textbf{Standard Ancestral Sampling}} & \multicolumn{4}{c}{\textbf{Greedy-Tail Sampling (final 3 steps)}} \\
\cmidrule(lr){3-6} \cmidrule(lr){7-10}
 & & \multicolumn{2}{c}{NELBO} & \multicolumn{2}{c}{NELBO + \textcolor{purplereg}{CTR-Reg}} & \multicolumn{2}{c}{NELBO} & \multicolumn{2}{c}{NELBO + \textcolor{purplereg}{CTR-Reg}} \\
\cmidrule(lr){3-4} \cmidrule(lr){5-6} \cmidrule(lr){7-8} \cmidrule(lr){9-10}
\textbf{Model} & \textbf{Steps} & \textbf{PPL} & \textbf{Ent.} & \textbf{PPL} & \textbf{Ent.} & \textbf{PPL} & \textbf{Ent.} & \textbf{PPL} & \textbf{Ent.} \\
\specialrule{1.1pt}{2pt}{2pt}
\multirow{7}{*}{\textbf{DUO}}
 & 16 & 147.50 & \entok{4.94} & \cellcolor{purplerow}\textbf{80.53} & \cellcolor{purplerow}\entok{5.08} & 126.75 & \enbad{4.56} & \cellcolor{purplerow}\textbf{63.63} & \cellcolor{purplerow}\entok{5.04} \\
 & 32 & 124.84 & \entok{4.97} & \cellcolor{purplerow}\textbf{71.23} & \cellcolor{purplerow}\entok{5.09} & 122.84 & \enbad{4.69} & \cellcolor{purplerow}\textbf{56.90} & \cellcolor{purplerow}\entok{5.06} \\
 & 64 & 117.41 & \entok{4.97} & \cellcolor{purplerow}\textbf{67.88} & \cellcolor{purplerow}\entok{5.10} & 119.18 & \enbad{4.71} & \cellcolor{purplerow}\textbf{55.10} & \cellcolor{purplerow}\entok{5.06} \\
 & 128 & 111.73 & \entok{4.97} & \cellcolor{purplerow}\textbf{66.25} & \cellcolor{purplerow}\entok{5.09} & 118.33 & \enbad{4.73} & \cellcolor{purplerow}\textbf{54.03} & \cellcolor{purplerow}\entok{5.07} \\
 & 256 & 111.39 & \entok{4.97} & \cellcolor{purplerow}\textbf{65.27} & \cellcolor{purplerow}\entok{5.10} & 114.16 & \enbad{4.72} & \cellcolor{purplerow}\textbf{52.86} & \cellcolor{purplerow}\entok{5.06} \\
 & 512 & 110.88 & \entok{4.96} & \cellcolor{purplerow}\textbf{64.67} & \cellcolor{purplerow}\entok{5.08} & 112.91 & \enbad{4.72} & \cellcolor{purplerow}\textbf{53.32} & \cellcolor{purplerow}\entok{5.06} \\
 & 1024 & 110.61 & \entok{4.96} & \cellcolor{purplerow}\textbf{64.06} & \cellcolor{purplerow}\entok{5.09} & 112.74 & \enbad{4.72} & \cellcolor{purplerow}\textbf{52.77} & \cellcolor{purplerow}\entok{5.06} \\
\specialrule{1.1pt}{2pt}{2pt}
\multirow{7}{*}{\textbf{UDLM}}
 & 16 & 124.61 & \entok{5.15} & \cellcolor{purplerow}\textbf{124.43} & \cellcolor{purplerow}\entok{5.15} & 105.39 & \enbad{4.68} & \cellcolor{purplerow}\textbf{56.73} & \cellcolor{purplerow}\entok{5.00} \\
 & 32 & 105.20 & \entok{5.16} & \cellcolor{purplerow}\textbf{103.89} & \cellcolor{purplerow}\entok{5.16} & 98.77 & \enbad{4.78} & \cellcolor{purplerow}\textbf{51.42} & \cellcolor{purplerow}\entok{5.03} \\
 & 64 & \textbf{97.16} & \entok{5.16} & \cellcolor{purplerow}97.80 & \cellcolor{purplerow}\entok{5.16} & 95.27 & \enbad{4.81} & \cellcolor{purplerow}\textbf{49.33} & \cellcolor{purplerow}\entok{5.04} \\
 & 128 & \textbf{92.71} & \entok{5.16} & \cellcolor{purplerow}93.71 & \cellcolor{purplerow}\entok{5.16} & 92.24 & \enbad{4.82} & \cellcolor{purplerow}\textbf{48.34} & \cellcolor{purplerow}\entok{5.05} \\
 & 256 & \textbf{93.57} & \entok{5.17} & \cellcolor{purplerow}95.02 & \cellcolor{purplerow}\entok{5.17} & 93.51 & \enbad{4.82} & \cellcolor{purplerow}\textbf{47.89} & \cellcolor{purplerow}\entok{5.06} \\
 & 512 & 95.64 & \entok{5.17} & \cellcolor{purplerow}\textbf{93.28} & \cellcolor{purplerow}\entok{5.18} & 91.31 & \enbad{4.82} & \cellcolor{purplerow}\textbf{47.74} & \cellcolor{purplerow}\entok{5.05} \\
 & 1024 & \textbf{94.98} & \entok{5.17} & \cellcolor{purplerow}95.67 & \cellcolor{purplerow}\entok{5.18} & 91.96 & \enbad{4.82} & \cellcolor{purplerow}\textbf{47.61} & \cellcolor{purplerow}\entok{5.05} \\
\specialrule{1.1pt}{2pt}{2pt}
\multirow{7}{*}{\textbf{SEDD}}
 & 16 & 302.62 & \entok{5.17} & \cellcolor{purplerow}\textbf{292.20} & \cellcolor{purplerow}\entok{5.17} & 77.43 & \enbad{4.00} & \cellcolor{purplerow}\textbf{73.43} & \cellcolor{purplerow}\enbad{4.80} \\
 & 32 & 237.53 & \entok{5.19} & \cellcolor{purplerow}\textbf{234.26} & \cellcolor{purplerow}\entok{5.20} & \textbf{84.05} & \enbad{4.47} & \cellcolor{purplerow}84.85 & \cellcolor{purplerow}\entok{4.89} \\
 & 64 & \textbf{206.40} & \entok{5.20} & \cellcolor{purplerow}206.46 & \cellcolor{purplerow}\entok{5.21} & \textbf{82.15} & \enbad{4.61} & \cellcolor{purplerow}87.01 & \cellcolor{purplerow}\entok{4.96} \\
 & 128 & 201.27 & \entok{5.22} & \cellcolor{purplerow}\textbf{201.23} & \cellcolor{purplerow}\entok{5.22} & \textbf{79.86} & \enbad{4.65} & \cellcolor{purplerow}90.03 & \cellcolor{purplerow}\entok{5.00} \\
 & 256 & 191.58 & \entok{5.21} & \cellcolor{purplerow}\textbf{190.13} & \cellcolor{purplerow}\entok{5.22} & \textbf{79.19} & \enbad{4.66} & \cellcolor{purplerow}89.38 & \cellcolor{purplerow}\entok{5.00} \\
 & 512 & \textbf{193.99} & \entok{5.22} & \cellcolor{purplerow}194.59 & \cellcolor{purplerow}\entok{5.23} & \textbf{79.82} & \enbad{4.68} & \cellcolor{purplerow}90.34 & \cellcolor{purplerow}\entok{5.02} \\
 & 1024 & \textbf{194.34} & \entok{5.23} & \cellcolor{purplerow}198.93 & \cellcolor{purplerow}\entok{5.23} & \textbf{80.83} & \enbad{4.68} & \cellcolor{purplerow}93.38 & \cellcolor{purplerow}\entok{5.03} \\
\specialrule{1.1pt}{2pt}{2pt}
\end{tabular}%
}
\end{table}

%% file: tables/rand_perturb_exp_full_coruption_rate.tex
% Required packages: booktabs, multirow, xcolor[table], graphicx
\begin{table}[t]
\centering
\small
\renewcommand{\arraystretch}{1.05}
\setlength{\tabcolsep}{5pt}
\caption{\textbf{Per-corruption-rate results on AG News.} Full breakdown of Table~\ref{tab:corruption_exp_generalization} by corruption rate $p \in \{10,20,30,40\}\%$ on AG News, for each of three uniform-state diffusion language models, comparing NELBO alone (black) against NELBO + \textcolor{purplereg}{\textbf{CTR-Reg}} (purple). The \textbf{bolded} value in each pair is the higher of the two; \textit{Avg} rows report the corruption-rate-averaged values used in Table~\ref{tab:corruption_exp_generalization}.}
\label{tab:appendix-agnews}
\resizebox{0.7\linewidth}{!}{%
\begin{tabular}{lllccc}
\specialrule{1.1pt}{2pt}{2pt}
\textbf{Model} & \textbf{Loss} & \textbf{Corruption} & \textbf{Corrupted-Token} & \textbf{Clean-Token} & \textbf{Overall} \\
 & & \textbf{Rate ($p$)} & \textbf{Acc.\ (\%)} & \textbf{Acc.\ (\%)} & \textbf{Acc.\ (\%)} \\
\specialrule{1.1pt}{2pt}{2pt}
\multirow{10}{*}{\textbf{DUO}}
 & \multirow{5}{*}{NELBO}
 & 10\% & \textbf{52.6} & 49.1 & 49.5 \\
 &  & 20\% & 45.6 & 44.1 & 44.4 \\
 &  & 30\% & 38.0 & 38.2 & 38.2 \\
 &  & 40\% & 29.6 & 31.4 & 30.7 \\
 & & \textit{Avg} & 41.5 & 40.7 & 40.7 \\
\cmidrule(lr){2-6}
 & \multirow{5}{*}{\textcolor{purplereg}{+ CTR-Reg}}
 & 10\% & \cellcolor{purplerow}\textcolor{purplereg}{52.5} & \cellcolor{purplerow}\textcolor{purplereg}{\textbf{80.3}} & \cellcolor{purplerow}\textcolor{purplereg}{\textbf{77.5}} \\
 &  & 20\% & \cellcolor{purplerow}\textcolor{purplereg}{\textbf{46.0}} & \cellcolor{purplerow}\textcolor{purplereg}{\textbf{75.8}} & \cellcolor{purplerow}\textcolor{purplereg}{\textbf{69.9}} \\
 &  & 30\% & \cellcolor{purplerow}\textcolor{purplereg}{\textbf{38.4}} & \cellcolor{purplerow}\textcolor{purplereg}{\textbf{70.3}} & \cellcolor{purplerow}\textcolor{purplereg}{\textbf{60.7}} \\
 &  & 40\% & \cellcolor{purplerow}\textcolor{purplereg}{\textbf{30.2}} & \cellcolor{purplerow}\textcolor{purplereg}{\textbf{63.5}} & \cellcolor{purplerow}\textcolor{purplereg}{\textbf{50.2}} \\
 & & \textit{Avg} & \cellcolor{purplerow}\textcolor{purplereg}{\textbf{41.8}} & \cellcolor{purplerow}\textcolor{purplereg}{\textbf{72.5}} & \cellcolor{purplerow}\textcolor{purplereg}{\textbf{64.6}} \\
\specialrule{1.1pt}{2pt}{2pt}
\multirow{10}{*}{\textbf{SEDD}}
 & \multirow{5}{*}{NELBO}
 & 10\% & \textbf{43.9} & 56.2 & 54.9 \\
 &  & 20\% & \textbf{37.4} & 49.6 & 47.1 \\
 &  & 30\% & \textbf{30.1} & 41.7 & 38.2 \\
 &  & 40\% & \textbf{22.9} & 33.1 & 29.0 \\
 & & \textit{Avg} & \textbf{33.6} & 45.1 & 42.3 \\
\cmidrule(lr){2-6}
 & \multirow{5}{*}{\textcolor{purplereg}{+ CTR-Reg}}
 & 10\% & \cellcolor{purplerow}\textcolor{purplereg}{43.0} & \cellcolor{purplerow}\textcolor{purplereg}{\textbf{82.4}} & \cellcolor{purplerow}\textcolor{purplereg}{\textbf{78.4}} \\
 &  & 20\% & \cellcolor{purplerow}\textcolor{purplereg}{36.4} & \cellcolor{purplerow}\textcolor{purplereg}{\textbf{75.8}} & \cellcolor{purplerow}\textcolor{purplereg}{\textbf{67.9}} \\
 &  & 30\% & \cellcolor{purplerow}\textcolor{purplereg}{29.1} & \cellcolor{purplerow}\textcolor{purplereg}{\textbf{68.6}} & \cellcolor{purplerow}\textcolor{purplereg}{\textbf{56.7}} \\
 &  & 40\% & \cellcolor{purplerow}\textcolor{purplereg}{22.0} & \cellcolor{purplerow}\textcolor{purplereg}{\textbf{60.8}} & \cellcolor{purplerow}\textcolor{purplereg}{\textbf{45.3}} \\
 & & \textit{Avg} & \cellcolor{purplerow}\textcolor{purplereg}{32.6} & \cellcolor{purplerow}\textcolor{purplereg}{\textbf{71.9}} & \cellcolor{purplerow}\textcolor{purplereg}{\textbf{62.1}} \\
\specialrule{1.1pt}{2pt}{2pt}
\multirow{10}{*}{\textbf{UDLM}}
 & \multirow{5}{*}{NELBO}
 & 10\% & \textbf{52.3} & 53.4 & 53.3 \\
 &  & 20\% & \textbf{46.4} & 49.0 & 48.4 \\
 &  & 30\% & \textbf{39.7} & 43.6 & 42.5 \\
 &  & 40\% & \textbf{32.9} & 37.8 & 35.8 \\
 & & \textit{Avg} & \textbf{42.8} & 46.0 & 45.0 \\
\cmidrule(lr){2-6}
 & \multirow{5}{*}{\textcolor{purplereg}{+ CTR-Reg}}
 & 10\% & \cellcolor{purplerow}\textcolor{purplereg}{51.9} & \cellcolor{purplerow}\textcolor{purplereg}{\textbf{77.6}} & \cellcolor{purplerow}\textcolor{purplereg}{\textbf{75.0}} \\
 &  & 20\% & \cellcolor{purplerow}\textcolor{purplereg}{46.2} & \cellcolor{purplerow}\textcolor{purplereg}{\textbf{72.7}} & \cellcolor{purplerow}\textcolor{purplereg}{\textbf{67.4}} \\
 &  & 30\% & \cellcolor{purplerow}\textcolor{purplereg}{39.5} & \cellcolor{purplerow}\textcolor{purplereg}{\textbf{65.8}} & \cellcolor{purplerow}\textcolor{purplereg}{\textbf{57.9}} \\
 &  & 40\% & \cellcolor{purplerow}\textcolor{purplereg}{32.7} & \cellcolor{purplerow}\textcolor{purplereg}{\textbf{56.7}} & \cellcolor{purplerow}\textcolor{purplereg}{\textbf{47.1}} \\
 & & \textit{Avg} & \cellcolor{purplerow}\textcolor{purplereg}{42.6} & \cellcolor{purplerow}\textcolor{purplereg}{\textbf{68.2}} & \cellcolor{purplerow}\textcolor{purplereg}{\textbf{61.9}} \\
\specialrule{1.1pt}{2pt}{2pt}
\end{tabular}%
}
\end{table}

% Required packages: booktabs, multirow, xcolor[table], graphicx
\begin{table}[t]
\centering
\small
\renewcommand{\arraystretch}{1.05}
\setlength{\tabcolsep}{5pt}
\caption{\textbf{Per-corruption-rate results on arXiv.} Full breakdown of Table~\ref{tab:corruption_exp_generalization} by corruption rate $p \in \{10,20,30,40\}\%$ on arXiv, for each of three uniform-state diffusion language models, comparing NELBO alone (black) against NELBO + \textcolor{purplereg}{\textbf{CTR-Reg}} (purple). The \textbf{bolded} value in each pair is the higher of the two; \textit{Avg} rows report the corruption-rate-averaged values used in Table~\ref{tab:corruption_exp_generalization}.}
\label{tab:appendix-arxiv}
\resizebox{0.7\linewidth}{!}{%
\begin{tabular}{lllccc}
\specialrule{1.1pt}{2pt}{2pt}
\textbf{Model} & \textbf{Loss} & \textbf{Corruption} & \textbf{Corrupted-Token} & \textbf{Clean-Token} & \textbf{Overall} \\
 & & \textbf{Rate ($p$)} & \textbf{Acc.\ (\%)} & \textbf{Acc.\ (\%)} & \textbf{Acc.\ (\%)} \\
\specialrule{1.1pt}{2pt}{2pt}
\multirow{10}{*}{\textbf{DUO}}
 & \multirow{5}{*}{NELBO}
 & 10\% & 57.2 & 59.8 & 59.5 \\
 &  & 20\% & 52.0 & 55.8 & 55.1 \\
 &  & 30\% & 45.9 & 50.8 & 49.3 \\
 &  & 40\% & 38.2 & 44.4 & 41.9 \\
 & & \textit{Avg} & 48.4 & 52.7 & 51.5 \\
\cmidrule(lr){2-6}
 & \multirow{5}{*}{\textcolor{purplereg}{+ CTR-Reg}}
 & 10\% & \cellcolor{purplerow}\textcolor{purplereg}{\textbf{58.2}} & \cellcolor{purplerow}\textcolor{purplereg}{\textbf{84.5}} & \cellcolor{purplerow}\textcolor{purplereg}{\textbf{81.8}} \\
 &  & 20\% & \cellcolor{purplerow}\textcolor{purplereg}{\textbf{53.2}} & \cellcolor{purplerow}\textcolor{purplereg}{\textbf{81.8}} & \cellcolor{purplerow}\textcolor{purplereg}{\textbf{76.1}} \\
 &  & 30\% & \cellcolor{purplerow}\textcolor{purplereg}{\textbf{47.2}} & \cellcolor{purplerow}\textcolor{purplereg}{\textbf{78.4}} & \cellcolor{purplerow}\textcolor{purplereg}{\textbf{69.0}} \\
 &  & 40\% & \cellcolor{purplerow}\textcolor{purplereg}{\textbf{40.0}} & \cellcolor{purplerow}\textcolor{purplereg}{\textbf{73.7}} & \cellcolor{purplerow}\textcolor{purplereg}{\textbf{60.2}} \\
 & & \textit{Avg} & \cellcolor{purplerow}\textcolor{purplereg}{\textbf{49.7}} & \cellcolor{purplerow}\textcolor{purplereg}{\textbf{79.6}} & \cellcolor{purplerow}\textcolor{purplereg}{\textbf{71.8}} \\
\specialrule{1.1pt}{2pt}{2pt}
\multirow{10}{*}{\textbf{SEDD}}
 & \multirow{5}{*}{NELBO}
 & 10\% & \textbf{53.1} & 62.6 & 61.7 \\
 &  & 20\% & \textbf{47.2} & 57.9 & 55.8 \\
 &  & 30\% & \textbf{40.1} & 51.9 & 48.4 \\
 &  & 40\% & \textbf{31.9} & 44.4 & 39.4 \\
 & & \textit{Avg} & \textbf{43.1} & 54.2 & 51.3 \\
\cmidrule(lr){2-6}
 & \multirow{5}{*}{\textcolor{purplereg}{+ CTR-Reg}}
 & 10\% & \cellcolor{purplerow}\textcolor{purplereg}{53.8} & \cellcolor{purplerow}\textcolor{purplereg}{\textbf{87.5}} & \cellcolor{purplerow}\textcolor{purplereg}{\textbf{84.0}} \\
 &  & 20\% & \cellcolor{purplerow}\textcolor{purplereg}{47.0} & \cellcolor{purplerow}\textcolor{purplereg}{\textbf{83.4}} & \cellcolor{purplerow}\textcolor{purplereg}{\textbf{75.9}} \\
 &  & 30\% & \cellcolor{purplerow}\textcolor{purplereg}{39.1} & \cellcolor{purplerow}\textcolor{purplereg}{\textbf{78.2}} & \cellcolor{purplerow}\textcolor{purplereg}{\textbf{66.2}} \\
 &  & 40\% & \cellcolor{purplerow}\textcolor{purplereg}{30.4} & \cellcolor{purplerow}\textcolor{purplereg}{\textbf{71.6}} & \cellcolor{purplerow}\textcolor{purplereg}{\textbf{54.7}} \\
 & & \textit{Avg} & \cellcolor{purplerow}\textcolor{purplereg}{42.6} & \cellcolor{purplerow}\textcolor{purplereg}{\textbf{80.2}} & \cellcolor{purplerow}\textcolor{purplereg}{\textbf{70.2}} \\
\specialrule{1.1pt}{2pt}{2pt}
\multirow{10}{*}{\textbf{UDLM}}
 & \multirow{5}{*}{NELBO}
 & 10\% & 54.2 & 56.7 & 56.4 \\
 &  & 20\% & 50.1 & 53.6 & 52.9 \\
 &  & 30\% & 45.4 & 49.9 & 48.5 \\
 &  & 40\% & \textbf{40.0} & 45.7 & 43.4 \\
 & & \textit{Avg} & 47.4 & 51.5 & 50.3 \\
\cmidrule(lr){2-6}
 & \multirow{5}{*}{\textcolor{purplereg}{+ CTR-Reg}}
 & 10\% & \cellcolor{purplerow}\textcolor{purplereg}{\textbf{54.6}} & \cellcolor{purplerow}\textcolor{purplereg}{\textbf{79.9}} & \cellcolor{purplerow}\textcolor{purplereg}{\textbf{77.4}} \\
 &  & 20\% & \cellcolor{purplerow}\textcolor{purplereg}{\textbf{50.2}} & \cellcolor{purplerow}\textcolor{purplereg}{\textbf{76.2}} & \cellcolor{purplerow}\textcolor{purplereg}{\textbf{71.0}} \\
 &  & 30\% & \cellcolor{purplerow}\textcolor{purplereg}{\textbf{45.4}} & \cellcolor{purplerow}\textcolor{purplereg}{\textbf{71.2}} & \cellcolor{purplerow}\textcolor{purplereg}{\textbf{63.5}} \\
 &  & 40\% & \cellcolor{purplerow}\textcolor{purplereg}{40.0} & \cellcolor{purplerow}\textcolor{purplereg}{\textbf{64.6}} & \cellcolor{purplerow}\textcolor{purplereg}{\textbf{54.8}} \\
 & & \textit{Avg} & \cellcolor{purplerow}\textcolor{purplereg}{\textbf{47.6}} & \cellcolor{purplerow}\textcolor{purplereg}{\textbf{73.0}} & \cellcolor{purplerow}\textcolor{purplereg}{\textbf{66.7}} \\
\specialrule{1.1pt}{2pt}{2pt}
\end{tabular}%
}
\end{table}

% Required packages: booktabs, multirow, xcolor[table], graphicx
\begin{table}[t]
\centering
\small
\renewcommand{\arraystretch}{1.05}
\setlength{\tabcolsep}{5pt}
\caption{\textbf{Per-corruption-rate results on LM1B.} Full breakdown of Table~\ref{tab:corruption_exp_generalization} by corruption rate $p \in \{10,20,30,40\}\%$ on LM1B, for each of three uniform-state diffusion language models, comparing NELBO alone (black) against NELBO + \textcolor{purplereg}{\textbf{CTR-Reg}} (purple). The \textbf{bolded} value in each pair is the higher of the two; \textit{Avg} rows report the corruption-rate-averaged values used in Table~\ref{tab:corruption_exp_generalization}.}
\label{tab:appendix-lm1b}
\resizebox{0.7\linewidth}{!}{%
\begin{tabular}{lllccc}
\specialrule{1.1pt}{2pt}{2pt}
\textbf{Model} & \textbf{Loss} & \textbf{Corruption} & \textbf{Corrupted-Token} & \textbf{Clean-Token} & \textbf{Overall} \\
 & & \textbf{Rate ($p$)} & \textbf{Acc.\ (\%)} & \textbf{Acc.\ (\%)} & \textbf{Acc.\ (\%)} \\
\specialrule{1.1pt}{2pt}{2pt}
\multirow{10}{*}{\textbf{DUO}}
 & \multirow{5}{*}{NELBO}
 & 10\% & 44.9 & 43.8 & 43.9 \\
 &  & 20\% & 38.7 & 39.3 & 39.1 \\
 &  & 30\% & 32.0 & 34.1 & 33.5 \\
 &  & 40\% & 25.0 & 28.2 & 26.9 \\
 & & \textit{Avg} & 35.2 & 36.3 & 35.8 \\
\cmidrule(lr){2-6}
 & \multirow{5}{*}{\textcolor{purplereg}{+ CTR-Reg}}
 & 10\% & \cellcolor{purplerow}\textcolor{purplereg}{\textbf{46.4}} & \cellcolor{purplerow}\textcolor{purplereg}{\textbf{77.7}} & \cellcolor{purplerow}\textcolor{purplereg}{\textbf{74.5}} \\
 &  & 20\% & \cellcolor{purplerow}\textcolor{purplereg}{\textbf{40.1}} & \cellcolor{purplerow}\textcolor{purplereg}{\textbf{73.1}} & \cellcolor{purplerow}\textcolor{purplereg}{\textbf{66.5}} \\
 &  & 30\% & \cellcolor{purplerow}\textcolor{purplereg}{\textbf{33.4}} & \cellcolor{purplerow}\textcolor{purplereg}{\textbf{67.6}} & \cellcolor{purplerow}\textcolor{purplereg}{\textbf{57.4}} \\
 &  & 40\% & \cellcolor{purplerow}\textcolor{purplereg}{\textbf{26.3}} & \cellcolor{purplerow}\textcolor{purplereg}{\textbf{61.0}} & \cellcolor{purplerow}\textcolor{purplereg}{\textbf{47.1}} \\
 & & \textit{Avg} & \cellcolor{purplerow}\textcolor{purplereg}{\textbf{36.5}} & \cellcolor{purplerow}\textcolor{purplereg}{\textbf{69.9}} & \cellcolor{purplerow}\textcolor{purplereg}{\textbf{61.4}} \\
\specialrule{1.1pt}{2pt}{2pt}
\multirow{10}{*}{\textbf{SEDD}}
 & \multirow{5}{*}{NELBO}
 & 10\% & \textbf{38.1} & 53.5 & 52.0 \\
 &  & 20\% & \textbf{32.0} & 46.8 & 43.9 \\
 &  & 30\% & 25.9 & 39.0 & 35.0 \\
 &  & 40\% & 19.8 & 30.6 & 26.3 \\
 & & \textit{Avg} & 28.9 & 42.5 & 39.3 \\
\cmidrule(lr){2-6}
 & \multirow{5}{*}{\textcolor{purplereg}{+ CTR-Reg}}
 & 10\% & \cellcolor{purplerow}\textcolor{purplereg}{37.8} & \cellcolor{purplerow}\textcolor{purplereg}{\textbf{82.7}} & \cellcolor{purplerow}\textcolor{purplereg}{\textbf{78.2}} \\
 &  & 20\% & \cellcolor{purplerow}\textcolor{purplereg}{32.0} & \cellcolor{purplerow}\textcolor{purplereg}{\textbf{76.2}} & \cellcolor{purplerow}\textcolor{purplereg}{\textbf{67.3}} \\
 &  & 30\% & \cellcolor{purplerow}\textcolor{purplereg}{\textbf{25.9}} & \cellcolor{purplerow}\textcolor{purplereg}{\textbf{69.3}} & \cellcolor{purplerow}\textcolor{purplereg}{\textbf{56.3}} \\
 &  & 40\% & \cellcolor{purplerow}\textcolor{purplereg}{\textbf{19.9}} & \cellcolor{purplerow}\textcolor{purplereg}{\textbf{61.9}} & \cellcolor{purplerow}\textcolor{purplereg}{\textbf{45.1}} \\
 & & \textit{Avg} & \cellcolor{purplerow}\textcolor{purplereg}{28.9} & \cellcolor{purplerow}\textcolor{purplereg}{\textbf{72.5}} & \cellcolor{purplerow}\textcolor{purplereg}{\textbf{61.7}} \\
\specialrule{1.1pt}{2pt}{2pt}
\multirow{10}{*}{\textbf{UDLM}}
 & \multirow{5}{*}{NELBO}
 & 10\% & \textbf{47.8} & 48.3 & 48.3 \\
 &  & 20\% & \textbf{42.0} & 43.8 & 43.4 \\
 &  & 30\% & \textbf{35.8} & 38.6 & 37.8 \\
 &  & 40\% & \textbf{29.4} & 32.9 & 31.5 \\
 & & \textit{Avg} & \textbf{38.8} & 40.9 & 40.3 \\
\cmidrule(lr){2-6}
 & \multirow{5}{*}{\textcolor{purplereg}{+ CTR-Reg}}
 & 10\% & \cellcolor{purplerow}\textcolor{purplereg}{47.0} & \cellcolor{purplerow}\textcolor{purplereg}{\textbf{75.6}} & \cellcolor{purplerow}\textcolor{purplereg}{\textbf{72.8}} \\
 &  & 20\% & \cellcolor{purplerow}\textcolor{purplereg}{41.2} & \cellcolor{purplerow}\textcolor{purplereg}{\textbf{70.6}} & \cellcolor{purplerow}\textcolor{purplereg}{\textbf{64.7}} \\
 &  & 30\% & \cellcolor{purplerow}\textcolor{purplereg}{35.0} & \cellcolor{purplerow}\textcolor{purplereg}{\textbf{63.8}} & \cellcolor{purplerow}\textcolor{purplereg}{\textbf{55.2}} \\
 &  & 40\% & \cellcolor{purplerow}\textcolor{purplereg}{28.7} & \cellcolor{purplerow}\textcolor{purplereg}{\textbf{54.8}} & \cellcolor{purplerow}\textcolor{purplereg}{\textbf{44.4}} \\
 & & \textit{Avg} & \cellcolor{purplerow}\textcolor{purplereg}{38.0} & \cellcolor{purplerow}\textcolor{purplereg}{\textbf{66.2}} & \cellcolor{purplerow}\textcolor{purplereg}{\textbf{59.3}} \\
\specialrule{1.1pt}{2pt}{2pt}
\end{tabular}%
}
\end{table}

% Required packages: booktabs, multirow, xcolor[table], graphicx
\begin{table}[t]
\centering
\small
\renewcommand{\arraystretch}{1.05}
\setlength{\tabcolsep}{5pt}
\caption{\textbf{Per-corruption-rate results on PubMed.} Full breakdown of Table~\ref{tab:corruption_exp_generalization} by corruption rate $p \in \{10,20,30,40\}\%$ on PubMed, for each of three uniform-state diffusion language models, comparing NELBO alone (black) against NELBO + \textcolor{purplereg}{\textbf{CTR-Reg}} (purple). The \textbf{bolded} value in each pair is the higher of the two; \textit{Avg} rows report the corruption-rate-averaged values used in Table~\ref{tab:corruption_exp_generalization}.}
\label{tab:appendix-pubmed}
\resizebox{0.7\linewidth}{!}{%
\begin{tabular}{lllccc}
\specialrule{1.1pt}{2pt}{2pt}
\textbf{Model} & \textbf{Loss} & \textbf{Corruption} & \textbf{Corrupted-Token} & \textbf{Clean-Token} & \textbf{Overall} \\
 & & \textbf{Rate ($p$)} & \textbf{Acc.\ (\%)} & \textbf{Acc.\ (\%)} & \textbf{Acc.\ (\%)} \\
\specialrule{1.1pt}{2pt}{2pt}
\multirow{10}{*}{\textbf{DUO}}
 & \multirow{5}{*}{NELBO}
 & 10\% & 56.3 & 58.2 & 58.0 \\
 &  & 20\% & 50.3 & 53.6 & 53.0 \\
 &  & 30\% & 43.4 & 48.0 & 46.6 \\
 &  & 40\% & 35.0 & 40.9 & 38.5 \\
 & & \textit{Avg} & 46.3 & 50.2 & 49.0 \\
\cmidrule(lr){2-6}
 & \multirow{5}{*}{\textcolor{purplereg}{+ CTR-Reg}}
 & 10\% & \cellcolor{purplerow}\textcolor{purplereg}{\textbf{57.3}} & \cellcolor{purplerow}\textcolor{purplereg}{\textbf{83.8}} & \cellcolor{purplerow}\textcolor{purplereg}{\textbf{81.2}} \\
 &  & 20\% & \cellcolor{purplerow}\textcolor{purplereg}{\textbf{51.6}} & \cellcolor{purplerow}\textcolor{purplereg}{\textbf{80.9}} & \cellcolor{purplerow}\textcolor{purplereg}{\textbf{75.1}} \\
 &  & 30\% & \cellcolor{purplerow}\textcolor{purplereg}{\textbf{44.9}} & \cellcolor{purplerow}\textcolor{purplereg}{\textbf{77.1}} & \cellcolor{purplerow}\textcolor{purplereg}{\textbf{67.4}} \\
 &  & 40\% & \cellcolor{purplerow}\textcolor{purplereg}{\textbf{36.9}} & \cellcolor{purplerow}\textcolor{purplereg}{\textbf{71.8}} & \cellcolor{purplerow}\textcolor{purplereg}{\textbf{57.8}} \\
 & & \textit{Avg} & \cellcolor{purplerow}\textcolor{purplereg}{\textbf{47.7}} & \cellcolor{purplerow}\textcolor{purplereg}{\textbf{78.4}} & \cellcolor{purplerow}\textcolor{purplereg}{\textbf{70.4}} \\
\specialrule{1.1pt}{2pt}{2pt}
\multirow{10}{*}{\textbf{SEDD}}
 & \multirow{5}{*}{NELBO}
 & 10\% & \textbf{50.7} & 61.4 & 60.3 \\
 &  & 20\% & \textbf{44.0} & 56.2 & 53.8 \\
 &  & 30\% & \textbf{36.2} & 49.5 & 45.6 \\
 &  & 40\% & \textbf{27.8} & 41.4 & 36.0 \\
 & & \textit{Avg} & \textbf{39.7} & 52.1 & 48.9 \\
\cmidrule(lr){2-6}
 & \multirow{5}{*}{\textcolor{purplereg}{+ CTR-Reg}}
 & 10\% & \cellcolor{purplerow}\textcolor{purplereg}{50.2} & \cellcolor{purplerow}\textcolor{purplereg}{\textbf{86.5}} & \cellcolor{purplerow}\textcolor{purplereg}{\textbf{82.8}} \\
 &  & 20\% & \cellcolor{purplerow}\textcolor{purplereg}{42.9} & \cellcolor{purplerow}\textcolor{purplereg}{\textbf{81.6}} & \cellcolor{purplerow}\textcolor{purplereg}{\textbf{73.9}} \\
 &  & 30\% & \cellcolor{purplerow}\textcolor{purplereg}{34.6} & \cellcolor{purplerow}\textcolor{purplereg}{\textbf{75.7}} & \cellcolor{purplerow}\textcolor{purplereg}{\textbf{63.4}} \\
 &  & 40\% & \cellcolor{purplerow}\textcolor{purplereg}{26.0} & \cellcolor{purplerow}\textcolor{purplereg}{\textbf{68.3}} & \cellcolor{purplerow}\textcolor{purplereg}{\textbf{51.4}} \\
 & & \textit{Avg} & \cellcolor{purplerow}\textcolor{purplereg}{38.5} & \cellcolor{purplerow}\textcolor{purplereg}{\textbf{78.0}} & \cellcolor{purplerow}\textcolor{purplereg}{\textbf{67.9}} \\
\specialrule{1.1pt}{2pt}{2pt}
\multirow{10}{*}{\textbf{UDLM}}
 & \multirow{5}{*}{NELBO}
 & 10\% & 53.3 & 55.8 & 55.5 \\
 &  & 20\% & 48.4 & 52.0 & 51.3 \\
 &  & 30\% & 43.0 & 47.7 & 46.3 \\
 &  & 40\% & 37.0 & 42.7 & 40.4 \\
 & & \textit{Avg} & 45.4 & 49.6 & 48.4 \\
\cmidrule(lr){2-6}
 & \multirow{5}{*}{\textcolor{purplereg}{+ CTR-Reg}}
 & 10\% & \cellcolor{purplerow}\textcolor{purplereg}{\textbf{54.1}} & \cellcolor{purplerow}\textcolor{purplereg}{\textbf{80.4}} & \cellcolor{purplerow}\textcolor{purplereg}{\textbf{77.8}} \\
 &  & 20\% & \cellcolor{purplerow}\textcolor{purplereg}{\textbf{49.0}} & \cellcolor{purplerow}\textcolor{purplereg}{\textbf{76.4}} & \cellcolor{purplerow}\textcolor{purplereg}{\textbf{70.9}} \\
 &  & 30\% & \cellcolor{purplerow}\textcolor{purplereg}{\textbf{43.3}} & \cellcolor{purplerow}\textcolor{purplereg}{\textbf{70.9}} & \cellcolor{purplerow}\textcolor{purplereg}{\textbf{62.6}} \\
 &  & 40\% & \cellcolor{purplerow}\textcolor{purplereg}{\textbf{37.1}} & \cellcolor{purplerow}\textcolor{purplereg}{\textbf{63.4}} & \cellcolor{purplerow}\textcolor{purplereg}{\textbf{52.9}} \\
 & & \textit{Avg} & \cellcolor{purplerow}\textcolor{purplereg}{\textbf{45.9}} & \cellcolor{purplerow}\textcolor{purplereg}{\textbf{72.8}} & \cellcolor{purplerow}\textcolor{purplereg}{\textbf{66.0}} \\
\specialrule{1.1pt}{2pt}{2pt}
\end{tabular}%
}
\end{table}

% Required packages: booktabs, multirow, xcolor[table], graphicx
\begin{table}[t]
\centering
\small
\renewcommand{\arraystretch}{1.05}
\setlength{\tabcolsep}{5pt}
\caption{\textbf{Per-corruption-rate results on WT-103.} Full breakdown of Table~\ref{tab:corruption_exp_generalization} by corruption rate $p \in \{10,20,30,40\}\%$ on WT-103, for each of three uniform-state diffusion language models, comparing NELBO alone (black) against NELBO + \textcolor{purplereg}{\textbf{CTR-Reg}} (purple). The \textbf{bolded} value in each pair is the higher of the two; \textit{Avg} rows report the corruption-rate-averaged values used in Table~\ref{tab:corruption_exp_generalization}.}
\label{tab:appendix-wt103}
\resizebox{0.7\linewidth}{!}{%
\begin{tabular}{lllccc}
\specialrule{1.1pt}{2pt}{2pt}
\textbf{Model} & \textbf{Loss} & \textbf{Corruption} & \textbf{Corrupted-Token} & \textbf{Clean-Token} & \textbf{Overall} \\
 & & \textbf{Rate ($p$)} & \textbf{Acc.\ (\%)} & \textbf{Acc.\ (\%)} & \textbf{Acc.\ (\%)} \\
\specialrule{1.1pt}{2pt}{2pt}
\multirow{10}{*}{\textbf{DUO}}
 & \multirow{5}{*}{NELBO}
 & 10\% & 56.5 & 57.5 & 57.4 \\
 &  & 20\% & 49.8 & 52.9 & 52.3 \\
 &  & 30\% & 42.2 & 46.9 & 45.5 \\
 &  & 40\% & 33.6 & 39.3 & 37.0 \\
 & & \textit{Avg} & 45.5 & 49.1 & 48.0 \\
\cmidrule(lr){2-6}
 & \multirow{5}{*}{\textcolor{purplereg}{+ CTR-Reg}}
 & 10\% & \cellcolor{purplerow}\textcolor{purplereg}{\textbf{57.8}} & \cellcolor{purplerow}\textcolor{purplereg}{\textbf{87.0}} & \cellcolor{purplerow}\textcolor{purplereg}{\textbf{84.1}} \\
 &  & 20\% & \cellcolor{purplerow}\textcolor{purplereg}{\textbf{51.3}} & \cellcolor{purplerow}\textcolor{purplereg}{\textbf{83.6}} & \cellcolor{purplerow}\textcolor{purplereg}{\textbf{77.2}} \\
 &  & 30\% & \cellcolor{purplerow}\textcolor{purplereg}{\textbf{44.1}} & \cellcolor{purplerow}\textcolor{purplereg}{\textbf{79.3}} & \cellcolor{purplerow}\textcolor{purplereg}{\textbf{68.7}} \\
 &  & 40\% & \cellcolor{purplerow}\textcolor{purplereg}{\textbf{35.5}} & \cellcolor{purplerow}\textcolor{purplereg}{\textbf{73.1}} & \cellcolor{purplerow}\textcolor{purplereg}{\textbf{58.1}} \\
 & & \textit{Avg} & \cellcolor{purplerow}\textcolor{purplereg}{\textbf{47.2}} & \cellcolor{purplerow}\textcolor{purplereg}{\textbf{80.8}} & \cellcolor{purplerow}\textcolor{purplereg}{\textbf{72.0}} \\
\specialrule{1.1pt}{2pt}{2pt}
\multirow{10}{*}{\textbf{SEDD}}
 & \multirow{5}{*}{NELBO}
 & 10\% & \textbf{50.2} & 66.2 & 64.6 \\
 &  & 20\% & \textbf{43.5} & 60.2 & 56.9 \\
 &  & 30\% & \textbf{35.3} & 52.5 & 47.4 \\
 &  & 40\% & \textbf{26.8} & 43.4 & 36.8 \\
 & & \textit{Avg} & \textbf{39.0} & 55.6 & 51.4 \\
\cmidrule(lr){2-6}
 & \multirow{5}{*}{\textcolor{purplereg}{+ CTR-Reg}}
 & 10\% & \cellcolor{purplerow}\textcolor{purplereg}{49.2} & \cellcolor{purplerow}\textcolor{purplereg}{\textbf{88.3}} & \cellcolor{purplerow}\textcolor{purplereg}{\textbf{84.4}} \\
 &  & 20\% & \cellcolor{purplerow}\textcolor{purplereg}{42.7} & \cellcolor{purplerow}\textcolor{purplereg}{\textbf{83.1}} & \cellcolor{purplerow}\textcolor{purplereg}{\textbf{75.1}} \\
 &  & 30\% & \cellcolor{purplerow}\textcolor{purplereg}{34.3} & \cellcolor{purplerow}\textcolor{purplereg}{\textbf{77.0}} & \cellcolor{purplerow}\textcolor{purplereg}{\textbf{64.2}} \\
 &  & 40\% & \cellcolor{purplerow}\textcolor{purplereg}{26.0} & \cellcolor{purplerow}\textcolor{purplereg}{\textbf{69.6}} & \cellcolor{purplerow}\textcolor{purplereg}{\textbf{52.3}} \\
 & & \textit{Avg} & \cellcolor{purplerow}\textcolor{purplereg}{38.1} & \cellcolor{purplerow}\textcolor{purplereg}{\textbf{79.5}} & \cellcolor{purplerow}\textcolor{purplereg}{\textbf{69.0}} \\
\specialrule{1.1pt}{2pt}{2pt}
\multirow{10}{*}{\textbf{UDLM}}
 & \multirow{5}{*}{NELBO}
 & 10\% & \textbf{57.1} & 58.9 & 58.7 \\
 &  & 20\% & \textbf{51.2} & 54.7 & 54.0 \\
 &  & 30\% & \textbf{45.0} & 49.8 & 48.4 \\
 &  & 40\% & \textbf{38.3} & 43.7 & 41.5 \\
 & & \textit{Avg} & \textbf{47.9} & 51.8 & 50.7 \\
\cmidrule(lr){2-6}
 & \multirow{5}{*}{\textcolor{purplereg}{+ CTR-Reg}}
 & 10\% & \cellcolor{purplerow}\textcolor{purplereg}{56.6} & \cellcolor{purplerow}\textcolor{purplereg}{\textbf{84.2}} & \cellcolor{purplerow}\textcolor{purplereg}{\textbf{81.5}} \\
 &  & 20\% & \cellcolor{purplerow}\textcolor{purplereg}{50.5} & \cellcolor{purplerow}\textcolor{purplereg}{\textbf{80.3}} & \cellcolor{purplerow}\textcolor{purplereg}{\textbf{74.3}} \\
 &  & 30\% & \cellcolor{purplerow}\textcolor{purplereg}{44.5} & \cellcolor{purplerow}\textcolor{purplereg}{\textbf{74.6}} & \cellcolor{purplerow}\textcolor{purplereg}{\textbf{65.6}} \\
 &  & 40\% & \cellcolor{purplerow}\textcolor{purplereg}{37.8} & \cellcolor{purplerow}\textcolor{purplereg}{\textbf{66.5}} & \cellcolor{purplerow}\textcolor{purplereg}{\textbf{55.0}} \\
 & & \textit{Avg} & \cellcolor{purplerow}\textcolor{purplereg}{47.4} & \cellcolor{purplerow}\textcolor{purplereg}{\textbf{76.4}} & \cellcolor{purplerow}\textcolor{purplereg}{\textbf{69.1}} \\
\specialrule{1.1pt}{2pt}{2pt}
\end{tabular}%
}
\end{table}

%% file: tables/mauve-detail-results.tex
% Required packages: booktabs, multirow, xcolor[table], graphicx
\begin{table}[t]
\centering
\footnotesize
\renewcommand{\arraystretch}{1.05}
\setlength{\tabcolsep}{4pt}
\caption{\textbf{Exact PPL, entropy, and MAUVE values underlying Figures~\ref{fig:mauve-greedy-tail}.} For each of three uniform-state diffusion language models and each greedy-tail step count $g \in \{0,1,3,6,9\}$ (applied after 500 ancestral sampling steps), we report perplexity (PPL, computed with GPT2-Large; \textbf{bolded} value is the lower, i.e.\ better, of the two), generation entropy, and MAUVE score (\textbf{bolded} value is the higher, i.e.\ better, of the two) for the NELBO-only baseline and for NELBO + \textcolor{purplereg}{\textbf{CTR-Reg}}. Entropy values are colored \entok{green} when they meet or exceed the real-OpenWebText reference entropy ($4.83$) and \enbad{red} when they fall short. As $g$ increases, the NELBO-only baseline's entropy drops below threshold and MAUVE collapses even as its PPL appears to improve, indicating reduced diversity rather than genuine quality; CTR-Reg keeps entropy green and MAUVE high while matching or improving PPL.}
\label{tab:appendix-greedy-tail-full}
\resizebox{0.7\linewidth}{!}{%
\begin{tabular}{llcccccc}
\specialrule{1.1pt}{2pt}{2pt}
& & \multicolumn{3}{c}{\textbf{NELBO}} & \multicolumn{3}{c}{\textbf{NELBO + \textcolor{purplereg}{CTR-Reg}}} \\
\cmidrule(lr){3-5} \cmidrule(lr){6-8}
\textbf{Model} & \textbf{$g$} & \textbf{PPL} & \textbf{Ent.} & \textbf{MAUVE} & \textbf{PPL} & \textbf{Ent.} & \textbf{MAUVE} \\
\specialrule{1.1pt}{2pt}{2pt}
\multirow{5}{*}{\textbf{DUO}}
 & 0 & \textbf{104.96} & \entok{5.187} & 0.701 & \cellcolor{purplerow}\textcolor{purplereg}{105.57} & \cellcolor{purplerow}\textcolor{purplereg}{\entok{5.188}} & \cellcolor{purplerow}\textcolor{purplereg}{\textbf{0.713}} \\
 & 1 & 108.47 & \entok{4.950} & 0.414 & \cellcolor{purplerow}\textcolor{purplereg}{\textbf{64.15}} & \cellcolor{purplerow}\textcolor{purplereg}{\entok{5.087}} & \cellcolor{purplerow}\textcolor{purplereg}{\textbf{0.760}} \\
 & 3 & 113.02 & \enbad{4.720} & 0.180 & \cellcolor{purplerow}\textcolor{purplereg}{\textbf{52.70}} & \cellcolor{purplerow}\textcolor{purplereg}{\entok{5.057}} & \cellcolor{purplerow}\textcolor{purplereg}{\textbf{0.804}} \\
 & 6 & 77.57 & \enbad{4.250} & 0.142 & \cellcolor{purplerow}\textcolor{purplereg}{\textbf{48.96}} & \cellcolor{purplerow}\textcolor{purplereg}{\entok{5.054}} & \cellcolor{purplerow}\textcolor{purplereg}{\textbf{0.805}} \\
 & 9 & 48.56 & \enbad{3.816} & 0.166 & \cellcolor{purplerow}\textcolor{purplereg}{\textbf{47.69}} & \cellcolor{purplerow}\textcolor{purplereg}{\entok{5.049}} & \cellcolor{purplerow}\textcolor{purplereg}{\textbf{0.807}} \\
\specialrule{1.1pt}{2pt}{2pt}
\multirow{5}{*}{\textbf{UDLM}}
 & 0 & 94.33 & \entok{5.177} & 0.743 & \cellcolor{purplerow}\textcolor{purplereg}{\textbf{93.57}} & \cellcolor{purplerow}\textcolor{purplereg}{\entok{5.175}} & \cellcolor{purplerow}\textcolor{purplereg}{\textbf{0.744}} \\
 & 1 & 80.90 & \entok{4.960} & 0.470 & \cellcolor{purplerow}\textcolor{purplereg}{\textbf{57.94}} & \cellcolor{purplerow}\textcolor{purplereg}{\entok{5.082}} & \cellcolor{purplerow}\textcolor{purplereg}{\textbf{0.790}} \\
 & 3 & 91.84 & \enbad{4.822} & 0.220 & \cellcolor{purplerow}\textcolor{purplereg}{\textbf{47.38}} & \cellcolor{purplerow}\textcolor{purplereg}{\entok{5.050}} & \cellcolor{purplerow}\textcolor{purplereg}{\textbf{0.809}} \\
 & 6 & 84.36 & \enbad{4.580} & 0.125 & \cellcolor{purplerow}\textcolor{purplereg}{\textbf{44.15}} & \cellcolor{purplerow}\textcolor{purplereg}{\entok{5.050}} & \cellcolor{purplerow}\textcolor{purplereg}{\textbf{0.810}} \\
 & 9 & 65.13 & \enbad{4.290} & 0.141 & \cellcolor{purplerow}\textcolor{purplereg}{\textbf{44.15}} & \cellcolor{purplerow}\textcolor{purplereg}{\entok{5.052}} & \cellcolor{purplerow}\textcolor{purplereg}{\textbf{0.828}} \\
\specialrule{1.1pt}{2pt}{2pt}
\multirow{5}{*}{\textbf{SEDD}}
 & 0 & \textbf{195.09} & \entok{5.228} & \textbf{0.478} & \cellcolor{purplerow}\textcolor{purplereg}{196.24} & \cellcolor{purplerow}\textcolor{purplereg}{\entok{5.235}} & \cellcolor{purplerow}\textcolor{purplereg}{0.451} \\
 & 1 & \textbf{98.37} & \entok{4.915} & 0.266 & \cellcolor{purplerow}\textcolor{purplereg}{111.08} & \cellcolor{purplerow}\textcolor{purplereg}{\entok{5.067}} & \cellcolor{purplerow}\textcolor{purplereg}{\textbf{0.471}} \\
 & 3 & \textbf{79.27} & \enbad{4.673} & 0.155 & \cellcolor{purplerow}\textcolor{purplereg}{90.53} & \cellcolor{purplerow}\textcolor{purplereg}{\entok{5.017}} & \cellcolor{purplerow}\textcolor{purplereg}{\textbf{0.540}} \\
 & 6 & \textbf{60.40} & \enbad{4.362} & 0.146 & \cellcolor{purplerow}\textcolor{purplereg}{86.58} & \cellcolor{purplerow}\textcolor{purplereg}{\entok{5.007}} & \cellcolor{purplerow}\textcolor{purplereg}{\textbf{0.560}} \\
 & 9 & \textbf{48.95} & \enbad{4.055} & 0.152 & \cellcolor{purplerow}\textcolor{purplereg}{84.56} & \cellcolor{purplerow}\textcolor{purplereg}{\entok{5.000}} & \cellcolor{purplerow}\textcolor{purplereg}{\textbf{0.576}} \\
\specialrule{1.1pt}{2pt}{2pt}
\end{tabular}%
}
\end{table}

%% file: tables/duo-524-corruption-rate-detail.tex
% Required packages: booktabs, multirow, xcolor[table], graphicx
% Colors assumed already defined from the main 524B corruption table:
% \definecolor{bluereg}{HTML}{1F5FA6}
% \definecolor{bluerow}{HTML}{EAF1FB}

\begin{table}[t]
\centering
\small
\renewcommand{\arraystretch}{1.05}
\setlength{\tabcolsep}{6pt}
\caption{\textbf{Per-corruption-rate results for the official DUO checkpoint (524B tokens).} Full breakdown of the \textcolor{bluereg}{\textbf{524B-token DUO}} row of Table~\ref{tab:appendix-524b-corruption} by corruption rate $p \in \{10,20,30,40\}\%$, across all six evaluation corpora. In each row, the \textbf{bolded} value is the higher of corrupted-token vs.\ clean-token accuracy; \textit{Avg} rows report the corruption-rate-averaged values used in Table~\ref{tab:appendix-524b-corruption}.}
\label{tab:appendix-524b-percorruption}
\resizebox{0.75\linewidth}{!}{%
\begin{tabular}{llccc}
\specialrule{1.1pt}{2pt}{2pt}
\textbf{Dataset} & \textbf{Corruption Rate ($p$)} & \textbf{Corrupted-Token Acc.\ (\%)} & \textbf{Clean-Token Acc.\ (\%)} & \textbf{Overall Acc.\ (\%)} \\
\specialrule{1.1pt}{2pt}{2pt}

\multirow{5}{*}{\cellcolor{bluerow}\textcolor{bluereg}{\textbf{OWT}}}
 & \cellcolor{bluerow}10\% & \cellcolor{bluerow}\textbf{71.0} & \cellcolor{bluerow}69.6 & \cellcolor{bluerow}69.7 \\
 & \cellcolor{bluerow}20\% & \cellcolor{bluerow}\textbf{65.5} & \cellcolor{bluerow}65.2 & \cellcolor{bluerow}65.3 \\
 & \cellcolor{bluerow}30\% & \cellcolor{bluerow}58.3 & \cellcolor{bluerow}\textbf{59.7} & \cellcolor{bluerow}59.3 \\
 & \cellcolor{bluerow}40\% & \cellcolor{bluerow}48.1 & \cellcolor{bluerow}\textbf{51.4} & \cellcolor{bluerow}50.1 \\
 & \cellcolor{bluerow}\textit{Avg} & \cellcolor{bluerow}60.7 & \cellcolor{bluerow}\textbf{61.5} & \cellcolor{bluerow}61.1 \\
\specialrule{1.1pt}{2pt}{2pt}

\multirow{5}{*}{\textbf{AG News}}
 & 10\% & \textbf{61.8} & 58.7 & 59.0 \\
 & 20\% & \textbf{54.4} & 53.2 & 53.4 \\
 & 30\% & 45.2 & \textbf{46.8} & 46.3 \\
 & 40\% & 34.0 & \textbf{38.4} & 36.7 \\
 & \textit{Avg} & 48.8 & \textbf{49.3} & 48.9 \\
\specialrule{1.1pt}{2pt}{2pt}

\multirow{5}{*}{\cellcolor{bluerow}\textcolor{bluereg}{\textbf{arXiv}}}
 & \cellcolor{bluerow}10\% & \cellcolor{bluerow}\textbf{65.2} & \cellcolor{bluerow}63.7 & \cellcolor{bluerow}63.8 \\
 & \cellcolor{bluerow}20\% & \cellcolor{bluerow}59.6 & \cellcolor{bluerow}59.6 & \cellcolor{bluerow}59.6 \\
 & \cellcolor{bluerow}30\% & \cellcolor{bluerow}51.9 & \cellcolor{bluerow}\textbf{54.1} & \cellcolor{bluerow}53.5 \\
 & \cellcolor{bluerow}40\% & \cellcolor{bluerow}41.0 & \cellcolor{bluerow}\textbf{45.8} & \cellcolor{bluerow}43.9 \\
 & \cellcolor{bluerow}\textit{Avg} & \cellcolor{bluerow}54.4 & \cellcolor{bluerow}\textbf{55.8} & \cellcolor{bluerow}55.2 \\
\specialrule{1.1pt}{2pt}{2pt}

\multirow{5}{*}{\textbf{LM1B}}
 & 10\% & \textbf{51.6} & 50.7 & 50.8 \\
 & 20\% & 44.5 & \textbf{45.4} & 45.3 \\
 & 30\% & 36.3 & \textbf{38.9} & 38.2 \\
 & 40\% & 27.0 & \textbf{30.8} & 29.3 \\
 & \textit{Avg} & 39.9 & \textbf{41.5} & 40.9 \\
\specialrule{1.1pt}{2pt}{2pt}

\multirow{5}{*}{\cellcolor{bluerow}\textcolor{bluereg}{\textbf{PubMed}}}
 & \cellcolor{bluerow}10\% & \cellcolor{bluerow}\textbf{65.6} & \cellcolor{bluerow}62.5 & \cellcolor{bluerow}62.8 \\
 & \cellcolor{bluerow}20\% & \cellcolor{bluerow}\textbf{59.3} & \cellcolor{bluerow}58.0 & \cellcolor{bluerow}58.2 \\
 & \cellcolor{bluerow}30\% & \cellcolor{bluerow}50.8 & \cellcolor{bluerow}\textbf{51.8} & \cellcolor{bluerow}51.5 \\
 & \cellcolor{bluerow}40\% & \cellcolor{bluerow}38.6 & \cellcolor{bluerow}\textbf{42.4} & \cellcolor{bluerow}40.9 \\
 & \cellcolor{bluerow}\textit{Avg} & \cellcolor{bluerow}53.6 & \cellcolor{bluerow}\textbf{53.7} & \cellcolor{bluerow}53.4 \\
\specialrule{1.1pt}{2pt}{2pt}

\multirow{5}{*}{\textbf{WT-103}}
 & 10\% & \textbf{64.9} & 64.5 & 64.6 \\
 & 20\% & 58.6 & \textbf{60.0} & 59.7 \\
 & 30\% & 49.6 & \textbf{53.0} & 51.9 \\
 & 40\% & 36.1 & \textbf{41.3} & 39.2 \\
 & \textit{Avg} & 52.3 & \textbf{54.7} & 53.9 \\
\specialrule{1.1pt}{2pt}{2pt}
\end{tabular}%
}
\end{table}

%% file: tables/lambda_ablation_ppl_entropy_detail.tex
% Requires: booktabs, multirow, xcolor[table], graphicx
% Colors assumed already defined elsewhere (entok/enbad, purplereg/purplerow, bluereg/bluerow)
\begin{table}[t]
\centering
\small
\renewcommand{\arraystretch}{1.05}
\setlength{\tabcolsep}{4pt}
\caption{\textbf{Exact PPL and entropy values underlying Figure~\ref{fig:lambda-greedy-tail}.} For DUO and UDLM and each greedy-tail step count $g \in \{0,1,2,3,4,5,6,8,10,12,14,16,18,20\}$ (total step budget fixed at 512), we report perplexity (PPL, GPT2-Large, 1024 generated samples; \textbf{bolded} value is the lowest of the three) and generation entropy for $\lambda=0$ (NELBO-only), $\lambda=0.02$ (\textcolor{purplereg}{\textbf{our default}}), and $\lambda=0.1$ (\textcolor{bluereg}{\textbf{blue}}). Entropy values are colored \entok{green} when they meet or exceed the real-OpenWebText reference entropy ($4.83$) and \enbad{red} when they fall short, matching Figure~\ref{fig:lambda-greedy-tail}.}
\label{tab:appendix-lambda-greedy-tail-values}
\resizebox{\linewidth}{!}{%
\begin{tabular}{lccccccc}
\specialrule{1.1pt}{2pt}{2pt}
 & & \multicolumn{2}{c}{\textbf{$\lambda=0$ (NELBO)}} & \multicolumn{2}{c}{\textbf{\textcolor{purplereg}{$\lambda=0.02$}}} & \multicolumn{2}{c}{\textbf{\textcolor{bluereg}{$\lambda=0.1$}}} \\
\cmidrule(lr){3-4} \cmidrule(lr){5-6} \cmidrule(lr){7-8}
\textbf{Model} & \textbf{Greedy-Tail Steps ($g$)} & \textbf{PPL ($\downarrow$)} & \textbf{Entropy} & \textbf{PPL ($\downarrow$)} & \textbf{Entropy} & \textbf{PPL ($\downarrow$)} & \textbf{Entropy} \\
\specialrule{1.1pt}{2pt}{2pt}
\multirow{14}{*}{\textbf{DUO}}
 & 0  & \textbf{105.37} & \entok{5.18} & \cellcolor{purplerow}105.86 & \cellcolor{purplerow}\entok{5.19} & \cellcolor{bluerow}109.03 & \cellcolor{bluerow}\entok{5.19} \\
 & 1  & 110.88 & \entok{4.96} & \cellcolor{purplerow}\textbf{64.67} & \cellcolor{purplerow}\entok{5.09} & \cellcolor{bluerow}73.24 & \cellcolor{bluerow}\entok{5.13} \\
 & 2  & 118.86 & \entok{4.87} & \cellcolor{purplerow}\textbf{56.93} & \cellcolor{purplerow}\entok{5.08} & \cellcolor{bluerow}67.82 & \cellcolor{bluerow}\entok{5.11} \\
 & 3  & 112.91 & \enbad{4.72} & \cellcolor{purplerow}\textbf{53.32} & \cellcolor{purplerow}\entok{5.06} & \cellcolor{bluerow}64.56 & \cellcolor{bluerow}\entok{5.11} \\
 & 4  & 102.94 & \enbad{4.57} & \cellcolor{purplerow}\textbf{51.16} & \cellcolor{purplerow}\entok{5.06} & \cellcolor{bluerow}63.52 & \cellcolor{bluerow}\entok{5.10} \\
 & 5  & 91.46  & \enbad{4.39} & \cellcolor{purplerow}\textbf{49.57} & \cellcolor{purplerow}\entok{5.05} & \cellcolor{bluerow}63.26 & \cellcolor{bluerow}\entok{5.12} \\
 & 6  & 76.15  & \enbad{4.24} & \cellcolor{purplerow}\textbf{49.17} & \cellcolor{purplerow}\entok{5.05} & \cellcolor{bluerow}64.81 & \cellcolor{bluerow}\entok{5.12} \\
 & 8  & 55.18  & \enbad{3.94} & \cellcolor{purplerow}\textbf{47.49} & \cellcolor{purplerow}\entok{5.05} & \cellcolor{bluerow}61.84 & \cellcolor{bluerow}\entok{5.11} \\
 & 10 & \textbf{42.01} & \enbad{3.70} & \cellcolor{purplerow}47.41 & \cellcolor{purplerow}\entok{5.05} & \cellcolor{bluerow}63.19 & \cellcolor{bluerow}\entok{5.11} \\
 & 12 & \textbf{32.79} & \enbad{3.48} & \cellcolor{purplerow}47.37 & \cellcolor{purplerow}\entok{5.05} & \cellcolor{bluerow}62.77 & \cellcolor{bluerow}\entok{5.13} \\
 & 14 & \textbf{27.37} & \enbad{3.34} & \cellcolor{purplerow}47.15 & \cellcolor{purplerow}\entok{5.06} & \cellcolor{bluerow}61.12 & \cellcolor{bluerow}\entok{5.11} \\
 & 16 & \textbf{22.49} & \enbad{3.19} & \cellcolor{purplerow}46.78 & \cellcolor{purplerow}\entok{5.05} & \cellcolor{bluerow}62.65 & \cellcolor{bluerow}\entok{5.11} \\
 & 18 & \textbf{19.65} & \enbad{3.09} & \cellcolor{purplerow}46.44 & \cellcolor{purplerow}\entok{5.04} & \cellcolor{bluerow}63.03 & \cellcolor{bluerow}\entok{5.11} \\
 & 20 & \textbf{18.90} & \enbad{3.05} & \cellcolor{purplerow}47.07 & \cellcolor{purplerow}\entok{5.04} & \cellcolor{bluerow}64.32 & \cellcolor{bluerow}\entok{5.13} \\
\specialrule{1.1pt}{2pt}{2pt}
\multirow{14}{*}{\textbf{UDLM}}
 & 0  & 95.64 & \entok{5.18} & \cellcolor{purplerow}\textbf{93.28} & \cellcolor{purplerow}\entok{5.17} & \cellcolor{bluerow}95.91 & \cellcolor{bluerow}\entok{5.18} \\
 & 1  & 81.10 & \entok{4.96} & \cellcolor{purplerow}\textbf{58.58} & \cellcolor{purplerow}\entok{5.08} & \cellcolor{bluerow}67.42 & \cellcolor{bluerow}\entok{5.12} \\
 & 2  & 93.13 & \entok{4.92} & \cellcolor{purplerow}\textbf{51.39} & \cellcolor{purplerow}\entok{5.06} & \cellcolor{bluerow}62.16 & \cellcolor{bluerow}\entok{5.12} \\
 & 3  & 91.31 & \enbad{4.82} & \cellcolor{purplerow}\textbf{47.74} & \cellcolor{purplerow}\entok{5.05} & \cellcolor{bluerow}58.97 & \cellcolor{bluerow}\entok{5.11} \\
 & 4  & 91.90 & \enbad{4.76} & \cellcolor{purplerow}\textbf{46.15} & \cellcolor{purplerow}\entok{5.06} & \cellcolor{bluerow}58.09 & \cellcolor{bluerow}\entok{5.11} \\
 & 5  & 87.46 & \enbad{4.67} & \cellcolor{purplerow}\textbf{44.31} & \cellcolor{purplerow}\entok{5.05} & \cellcolor{bluerow}57.59 & \cellcolor{bluerow}\entok{5.11} \\
 & 6  & 83.52 & \enbad{4.58} & \cellcolor{purplerow}\textbf{43.93} & \cellcolor{purplerow}\entok{5.04} & \cellcolor{bluerow}57.75 & \cellcolor{bluerow}\entok{5.11} \\
 & 8  & 73.10 & \enbad{4.40} & \cellcolor{purplerow}\textbf{43.09} & \cellcolor{purplerow}\entok{5.04} & \cellcolor{bluerow}58.05 & \cellcolor{bluerow}\entok{5.12} \\
 & 10 & 59.01 & \enbad{4.20} & \cellcolor{purplerow}\textbf{41.91} & \cellcolor{purplerow}\entok{5.04} & \cellcolor{bluerow}57.35 & \cellcolor{bluerow}\entok{5.11} \\
 & 12 & 48.85 & \enbad{4.02} & \cellcolor{purplerow}\textbf{42.96} & \cellcolor{purplerow}\entok{5.04} & \cellcolor{bluerow}57.53 & \cellcolor{bluerow}\entok{5.11} \\
 & 14 & \textbf{41.34} & \enbad{3.86} & \cellcolor{purplerow}42.60 & \cellcolor{purplerow}\entok{5.05} & \cellcolor{bluerow}57.74 & \cellcolor{bluerow}\entok{5.11} \\
 & 16 & \textbf{33.10} & \enbad{3.70} & \cellcolor{purplerow}42.38 & \cellcolor{purplerow}\entok{5.04} & \cellcolor{bluerow}57.96 & \cellcolor{bluerow}\entok{5.11} \\
 & 18 & \textbf{28.73} & \enbad{3.57} & \cellcolor{purplerow}42.02 & \cellcolor{purplerow}\entok{5.04} & \cellcolor{bluerow}57.60 & \cellcolor{bluerow}\entok{5.11} \\
 & 20 & \textbf{25.67} & \enbad{3.49} & \cellcolor{purplerow}42.61 & \cellcolor{purplerow}\entok{5.04} & \cellcolor{bluerow}58.35 & \cellcolor{bluerow}\entok{5.11} \\
\specialrule{1.1pt}{2pt}{2pt}
\end{tabular}%
}
\end{table}

%% file: tables/lamda-rand-corruption-detail.tex
% Add once if not already defined elsewhere in this file's compile order
\makeatletter
\@ifundefined{bluereg}{\definecolor{bluereg}{HTML}{1F5FA6}}{}
\@ifundefined{bluerow}{\definecolor{bluerow}{HTML}{EAF1FB}}{}
\makeatother

% Required packages: booktabs, multirow, xcolor[table], graphicx

\begin{table}[t]
\centering
\small
\renewcommand{\arraystretch}{1.05}
\setlength{\tabcolsep}{5pt}
\caption{\textbf{Per-corruption-rate results at $\lambda=0.1$ (\textcolor{bluereg}{\textbf{blue}}).} Full breakdown of the $\lambda=0.1$ row of Table~\ref{tab:appendix-lambda-corruption} by corruption rate $p \in \{10,20,30,40\}\%$, across all six evaluation corpora, for DUO and UDLM. \textit{Avg} rows report the corruption-rate-averaged values used there.}
\label{tab:appendix-lambda01-percorruption}
\resizebox{0.78\linewidth}{!}{%
\begin{tabular}{lllccc}
\specialrule{1.1pt}{2pt}{2pt}
\textbf{Dataset} & \textbf{Model} & \textbf{Corruption Rate ($p$)} & \textbf{Corrupted-Token Acc.\ (\%)} & \textbf{Clean-Token Acc.\ (\%)} & \textbf{Overall Acc.\ (\%)} \\
\specialrule{1.1pt}{2pt}{2pt}

\multirow{10}{*}{\cellcolor{bluerow}\textcolor{bluereg}{\textbf{AG News}}}
 & \multirow{5}{*}{\cellcolor{bluerow}\textcolor{bluereg}{\textbf{DUO}}}
 & \cellcolor{bluerow}10\% & \cellcolor{bluerow}52.6 & \cellcolor{bluerow}89.7 & \cellcolor{bluerow}86.0 \\
 & & \cellcolor{bluerow}20\% & \cellcolor{bluerow}45.9 & \cellcolor{bluerow}86.8 & \cellcolor{bluerow}78.6 \\
 & & \cellcolor{bluerow}30\% & \cellcolor{bluerow}38.3 & \cellcolor{bluerow}82.9 & \cellcolor{bluerow}69.5 \\
 & & \cellcolor{bluerow}40\% & \cellcolor{bluerow}29.9 & \cellcolor{bluerow}78.0 & \cellcolor{bluerow}58.7 \\
 & & \cellcolor{bluerow}\textit{Avg} & \cellcolor{bluerow}41.7 & \cellcolor{bluerow}84.3 & \cellcolor{bluerow}73.2 \\
\cmidrule(lr){2-6}
 & \multirow{5}{*}{\cellcolor{bluerow}\textcolor{bluereg}{\textbf{UDLM}}}
 & \cellcolor{bluerow}10\% & \cellcolor{bluerow}53.2 & \cellcolor{bluerow}88.2 & \cellcolor{bluerow}84.7 \\
 & & \cellcolor{bluerow}20\% & \cellcolor{bluerow}47.3 & \cellcolor{bluerow}84.4 & \cellcolor{bluerow}77.0 \\
 & & \cellcolor{bluerow}30\% & \cellcolor{bluerow}40.5 & \cellcolor{bluerow}79.2 & \cellcolor{bluerow}67.6 \\
 & & \cellcolor{bluerow}40\% & \cellcolor{bluerow}33.5 & \cellcolor{bluerow}72.4 & \cellcolor{bluerow}56.8 \\
 & & \cellcolor{bluerow}\textit{Avg} & \cellcolor{bluerow}43.7 & \cellcolor{bluerow}81.0 & \cellcolor{bluerow}71.5 \\

\specialrule{1.1pt}{2pt}{2pt}

\multirow{10}{*}{\textbf{arXiv}}
 & \multirow{5}{*}{\textbf{DUO}}
 & 10\% & 57.4 & 91.0 & 87.6 \\
 & & 20\% & 52.3 & 89.5 & 82.0 \\
 & & 30\% & 46.1 & 87.5 & 75.1 \\
 & & 40\% & 38.8 & 84.8 & 66.4 \\
 & & \textit{Avg} & 48.6 & 88.2 & 77.8 \\
\cmidrule(lr){2-6}
 & \multirow{5}{*}{\textbf{UDLM}}
 & 10\% & 54.9 & 87.9 & 84.6 \\
 & & 20\% & 50.1 & 85.0 & 78.0 \\
 & & 30\% & 45.1 & 81.7 & 70.7 \\
 & & 40\% & 39.6 & 77.5 & 62.3 \\
 & & \textit{Avg} & 47.4 & 83.0 & 73.9 \\

\specialrule{1.1pt}{2pt}{2pt}

\multirow{10}{*}{\cellcolor{bluerow}\textcolor{bluereg}{\textbf{LM1B}}}
 & \multirow{5}{*}{\cellcolor{bluerow}\textcolor{bluereg}{\textbf{DUO}}}
 & \cellcolor{bluerow}10\% & \cellcolor{bluerow}45.5 & \cellcolor{bluerow}87.7 & \cellcolor{bluerow}83.5 \\
 & & \cellcolor{bluerow}20\% & \cellcolor{bluerow}39.0 & \cellcolor{bluerow}84.7 & \cellcolor{bluerow}75.6 \\
 & & \cellcolor{bluerow}30\% & \cellcolor{bluerow}32.3 & \cellcolor{bluerow}80.8 & \cellcolor{bluerow}66.3 \\
 & & \cellcolor{bluerow}40\% & \cellcolor{bluerow}25.3 & \cellcolor{bluerow}75.9 & \cellcolor{bluerow}55.7 \\
 & & \cellcolor{bluerow}\textit{Avg} & \cellcolor{bluerow}35.5 & \cellcolor{bluerow}82.3 & \cellcolor{bluerow}70.2 \\
\cmidrule(lr){2-6}
 & \multirow{5}{*}{\cellcolor{bluerow}\textcolor{bluereg}{\textbf{UDLM}}}
 & \cellcolor{bluerow}10\% & \cellcolor{bluerow}47.0 & \cellcolor{bluerow}86.1 & \cellcolor{bluerow}82.2 \\
 & & \cellcolor{bluerow}20\% & \cellcolor{bluerow}41.0 & \cellcolor{bluerow}82.1 & \cellcolor{bluerow}73.9 \\
 & & \cellcolor{bluerow}30\% & \cellcolor{bluerow}34.7 & \cellcolor{bluerow}76.7 & \cellcolor{bluerow}64.1 \\
 & & \cellcolor{bluerow}40\% & \cellcolor{bluerow}28.3 & \cellcolor{bluerow}69.8 & \cellcolor{bluerow}53.2 \\
 & & \cellcolor{bluerow}\textit{Avg} & \cellcolor{bluerow}37.7 & \cellcolor{bluerow}78.7 & \cellcolor{bluerow}68.3 \\

\specialrule{1.1pt}{2pt}{2pt}

\multirow{10}{*}{\textbf{PubMed}}
 & \multirow{5}{*}{\textbf{DUO}}
 & 10\% & 56.7 & 90.2 & 86.9 \\
 & & 20\% & 50.8 & 88.6 & 81.1 \\
 & & 30\% & 44.0 & 86.4 & 73.7 \\
 & & 40\% & 35.9 & 83.3 & 64.3 \\
 & & \textit{Avg} & 46.8 & 87.1 & 76.5 \\
\cmidrule(lr){2-6}
 & \multirow{5}{*}{\textbf{UDLM}}
 & 10\% & 54.3 & 87.6 & 84.2 \\
 & & 20\% & 48.6 & 84.5 & 77.3 \\
 & & 30\% & 42.8 & 80.9 & 69.4 \\
 & & 40\% & 36.6 & 76.1 & 60.3 \\
 & & \textit{Avg} & 45.6 & 82.3 & 72.8 \\

\specialrule{1.1pt}{2pt}{2pt}

\multirow{10}{*}{\cellcolor{bluerow}\textcolor{bluereg}{\textbf{WT-103}}}
 & \multirow{5}{*}{\cellcolor{bluerow}\textcolor{bluereg}{\textbf{DUO}}}
 & \cellcolor{bluerow}10\% & \cellcolor{bluerow}57.3 & \cellcolor{bluerow}93.2 & \cellcolor{bluerow}89.6 \\
 & & \cellcolor{bluerow}20\% & \cellcolor{bluerow}50.4 & \cellcolor{bluerow}91.3 & \cellcolor{bluerow}83.1 \\
 & & \cellcolor{bluerow}30\% & \cellcolor{bluerow}42.9 & \cellcolor{bluerow}88.8 & \cellcolor{bluerow}75.0 \\
 & & \cellcolor{bluerow}40\% & \cellcolor{bluerow}34.3 & \cellcolor{bluerow}85.1 & \cellcolor{bluerow}64.8 \\
 & & \cellcolor{bluerow}\textit{Avg} & \cellcolor{bluerow}46.2 & \cellcolor{bluerow}89.6 & \cellcolor{bluerow}78.1 \\
\cmidrule(lr){2-6}
 & \multirow{5}{*}{\cellcolor{bluerow}\textcolor{bluereg}{\textbf{UDLM}}}
 & \cellcolor{bluerow}10\% & \cellcolor{bluerow}57.1 & \cellcolor{bluerow}92.0 & \cellcolor{bluerow}88.5 \\
 & & \cellcolor{bluerow}20\% & \cellcolor{bluerow}51.0 & \cellcolor{bluerow}89.2 & \cellcolor{bluerow}81.6 \\
 & & \cellcolor{bluerow}30\% & \cellcolor{bluerow}44.7 & \cellcolor{bluerow}85.6 & \cellcolor{bluerow}73.3 \\
 & & \cellcolor{bluerow}40\% & \cellcolor{bluerow}37.8 & \cellcolor{bluerow}80.2 & \cellcolor{bluerow}63.3 \\
 & & \cellcolor{bluerow}\textit{Avg} & \cellcolor{bluerow}47.7 & \cellcolor{bluerow}86.8 & \cellcolor{bluerow}76.7 \\

\specialrule{1.1pt}{2pt}{2pt}

\multirow{10}{*}{\textbf{OWT}}
 & \multirow{5}{*}{\textbf{DUO}}
 & 10\% & 63.9 & 94.8 & 91.8 \\
 & & 20\% & 58.5 & 93.2 & 86.3 \\
 & & 30\% & 51.9 & 90.9 & 79.2 \\
 & & 40\% & 44.2 & 87.6 & 70.2 \\
 & & \textit{Avg} & 54.6 & 91.6 & 81.9 \\
\cmidrule(lr){2-6}
 & \multirow{5}{*}{\textbf{UDLM}}
 & 10\% & 63.8 & 94.0 & 90.9 \\
 & & 20\% & 58.8 & 91.7 & 85.1 \\
 & & 30\% & 53.0 & 88.6 & 77.9 \\
 & & 40\% & 46.6 & 84.2 & 69.2 \\
 & & \textit{Avg} & 55.5 & 89.6 & 80.8 \\

\specialrule{1.1pt}{2pt}{2pt}
\end{tabular}%
}
\end{table}

%% file: tables/copy-rate-detail.tex
\begin{table}[t]
\centering
\footnotesize
\renewcommand{\arraystretch}{1.0}
\setlength{\tabcolsep}{4pt}
\caption{\textbf{Per-corruption-rate copy rate.} Full breakdown of the copy-rate rows of Table~\ref{tab:appendix-lambda-corruption} by corruption rate $p$, for DUO and UDLM at $\lambda=0$ (NELBO-only), $0.02$ (\textcolor{purplereg}{\textbf{our default}}), and $0.1$ (\textcolor{bluereg}{\textbf{blue}}), in \%. \textit{Avg} rows report the corruption-rate-averaged values used there.}
\label{tab:appendix-lambda-copyrate}
\resizebox{0.6\linewidth}{!}{%
\begin{tabular}{ll cGH cGH}
\specialrule{1.1pt}{2pt}{2pt}
& & \multicolumn{3}{c}{\textbf{DUO}} & \multicolumn{3}{c}{\textbf{UDLM}} \\
\cmidrule(lr){3-5}\cmidrule(lr){6-8}
\textbf{Dataset} & $\bm{p}$ & 0 & \textbf{0.02} & \textbf{0.1} & 0 & \textbf{0.02} & \textbf{0.1} \\
\specialrule{1.1pt}{2pt}{2pt}
\multirow{5}{*}{\textbf{AG News}}
 & 10\% & 0.004 & 0.03 & 0.15 & 0.007 & 0.03 & 0.08 \\
 & 20\% & 0.004 & 0.04 & 0.18 & 0.010 & 0.03 & 0.09 \\
 & 30\% & 0.004 & 0.05 & 0.23 & 0.005 & 0.03 & 0.09 \\
 & 40\% & 0.004 & 0.06 & 0.30 & 0.005 & 0.03 & 0.09 \\
 & \textit{Avg} & 0.004 & 0.05 & 0.22 & 0.007 & 0.03 & 0.09 \\
\midrule
\multirow{5}{*}{\textbf{arXiv}}
 & 10\% & 0.004 & 0.05 & 0.19 & 0.005 & 0.04 & 0.12 \\
 & 20\% & 0.005 & 0.06 & 0.21 & 0.006 & 0.04 & 0.12 \\
 & 30\% & 0.005 & 0.06 & 0.25 & 0.006 & 0.03 & 0.11 \\
 & 40\% & 0.006 & 0.07 & 0.31 & 0.007 & 0.02 & 0.09 \\
 & \textit{Avg} & 0.005 & 0.06 & 0.24 & 0.006 & 0.03 & 0.11 \\
\midrule
\multirow{5}{*}{\textbf{LM1B}}
 & 10\% & 0.002 & 0.040 & 0.18 & 0.004 & 0.03 & 0.13 \\
 & 20\% & 0.003 & 0.050 & 0.22 & 0.004 & 0.04 & 0.13 \\
 & 30\% & 0.003 & 0.057 & 0.25 & 0.005 & 0.03 & 0.12 \\
 & 40\% & 0.004 & 0.063 & 0.29 & 0.005 & 0.02 & 0.11 \\
 & \textit{Avg} & 0.003 & 0.05 & 0.24 & 0.005 & 0.03 & 0.12 \\
\midrule
\multirow{5}{*}{\textbf{PubMed}}
 & 10\% & 0.004 & 0.05 & 0.17 & 0.005 & 0.04 & 0.12 \\
 & 20\% & 0.005 & 0.06 & 0.20 & 0.006 & 0.04 & 0.12 \\
 & 30\% & 0.006 & 0.06 & 0.24 & 0.007 & 0.03 & 0.11 \\
 & 40\% & 0.006 & 0.07 & 0.29 & 0.007 & 0.03 & 0.10 \\
 & \textit{Avg} & 0.005 & 0.06 & 0.23 & 0.006 & 0.04 & 0.11 \\
\midrule
\multirow{5}{*}{\textbf{WT-103}}
 & 10\% & 0.008 & 0.04 & 0.15 & 0.004 & 0.03 & 0.12 \\
 & 20\% & 0.006 & 0.04 & 0.23 & 0.006 & 0.03 & 0.11 \\
 & 30\% & 0.005 & 0.05 & 0.25 & 0.009 & 0.03 & 0.11 \\
 & 40\% & 0.006 & 0.08 & 0.33 & 0.010 & 0.03 & 0.10 \\
 & \textit{Avg} & 0.006 & 0.05 & 0.24 & 0.007 & 0.03 & 0.11 \\
\midrule
\multirow{5}{*}{\textbf{OWT}}
 & 10\% & 0.003 & 0.03 & 0.12 & 0.002 & 0.026 & 0.09 \\
 & 20\% & 0.003 & 0.03 & 0.14 & 0.003 & 0.020 & 0.09 \\
 & 30\% & 0.003 & 0.04 & 0.17 & 0.003 & 0.020 & 0.09 \\
 & 40\% & 0.003 & 0.04 & 0.20 & 0.004 & 0.020 & 0.08 \\
 & \textit{Avg} & 0.003 & 0.04 & 0.16 & 0.003 & 0.02 & 0.09 \\
\specialrule{1.1pt}{2pt}{2pt}
\end{tabular}}
\end{table}